\def\METAPAPER{1}
\newif\ifmetapaper
\ifdefined\METAPAPER
  \metapapertrue
\else
  \metapaperfalse
\fi

\ifmetapaper
  \documentclass{fairmeta}
\else
  \documentclass{article}
  \usepackage[T1]{fontenc}
  \usepackage{iclr2027_conference,times}
\fi

\usepackage{amsmath,amssymb}
\usepackage{algorithm}
\usepackage{algpseudocode}
\usepackage{float}
\usepackage{tabularx}
\usepackage{xurl}
\ifmetapaper
\else
  \usepackage{graphicx}
  \usepackage{booktabs}
  \usepackage{multirow}
  \usepackage{hyperref}
  \usepackage{xcolor}
\fi

\definecolor{miraorange}{HTML}{D9582F}
\definecolor{miraorangebg}{HTML}{FFF3ED}

\newcommand{\bestoverall}[1]{%
  \begingroup\setlength{\fboxsep}{1pt}\fbox{#1}\endgroup}
\newcommand{\appendixcontentsline}[1]{%
  \noindent
  \hyperref[#1]{\textcolor{miraorange}{\textbf{\ref*{#1}}}\hspace{0.65em}%
  \nameref*{#1}}%
  \dotfill
  \hyperref[#1]{\textcolor{miraorange}{\pageref*{#1}}}\par}

\newcommand{\papertitle}{Learning What to Investigate Next: Meta-Reasoning for Long-Horizon Research Agents}
\title{\papertitle}

\ifmetapaper
  \author[1,2,\ast]{Ankur Samanta}
  \author[3]{Yonathan Efroni}
  \author[2]{Paul Sajda}
  \author[1,\dagger]{Kaveh Hassani}
  \author[1,\dagger]{Anirudh Goyal}
  \affiliation[1]{Meta AI}
  \affiliation[2]{Columbia University}
  \affiliation[3]{Tel Aviv University}
  \contribution[\ast]{Work done at Meta}
  \contribution[\dagger]{Joint last authors}
\else
  \author{Anonymous Authors}
\fi

\ifmetapaper
  \abstract{Long-horizon research agents must decide both how to investigate and what to
investigate next as evidence accumulates. This is hard to learn because such decisions are sparse in
long execution traces, and their consequences may emerge several
investigations later. We introduce \textbf{Meta-reasoning for Iterative
Research Agents (MIRA)}, a hierarchical architecture separating research
allocation from execution. An outer-loop meta-reasoner curates context from a
persistent research record, then writes a work order for the next investigation
or ends the episode. A fresh inner-loop executor carries out each work order,
making execution part of the transition between meta-reasoning actions. Without policy
training, MIRA improves long-horizon inference and allocates additional compute
more effectively in theorem proving and open-ended neural-architecture
research. Its decision boundaries also provide natural units for credit assignment.
At each boundary, we train a generative critic to forecast expected remaining return from
partial states, outperforming token-level alternatives. Cross-environment
pretraining improves forecasting and adaptation, yielding a transferable prior
for valuing partial progress. We use this prior to initialize
\textbf{MIRA-AC}, a generative actor--critic jointly trained to
forecast remaining return and choose the next investigation, without a
separate critic model. MIRA-AC concentrates policy optimization on meta-reasoning
decisions, enabling efficient long-horizon reinforcement learning without
directly optimizing the longer execution traces they initiate. Training MIRA-AC
on the model's own proxy hill-climbing signals improves gold performance across
four autoresearch environments; the actor transfers with cross-environment
value initialization. Together, these results show that meta-reasoning can be
learned as an explicit policy for directing long-horizon autonomous research.
}
  \correspondence{Ankur Samanta (\email{as7416@columbia.edu}), Anirudh Goyal (\email{agi@meta.com})}
  \date{\today}
\else
\fi

\begin{document}

\maketitle

\ifmetapaper
\else
  \begin{abstract}
  Long-horizon research agents must decide both how to investigate and what to
investigate next as evidence accumulates. This is hard to learn because such decisions are sparse in
long execution traces, and their consequences may emerge several
investigations later. We introduce \textbf{Meta-reasoning for Iterative
Research Agents (MIRA)}, a hierarchical architecture separating research
allocation from execution. An outer-loop meta-reasoner curates context from a
persistent research record, then writes a work order for the next investigation
or ends the episode. A fresh inner-loop executor carries out each work order,
making execution part of the transition between meta-reasoning actions. Without policy
training, MIRA improves long-horizon inference and allocates additional compute
more effectively in theorem proving and open-ended neural-architecture
research. Its decision boundaries also provide natural units for credit assignment.
At each boundary, we train a generative critic to forecast expected remaining return from
partial states, outperforming token-level alternatives. Cross-environment
pretraining improves forecasting and adaptation, yielding a transferable prior
for valuing partial progress. We use this prior to initialize
\textbf{MIRA-AC}, a generative actor--critic jointly trained to
forecast remaining return and choose the next investigation, without a
separate critic model. MIRA-AC concentrates policy optimization on meta-reasoning
decisions, enabling efficient long-horizon reinforcement learning without
directly optimizing the longer execution traces they initiate. Training MIRA-AC
on the model's own proxy hill-climbing signals improves gold performance across
four autoresearch environments; the actor transfers with cross-environment
value initialization. Together, these results show that meta-reasoning can be
learned as an explicit policy for directing long-horizon autonomous research.

  \end{abstract}
\fi

\section{Introduction}
\label{sec:intro}
Long-horizon research requires more than executing individual
investigations well.
As evidence accumulates, an agent must decide which hypothesis to test,
which result to verify, whether to refine or abandon a direction, and
when to stop.
These decisions draw on evidence from many
experiments~\citep{gao_autoscientists_2026,zou_fml-bench_2026}, which can
quickly exceed the model's context window~\citep{kausik2026cgdp}. The decisions
themselves are sparse within much longer execution traces, and their
consequences may be delayed~\citep{sutton_barto_2018,samanta_credit_2026}:
an unsuccessful experiment can eliminate an important hypothesis, while
a locally promising direction can ultimately lead to a dead end.
In this sense, the quality of a research trajectory often depends less on the
outcome of any single investigation than on the choices that determine what to
investigate next.
Learning to conduct research therefore requires isolating the decisions that
allocate future research effort from the executions they initiate and
assigning them credit for their downstream consequences. This motivates the central question of this work:

\begin{center}
    \emph{Can a research agent learn what to investigate next from the
    downstream consequences of its previous research decisions?}
\end{center}

We view this as a problem of \emph{meta-reasoning}: deciding which
computation a resource-bounded agent should perform next and when it should stop
\citep{russell_principles_1991}.
For research agents, we take the relevant computation to be an
\emph{investigation}.
\ifmetapaper
Sec.~\ref{sec:formalism} formalizes research at the level of investigations.
\fi
Sec.~\ref{sec:rq1} introduces \textbf{MIRA}, a hierarchical architecture that separates
deciding what to investigate from carrying out the investigation.
At decision $t$, the outer loop curates evidence from a persistent research
record into an observation $o_t$. The meta-reasoning policy then samples a
complete decision response $a_t$, comprising its reasoning followed by either a
work order or a termination decision.
For a work order, a fresh executor carries out the investigation and
updates the persistent research record.

At the meta-reasoning level, the potentially long executor trace is therefore
compressed into one decision-level transition
$\xi_t=(o_t,a_t,r_t,o_{t+1},d_t)$, where $d_t$ denotes true episode termination.
The complete decision response
is the action; its implementation, experimentation, and tool use are part of
the environment transition. This isolates the decisions that determine how
future research effort is allocated from the longer computation used to
execute them.

\begin{figure}[t]
  \centering
  \includegraphics[width=\linewidth]
    {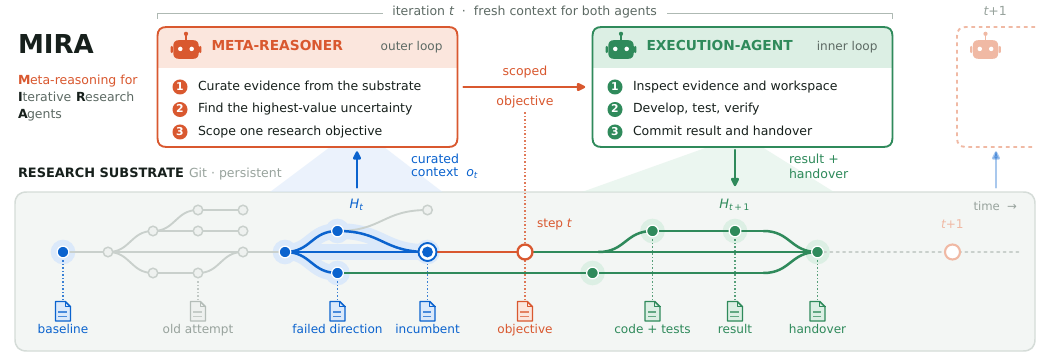}
  \caption{
  \textbf{MIRA separates research allocation from execution.}
At each research-decision boundary, the outer loop reads the persistent
research record $H_t$ and curates evidence bearing on the most consequential
unresolved uncertainty into an observation $o_t$. It then produces a
meta-reasoning action $a_t$, a complete decision response comprising its
reasoning followed by either a work order or a stopping decision.
For a work order, a fresh executor carries out the investigation
and writes its results back to the persistent record, producing
$H_{t+1}$.
The next outer-loop observation is then constructed from this updated
record; a stopping decision instead terminates the research episode.
Thus, each investigation, however long, is represented as one
transition at the meta-reasoning level.}
  \label{fig:mira-harness}
\end{figure}

Sec.~\ref{sec:rq2} develops a model of remaining research value.
Making research decisions explicit also creates a natural boundary for value
prediction and credit assignment. Because an investigation's immediate outcome
need not reflect its eventual contribution, we learn its expected remaining
research return $V^\pi(o_t)=\mathbb E_\pi[G_t\mid o_t]$.
We train a decision-level generative critic to forecast
this quantity once per research decision from the same observation available
to the policy. It uses the language model's native next-token distribution to
represent beliefs over future returns~\citep{zhang_generative_verifiers_2024,
bellemare_distributional_2017}. The critic forecasts the future potential of
the current curated research observation rather than ranking candidate work orders.

Sec.~\ref{sec:rq3} uses this signal to assign delayed credit to the research
policy. We introduce \textbf{MIRA-AC}, a shared generative actor--critic jointly
trained to forecast remaining return and choose what to investigate next from
each curated research observation. Because one model plays both roles, no
separate critic model is required. MIRA-AC reconstructs complete research
episodes to assign delayed credit to research decisions.
A single decision-level advantage is
propagated to every sampled
token in the corresponding decision response, including its reasoning and
resulting work order or termination decision. Context curation, executor
generations, experiments, and tool calls receive no direct policy loss.
MIRA-AC therefore concentrates policy optimization on \emph{what to investigate
next} without needing to directly optimize the much longer execution
trajectories. We train it with a buffered single-rollout asynchronous
optimization scheme~\citep{sao} that accommodates variable-duration research
episodes. Overall, we show that jointly training value forecasting and
meta-reasoning from the same curated research observations---while assigning credit
directly to the resulting decisions---can produce a stronger policy.

Sec.~\ref{sec:experiments} evaluates each stage in turn. First, untrained MIRA
improves long-horizon inference in theorem proving on
IMOProofBench~\citep{luong_towards_2025} and in open-ended neural-architecture
research with Residual Matrix Transformers~\citep{mak_residual_2025} and Loop
Transformers~\citep{geiping_scaling_2025}. In the latter settings, it selects
per-cell residual-write gates for RMT and signed attention with a leaky
recurrent update for the Loop Transformer. Second, a decision-level generative
critic forecasts remaining return more accurately than token-level alternatives,
and cross-environment pretraining improves value forecasting and adaptation.
Third, under proxy supervision, the strongest decision-level policy-learning
configuration improves the primary outcome over token-level regression across
symbolic regression~\citep{shojaee_llmsrbench_2025}, physics
discovery~\citep{wiemann_discoverphysics_2026}, Bayesian causal discovery on
networks from the Bayesian Network Repository~\citep{scutari_bnrepository}, and
CPU-based neural-architecture research. Crucially, training on proxy
hill-climbing feedback collected from the model's own research trajectories
improves gold-evaluation performance across all four environments. MIRA-AC
provides additional gains in several environments, and a policy trained without BNLearn
actor updates transfers to BNLearn when initialized from a critic pretrained
across environments, including BNLearn. This separation between proxy training
and gold evaluation mirrors autoresearch settings in which proxy evaluations
support inexpensive iteration, while gold evaluation may, for example, require
conducting in vivo experiments~\citep{stokes_deep_2020} or running a large-scale
production system~\citep{novikov_alphaevolve_2025}. Together, these results
show that research allocation can be effectively governed by an explicit,
learnable meta-reasoning policy rather than remaining an implicit byproduct of
long-horizon generation.

\ifmetapaper
  \section{Research as a Sequential Decision Process}
\label{sec:formalism}
\label{sec:prelims}

Long-horizon research interleaves two kinds of computation:
deciding \emph{what to investigate next} and carrying out the chosen
investigation.
Our central modeling choice is to place the learning boundary between
them.
One research decision is one action; when that action selects an investigation,
the potentially much longer computation used to execute it is part of the
environment transition.

\subsection{Decision Process}
\label{sec:decision-process}

Let $H_t$ denote the persistent research state immediately before
research decision $t$.
It contains the current artifact together with evidence accumulated from
earlier investigations and the remaining research budget.
Meta-reasoning includes both evidence curation and action selection. In this
work, we hold curation fixed and train the core action-selection policy. A fixed
context-curation procedure $C$ constructs the observation $o_t=C(H_t)$ presented
to that policy.
Retrieval and summarization used to construct $o_t$ belong to observation
construction rather than the action.

Conditioned on $o_t$, the policy selects
$a_t\sim\pi_\theta(\cdot\mid o_t)$, where $a_t$ is the complete sampled
meta-reasoning response: its emitted
reasoning followed by either a work order or a termination decision.
A work order specifies a research objective, a proposed approach, and
criteria for determining whether the investigation succeeded.

For a work order, an executor $\mathcal E$ carries out the investigation.
Its execution may contain many model generations, tool calls,
experiments, and revisions, but remains part of a single environment
transition, $H_t\xrightarrow{a_t,\mathcal E}H_{t+1}$.

Let $d_t\in\{0,1\}$ indicate true termination after decision $t$.
For a non-terminal transition, the next observation is
$o_{t+1}=C(H_{t+1})$.
For a terminal transition, we write $o_{t+1}=\bot$ and use the
convention $V(\bot)=0$.
A termination decision is itself a terminal research action.

At the meta-reasoning level, the stored transition is therefore
$\xi_t:=(o_t,a_t,r_t,o_{t+1},d_t)$.
Context resets, executor restarts, and learner-batch boundaries do not
set $d_t=1$.

Writing $T=\min\{t:d_t=1\}$, a logical research episode is
$\tau=(\xi_1,\ldots,\xi_T)$.

Thus, the unit of control is the research decision: the policy chooses
\emph{what} to investigate or whether to terminate, while the executor
determines \emph{how} a selected investigation is carried out.

\subsection{Research Feedback and Return}
\label{sec:returns}

Our environments expose proxy feedback during research and apply a separate
gold evaluation only after final submission. The gold evaluator reports the
task-specific final metrics and is never available to the agent during the
research episode.

\paragraph{Proxy feedback.}
Let $q_t\in[0,1]$ be the normalized proxy quality after decision $t$,
with $q_0$ the initial quality, and define
$\bar q_t=\max_{0\leq u\leq t}q_u$. The proxy reward is the improvement in
the incumbent, $r_t^{\mathrm{proxy}}=\bar q_t-\bar q_{t-1}$. Hence the
remaining proxy return telescopes:
$G_t^{\mathrm{proxy}}=\sum_{u=t}^{T}r_u^{\mathrm{proxy}}
=\bar q_T-\bar q_{t-1}$.
Each environment uses a monotone proxy normalization fixed before training.
The proxy and gold evaluations need not rank artifacts identically.

\paragraph{Gold-binary feedback.}
Gold evaluation may contain continuous metrics as well as an
environment-specific binary success criterion. For the \emph{Gold binary}
training condition, let $Y\in\{0,1\}$ denote this terminal success outcome.
Intermediate rewards are zero and the terminal reward is $Y$, so
$G_t^{\mathrm{gold\text{-}bin}}=Y$.

For either feedback signal, we write $G_t=\sum_{u=t}^{T}r_u$ for the
undiscounted remaining return.
Environment-specific definitions of $q_t$ and $Y$ are given in
App.~\ref{sec:environments}.

\subsection{Remaining Research Value}
\label{sec:research-value}

For a fixed continuation policy and research budget, we define
$V^\pi(o_t)=\mathbb E_\pi[G_t\mid o_t]$.
Because $o_t$ is a curated observation, this is an
observation-conditioned value rather than a value of the complete
research state. Under proxy feedback, $V^\pi(o_t)$ is the expected additional
best-so-far improvement from continuing research.
Under gold-binary feedback, it is the probability of success
under the continuation policy. It is not the maximum improvement theoretically
attainable from the underlying research state.

The remainder of the method instantiates this decision process:
MIRA exposes the research actions, a critic forecasts remaining value from the
resulting observations, and MIRA-AC assigns downstream credit to the policy
that selects them.

\fi

\section{MIRA: Exposing Meta-Reasoning Decisions}
\label{sec:rq1}

The decision-level formulation in Sec.~\ref{sec:formalism} requires an
agent architecture that exposes individual investigations as explicit actions
while preserving evidence across a much longer research trajectory.
In particular, the system must maintain research state across investigations,
separate the choice of what to investigate from the computation used to carry
it out, and produce a well-defined decision boundary at which outcomes can
later be valued and credited.
MIRA realizes these requirements with a persistent research substrate, an
outer-loop meta-reasoner, and a fresh inner-loop executor.

\subsection{MIRA Architecture}
\label{sec:mira}
Separating research allocation from execution requires evidence to persist
across investigations. We thus introduce MIRA, a hierarchical research harness in which an
outer-loop meta-reasoner and an inner-loop executor are instantiated afresh at
each iteration over a persistent research substrate.

\paragraph{Persistent research substrate.}
Each task has a Git repository that persists throughout the research episode.
It stores the current artifact, prior experiments and their outcomes, analyses,
code or proof artifacts, and handovers from earlier executors. The repository
implements the persistent research state $H_t$; model contexts are temporary.

\paragraph{Outer-loop meta-reasoner.}
At each decision boundary, the context-curation procedure constructs $o_t$ from
the persistent record. The meta-reasoning policy then samples the complete
decision response $a_t$: its emitted reasoning followed by either a work order
or a termination decision. A work order specifies the next objective, proposed
approach, and success criteria. The outer loop therefore allocates the next
unit of research effort.

\paragraph{Fresh inner-loop executor.}
Each work order launches a fresh execution agent through the Codex
harness~\citep{openai2025codex}, with the underlying model specified by the
experiment. It receives the fixed task instructions, the work order, and access
to the persistent workspace and task-specific tools. The executor performs the investigation,
updates the artifact, and commits a handover recording its evidence and
unresolved questions. Both outer- and inner-loop contexts reset between
decisions; only the persistent research substrate survives.
Fig.~\ref{fig:mira-harness} illustrates this separation between research
allocation and execution.

\subsection{MIRA Research Procedure}
\label{sec:mira-procedure}

Algorithm~\ref{alg:mira} formalizes this research procedure. Each outer-loop decision produces exactly one
decision-level transition, regardless of the length of the corresponding
executor trace. The shared procedure and executor interface remain fixed across
tasks; App.~\ref{app:mira-implementation} gives the context construction,
evaluation, and submission details.

\begin{algorithm}[t]
\caption{MIRA Research Episode}
\label{alg:mira}
\begin{algorithmic}[1]
\Require Initial research state $H_1$, maximum horizon $T_{\max}$
\Require Curator $C$, policy $\pi_\theta$, executor $\mathcal E$
\State $t\gets1$
\State $o_1\gets C(H_1)$
\While{true}
    \State Sample $a_t\sim\pi_\theta(\cdot\mid o_t)$
    \If{$a_t$ terminates the episode}
        \State Submit the current artifact and observe terminal reward $r_t$
        \State $d_t\gets1$, $o_{t+1}\gets\bot$
    \Else
        \State Execute the work order in $a_t$ with $\mathcal E$ and update $H_{t+1}$
        \State Observe $r_t$ and $d_t$, with $d_t=1$ at the horizon $t=T_{\max}$
        \State $o_{t+1}\gets
            \begin{cases}
                C(H_{t+1}), & d_t=0,\\
                \bot, & d_t=1
            \end{cases}$
    \EndIf
    \State Store $\xi_t=(o_t,a_t,r_t,o_{t+1},d_t)$
    \If{$d_t=1$}
        \State \Return $\tau=(\xi_1,\ldots,\xi_t)$
    \EndIf
    \State $t\gets t+1$
\EndWhile
\end{algorithmic}
\end{algorithm}

\subsection{Learning Boundary}
\label{sec:training-boundary}

MIRA's architectural separation also determines where MIRA-AC applies
learning. It optimizes the policy that generates decision responses and the
critic that forecasts remaining return, while holding curation, execution, and
the task environment fixed.

\begin{table}[t]
\centering
\caption{\textbf{Optimization boundary in MIRA-AC.}}
\label{tab:training-boundary}
\small
\begin{tabular}{lll}
\toprule
Component & Role & Trained? \\
\midrule
Curator $C$ & Constructs $o_t$ & No \\
Policy $\pi_\theta$ & Generates reasoning and work order / termination & Yes \\
Value critic & Forecasts remaining return & Yes \\
Executor $\mathcal E$ & Implements investigation & No \\
Tools / evaluators & Environment interaction & No \\
Git substrate $H_t$ & Persistent research state & --- \\
\bottomrule
\end{tabular}
\end{table}

Making research allocation explicit identifies where to assign credit. The
remaining challenge is that a decision's consequences may become clear only
after several later investigations; Sec.~\ref{sec:rq2} forecasts remaining
value from the resulting observations, and Sec.~\ref{sec:rq3} uses it for
policy optimization.

\section{Learning to Forecast the Future Value of Research Progress}
\label{sec:rq2}
\label{sec:value}

\begin{figure}[t]
  \centering
  \includegraphics[width=\linewidth]
    {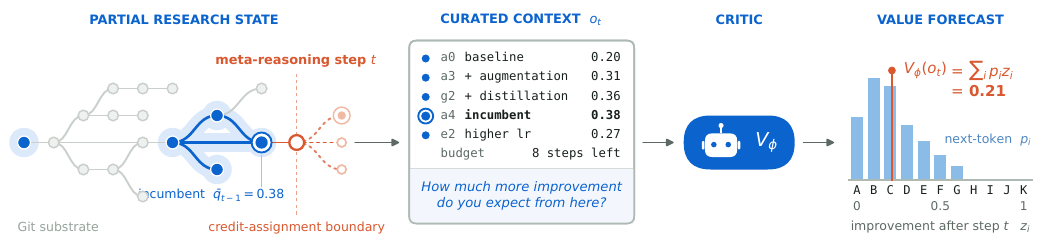}
  \caption{\textbf{Forecasting remaining research progress.}
  The critic receives the same curated observation $o_t$ as the policy and
  forecasts a distribution over remaining return. Its expectation is
  $V_\phi(o_t)$.}
  \label{fig:step-value-forecast}
\end{figure}

An investigation's immediate outcome need not reflect its eventual
contribution. A failed experiment may eliminate an incorrect hypothesis and
enable later progress, while an immediate improvement may lead to a dead end.
We therefore learn a value function $V^\pi(o_t)$ at MIRA's
research-decision boundaries.

The environments report progress in different raw units, so we use the gold
and proxy returns defined in Sec.~\ref{sec:returns}. Each environment maps its
proxy metric through a fixed environment-specific transform to a quality in
$[0,1]$.
These transforms are scaled to distinguish improvements attainable within the
research budget and convert partial trajectories into marginal rewards.
App.~\ref{sec:proxy-signals} describes their anchors, relation to gold
evaluation, and construction for new environments.

\ifmetapaper
\else
\ifmetapaper
\paragraph{Episodes and rewards.}
\else
\noindent\textbf{Returns and value.}
\label{sec:returns}
\fi
An episode consists of successive meta-reasoning decisions and their outcomes,
$\tau=(o_1,a_1,r_1,\ldots,o_T,a_T,r_T,o_{T+1})$, where $T\leq T_{\max}$. The
undiscounted return from step $t$ is $G_t=\sum_{u=t}^{T}r_u$. We consider two
reward signals according to the feedback available in the research environment.

\emph{Gold-binary feedback.} Let $Y\in\{0,1\}$ be the environment-specific
solved indicator produced by terminal gold evaluation and never exposed
during research. For HillClimbBench-CPU, $Y=1$ when the lower endpoint of the
paired 95\% confidence interval for improvement over the task reference
architecture is positive. Reward is zero before termination and $r_T=Y$, so
the return from every decision is $G_t^{\mathrm{gold}}=Y$.

\emph{Proxy-marginal feedback.} Let $q_t\in[0,1]$ denote normalized proxy
quality after decision $t$, and let $\bar q_t=\max_{0\leq u\leq t}q_u$ be the
best quality observed so far, including evaluations performed by the executor.
Each environment uses a normalization fixed before training and monotone in its
designated proxy metric; App.~\ref{sec:proxy-signals} gives their anchors and
App.~\ref{sec:environments} gives the four exact mappings. We define
$r_t^{\mathrm{proxy}}=\bar q_t-\bar q_{t-1}$, giving
$G_t^{\mathrm{proxy}}=\bar q_T-\bar q_{t-1}$. This reward assigns immediate
credit when a new incumbent is found while allowing earlier decisions to
receive credit for improvements achieved later. The proxy metric provides a
training signal but need not coincide with either the gold score produced by
final evaluation or the gold-binary outcome.

\ifmetapaper
\paragraph{Value of a curated research observation.}
\fi
For a fixed continuation policy $\pi$, the value of an observation is its
expected remaining return, $V^\pi(o_t)=\mathbb{E}_\pi[G_t\mid o_t]$. Under
gold-binary feedback, this is the probability of eventual success,
$V_{\mathrm{gold}}^\pi(o_t)=\Pr_\pi(Y=1\mid o_t)$. Under proxy-marginal
feedback, it is the expected additional improvement over the incumbent,
$V_{\mathrm{proxy}}^\pi(o_t)=\mathbb{E}_\pi[\bar q_T-\bar q_{t-1}\mid o_t]$.
These values depend on the continuation policy and remaining budget: they
describe the expected outcome of continuing from the current observation, not
the maximum improvement theoretically achievable from the underlying state.

\fi

\subsection{Decision-Level Generative Critic}
\label{sec:value_function}

The critic produces one value forecast immediately before each research
action. This aligns value forecasting with the unit of control, rather than with
intermediate token prefixes inside the language-model generation of a decision
response.

We parameterize the critic using the language model's next-token distribution.
Following \citet{samanta_bayesbench_2026}, we elicit its beliefs by renormalizing
the next-token log probabilities assigned to a set of single-token labels, each
representing a possible future-return value. Let
$\mathcal Z=\{z_1<\cdots<z_K\}$ denote these values. For observation $o_t$,
$p_\phi(z_i\mid o_t)=\exp\ell_{\phi,i}(o_t)/
\sum_{k=1}^{K}\exp\ell_{\phi,k}(o_t)$, and
$V_\phi(o_t)=\sum_{i=1}^{K}p_\phi(z_i\mid o_t)z_i$.

We use support $\{0.0,0.1,\ldots,1.0\}$ for proxy returns and
$\{0,1\}$ for gold-binary outcomes. The critic forecasts expected remaining
return under the continuation policy; it does not estimate maximum attainable
improvement or score candidate work orders.

\subsection{Critic Objective}
\label{sec:critic-objective}

The training target is the Monte Carlo return $G_t$ from the complete research
episode. For $z_j<G_t<z_{j+1}$, we project the target onto its neighboring
support points by setting
$y_j(G_t)=(z_{j+1}-G_t)/(z_{j+1}-z_j)$ and
$y_{j+1}(G_t)=(G_t-z_j)/(z_{j+1}-z_j)$, with all other components zero. If
$G_t=z_j$, then $y_j(G_t)=1$. The
projection preserves the target expectation.

For a batch $\mathcal B$, the critic minimizes
\begin{equation}
    \mathcal L_{\mathrm{critic}}(\phi)
    =
    \frac{1}{|\mathcal B|}
    \sum_{t\in\mathcal B}
    D_{\mathrm{KL}}
    \left(
        y(G_t)
        \,\Vert\,
        p_\phi(\cdot\mid o_t)
    \right).
    \label{eq:critic-loss}
\end{equation}
Operationally, we select the model's next-token log probabilities for the
$K$ support labels and renormalize them over that support with a log-softmax.
We then evaluate the KL objective above; its parameter-dependent term is
$-\sum_i y_i(G_t)\log p_\phi(z_i\mid o_t)$, because the target entropy is
constant with respect to $\phi$.

\subsection{Cross-Environment Value Pretraining}
\label{sec:value-transfer}

Research environments differ in artifacts, tools, and objectives, but partial
research observations may share predictive structure. For a target environment
$d\in\mathcal D$, leave-one-environment-out (LOO) pretraining uses trajectories
from the source set $\mathcal D\setminus\{d\}$, after which the critic is adapted
on target-environment support episodes and evaluated on held-out target query
episodes. The experiments below compare this initialization with target-only
and single-source training.

Value forecasting alone does not change the research policy. The next section
uses downstream outcomes to assign credit to the decisions that select what to
investigate.

\section{MIRA-AC: Learning What to Investigate Next}
\label{sec:rq3}
\label{sec:mira-ac}

We now ask whether value estimates at MIRA's decision
boundaries can improve the policy that selects research
objectives. We introduce MIRA-AC, a shared generative
actor--critic that assigns delayed credit across complete
research episodes and uses buffered asynchronous
optimization to accommodate their variable durations.

\begin{algorithm}[t]
\caption{MIRA-AC Training}
\label{alg:mira-ac}
\begin{algorithmic}[1]
\Require Critic-pretrained shared parameters $\theta_0$
\Require Research environments $\mathcal D$
\State $\theta\gets\theta_0$, $k\gets0$; publish $\theta^0\gets\theta$
\State Initialize actor buffer $\mathcal B_{\mathrm{actor}}$
\State Initialize critic buffer $\mathcal B_{\mathrm{critic}}$
\State Launch active MIRA episodes from $\mathcal D$ using $\theta^0$
\While{training budget remains}
    \State Advance active episodes asynchronously
    \State At each decision boundary, adopt the latest $\theta^k$ and record
           versioned behavior log-probabilities and critic baselines
    \For{each newly completed episode $\tau$}
        \State Immediately launch a replacement episode if budget remains
        \State Stitch stored fragments into the complete logical episode $\tau$
        \State Compute $G_t=\sum_{u=t}^{T}r_u$ for all $t$
        \State Add $\{(o_t,G_t)\}_{t=1}^{T}$ to
               $\mathcal B_{\mathrm{critic}}$
        \State Compute bridged $\{\hat A_t\}_{t=1}^{T}$ using recorded critic versions and
               Eq.~\ref{eq:decision-gae}
        \State Add actions, advantages, behavior log-probabilities, and actor--critic
               version metadata to $\mathcal B_{\mathrm{actor}}$
    \EndFor
    \State $\mathrm{updated}\gets\mathrm{false}$
    \If{$\mathcal B_{\mathrm{critic}}$ is ready}
        \State Update $\theta$ using $\mathcal L_{\mathrm{critic}}$
        \State Remove consumed critic examples; $\mathrm{updated}\gets\mathrm{true}$
    \EndIf
    \If{$\mathcal B_{\mathrm{actor}}$ is ready}
        \State Update $\theta$ using $\mathcal L_{\mathrm{actor}}$
        \State Remove consumed actor examples; $\mathrm{updated}\gets\mathrm{true}$
    \EndIf
    \If{$\mathrm{updated}$}
        \State $k\gets k+1$; atomically publish $\theta^k\gets\theta$
    \EndIf
\EndWhile
\end{algorithmic}
\end{algorithm}

\subsection{Shared Actor and Critic}
\label{sec:training}

MIRA-AC uses one language model with shared parameters $\theta$. In actor mode
it represents $\pi_\theta$; in critic mode its restricted next-token
distribution defines $V_\theta$ as in Sec.~\ref{sec:value_function}.

The shared model is initialized from a pretrained critic checkpoint. Actor and
critic objectives use separate optimizer states while updating the shared
model. Thus, unlike a separate actor--critic, value learning also updates
representations used to generate subsequent research decisions.

\subsection{Credit Assignment Across Research Decisions}
\label{sec:batch_online}

MIRA-AC reconstructs each logical episode $\tau$ before computing learning
targets. Context resets and learner-batch boundaries may partition an episode
in storage, but only $d_t=1$ stops value propagation.

The critic is trained on the exact return $G_t$. For actor training, we use
generalized advantage estimation~\citep{schulman_gae_2016} with $\gamma=1$ and
$\lambda_T=1-1/(\alpha T)$, with the fixed hyperparameter $\alpha$ given in
Table~\ref{tab:mira-ac-hparams}.

Because asynchronous collection allows successive decisions in one episode to
use different critic versions, let $c_t$ denote the critic version used at
decision $t$. At rollout time we record the current baseline
$v_t=V_{\theta^{c_t}}(o_t)$ and same-version successor bootstrap
$\widetilde v_{t+1}^{(c_t)}=V_{\theta^{c_t}}(o_{t+1})$.
The temporal-difference residual is
$\delta_t=r_t+\gamma(1-d_t)\widetilde v_{t+1}^{(c_t)}-v_t$.

At a non-terminal boundary where the critic version changes, define
$B_t=v_{t+1}-\widetilde v_{t+1}^{(c_t)}$, where
$v_{t+1}=V_{\theta^{c_{t+1}}}(o_{t+1})$; otherwise $B_t=0$. Advantages are
computed backward as
\begin{equation}
    \hat A_t
    =
    \delta_t
    +
    \gamma\lambda_T(1-d_t)
    \left(
        \hat A_{t+1}+B_t
    \right).
    \label{eq:decision-gae}
\end{equation}

The bridge aligns critic baselines across model versions; it does not modify
the Monte Carlo target. App.~\ref{app:trajectory_stitching} gives the complete
episode-reconstruction details.

\subsection{Decision-Level Actor Objective}
\label{sec:actor-objective}

Although each research action receives one decision-level advantage, the
language model generates it autoregressively. Let $a_{t,j}$ denote token $j$ of
action $a_t$, let $L_t$ be its sampled length, and let $b_t$ denote the
behavior-policy version that generated it. Define
$\rho_{t,j}=\pi_\theta(a_{t,j}\mid o_t,a_{t,<j})/
\pi_{\theta^{b_t}}(a_{t,j}\mid o_t,a_{t,<j})$.
We mask policy-gradient terms outside the allowed importance-ratio range using
$m_{t,j}=\mathbf 1[1-\epsilon_{\mathrm{low}}<\rho_{t,j}<
1+\epsilon_{\mathrm{high}}]$.

For a batch of $N$ research decisions, the actor objective is
\begin{equation}
\begin{split}
    \mathcal L_{\mathrm{actor}}(\theta)
    =
    \frac{1}{N}
    \sum_{t=1}^{N}
    \frac{1}{L_t}
    \sum_{j=1}^{L_t}
    \left[
        -m_{t,j}\tau_A\hat A_t\rho_{t,j}
        +
        \tau_{\mathrm{KL}}
        (\log\rho_{t,j})^2
    \right].
\end{split}
    \label{eq:actor-loss}
\end{equation}

Tokens outside the ratio interval are excluded from the policy-gradient term
but remain in the policy-mismatch penalty. The trainable span is the complete
sampled meta-reasoning decision response, including its emitted reasoning and
resulting work order or termination decision. Prompts, curated observations,
executor generations, tool calls, and experiments are masked.

\subsection{Buffered Single-Rollout Asynchronous Policy Optimization}
\label{sec:async}

Investigations have heterogeneous wall-clock durations, so synchronized
rollout cohorts can leave capacity idle. MIRA-AC instead collects completed
episodes continuously into separate actor and critic buffers, which trigger
updates independently on the shared model. If both are ready, the critic update
is applied first, followed by the actor update within the same atomic update
transaction. The resulting parameters are published once. A live episode adopts
newly published parameters only at its next research-decision boundary and may
therefore span multiple model versions.

\begin{figure*}[t]
  \centering
  \includegraphics[width=\textwidth]
    {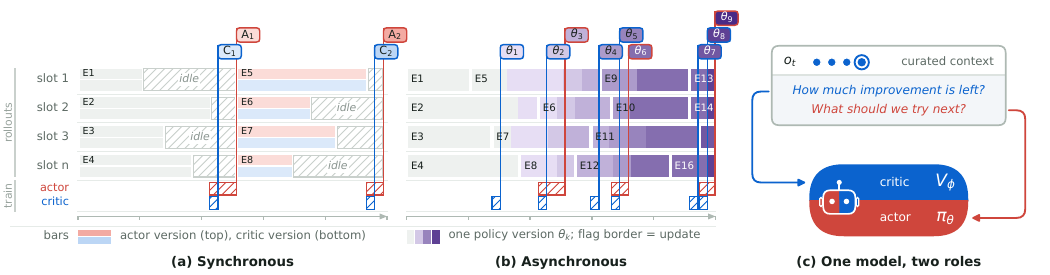}
  \caption{\textbf{Decoupled buffers allow MIRA-AC to learn asynchronously
  from variable-duration research episodes.} \textbf{(a)} Synchronous cohorts
  leave rollout slots idle while waiting for slow episodes. \textbf{(b)}
  MIRA-AC restarts freed slots immediately and updates the shared model whenever
  either the actor or critic buffer fills; live episodes adopt new weights at the
  next meta-reasoning boundary. \textbf{(c)} The same language model serves as
  actor and critic, forecasting remaining return and generating the next
  meta-reasoning step.}
  \label{fig:mira-ac-sync-vs-async}
\end{figure*}

Algorithm~\ref{alg:mira-ac} summarizes training. Buffer thresholds, optimizer
settings, rollout limits, and details of low-rank adaptation
(LoRA)~\citep{hu_lora_2022} are given in
App.~\ref{app:implementation-details}.

\FloatBarrier

\section{Experiments}
\label{sec:experiments}

We evaluate the approach in four stages.
First, we ask whether MIRA improves long-horizon inference before policy
training.
Second, we test whether remaining research value can be forecast at
MIRA's decision boundaries and whether this forecasting capability transfers
across research environments.
Third, we evaluate whether decision-level credit improves the research
policy and whether the learned policy transfers across environments.
Finally, we analyze how training changes research behavior and how MIRA's
decision-level decomposition affects where policy learning is applied and
the cost of collecting long-horizon trajectories.

These stages place different demands on the experimental setting.
To evaluate MIRA before learning, we use full-scale long-horizon settings
in theorem proving and open-ended neural-architecture research.
Value pretraining and policy learning, however, require many complete
research trajectories.
We therefore evaluate the learning components in scalable autoresearch
environments that support repeated training and held-out evaluation.

\subsection{Does MIRA Improve Long-Horizon Inference Without Policy Training?}
\label{sec:exp-mira}

\paragraph{Experimental setup.}
All comparisons in this subsection use frozen model weights. We compare GPT-5.5,
Codex, and MIRA.
GPT-5.5 uses the task interface directly; Codex operates as one persistent
tool-using session; and MIRA assigns each investigation to a fresh Codex inner
loop. The Codex and MIRA conditions use the same underlying
GPT-5.5 model, Codex harness, and tools, while MIRA adds explicit meta-reasoning
and curated context resets.
We test the three conditions on IMOProofBench~\citep{luong_towards_2025}
and HillClimbBench-RMT, and compare MIRA with
Codex on HillClimbBench-Loop. For theorem
proving, we also include output-token-matched
continuation controls.

\begin{figure}[t]
  \centering
  \begin{minipage}[t]{0.49\linewidth}
    \centering
    \includegraphics[width=\linewidth]
      {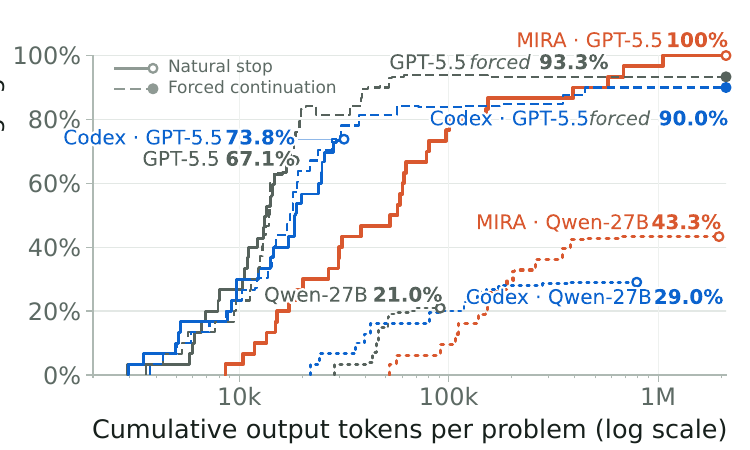}
    \textbf{(a)} IMOProofBench
  \end{minipage}\hfill
  \begin{minipage}[t]{0.49\linewidth}
    \centering
    \includegraphics[width=\linewidth]
      {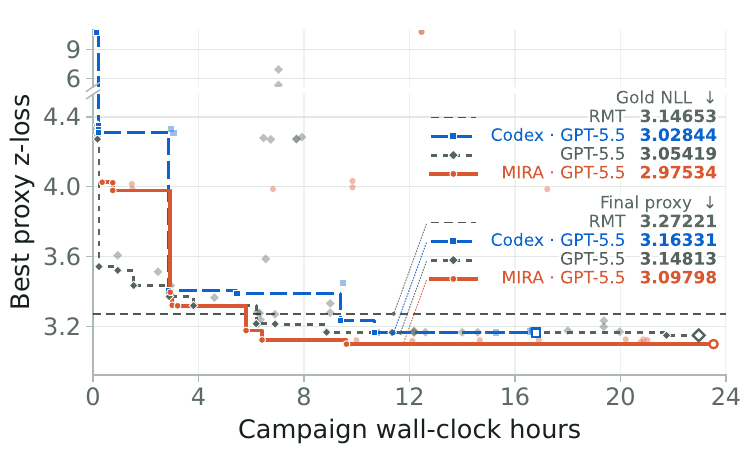}
    \textbf{(b)} HillClimbBench-RMT
  \end{minipage}
  \caption{\textbf{MIRA improves long-horizon theorem proving and
  neural-architecture hill climbing, outperforming baselines.}
  \textbf{(a)} Theorem-proving performance at
  natural stopping and under output-token-matched
  continuation. \textbf{(b)} Best loss observed
  during the matched 24-hour architecture-research
  campaigns.}
  \label{fig:rq1-hillclimbs}
\end{figure}

\paragraph{Theorem proving.}
\label{sec:rq1-imo}

\begin{figure}[t]
  \centering
  \includegraphics[width=\linewidth]
    {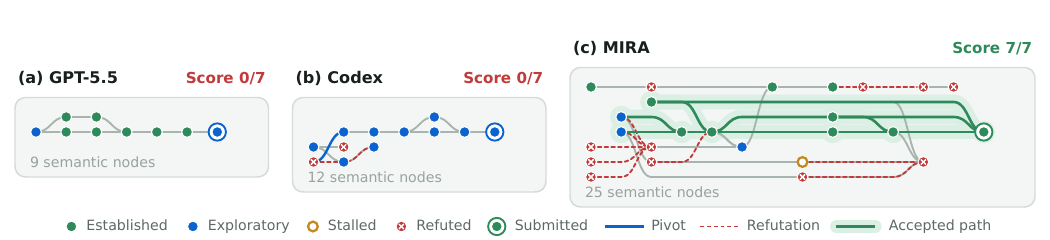}
  \caption{\textbf{MIRA more effectively allocates additional
  inference across proof directions.}
  On IMOProofBench Advanced Problem 23, GPT-5.5 and Codex pursue a small
  number of proof directions, while MIRA tests several directions, records
  refutations, and reaches the accepted proof.}
  \label{fig:imoproofbench-pb23-ribbon}
\end{figure}

On IMOProofBench Advanced, we count a problem as solved only when the grader
awards full credit. Using GPT-5.5, MIRA raises the
mean normalized grader score from 67.1\% under direct inference
to 100\%, solving all 30 problems in
each of three trials. With Qwen3.6-27B~\citep{qwen_qwen36_27b_2026},
MIRA also more than doubles
the model's score
(Fig.~\ref{fig:rq1-hillclimbs}(a)).
All conditions receive the same runtime budget, but MIRA makes better use of
its inference budget to iterate on its solution, while the baselines submit
much earlier. To determine whether its gains come from this allocation or
simply from generating more tokens, we force the baselines to continue until
they match MIRA's output-token count on each problem. This recovers
most of the performance gap, but the residual gap
is concentrated on four problems. On these
problems, GPT-5.5 and Codex primarily extend or repair
existing proof directions, whereas MIRA more often
revisits earlier evidence and explores alternatives.
We characterize this behavior using strategic
inertia, the fraction of consecutive semantic
steps within the same research direction. Across
natural-stopping runs using GPT-5.5, mean inertia is
0.958 for direct inference, 0.862 for Codex, and
0.772 for MIRA. MIRA's lower inertia means that it switches between proof
directions more often.
Fig.~\ref{fig:imoproofbench-pb23-ribbon}
illustrates this behavior on Problem 23;
App.~\ref{app:semantic-graphs} provides the
metric definition and additional examples.

\paragraph{Neural-architecture research.}

We introduce HillClimbBench to evaluate MIRA in
open-ended experimental research. Each environment starts
from a task's paper-derived reference architecture
and asks the agent to improve it through code
changes and experiments. We study
two architecture-research tasks; Sec.~\ref{sec:rq3} uses a separate
50-task collection, HillClimbBench-CPU, for policy learning. Each task provides
proxy evaluations that the agent can use to hill-climb during research. After
submission, a held-out gold evaluation scores the final design.

\paragraph{Residual Matrix Transformer.}
\label{sec:rq1-rmt}

The Residual Matrix Transformer (RMT)~\citep{mak_residual_2025} replaces
the usual vector-valued residual stream of the Transformer
architecture~\citep{vaswani_attention_2017} with a matrix that each layer reads
from and writes to. The research problem is to regulate where attention and MLP
updates write into this state without discarding its matrix structure.
HillClimbBench-RMT asks the agent to improve the architecture under that
constraint.
We run matched 24-hour GPT-5.5 campaigns under the three orchestration
conditions. GPT-5.5 stays close to conventional
training and scaling choices, while Codex refines an
RMT-specific normalization family. After more elaborate
routing mechanisms prove ineffective at scale, MIRA shifts
from normalization and cross-axis mixing to local write
control and selects learned per-cell residual-write gates.
MIRA achieves the lowest proxy and
gold loss among the three campaigns
(Fig.~\ref{fig:rq1-hillclimbs}(b)).
App.~\ref{app:rmt-analysis} provides the matched controls, architecture,
initialization, and ablations.

\paragraph{Loop Transformer.}
\label{sec:rq1-loop}

\begin{figure}[t]
  \centering
  \includegraphics[width=\linewidth]
    {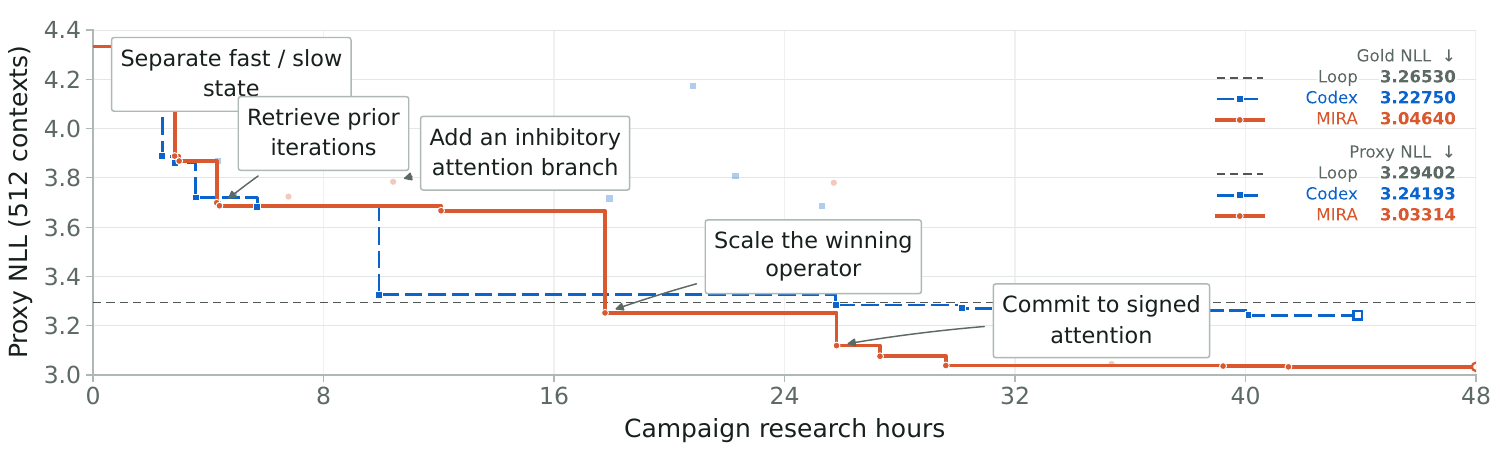}
  \caption{\textbf{MIRA with GPT-6 Astra improves the Loop Transformer by
  adding signed attention and a leaky recurrent update,
  outperforming Codex in architectural hill climbing.} Across matched 48-hour campaigns,
  MIRA progresses from separating fast and slow states and retrieving prior
  iterations to testing an inhibitory attention branch, scaling the successful
  operator, and selecting the combined architecture. The
  resulting model achieves lower proxy and gold negative log-likelihood (NLL)
  than Codex and the reference implementation. Curves track the best proxy NLL observed
  over time; faint markers show all evaluations,
  and hollow markers show where each run ends.
  Callouts identify major changes in MIRA's research direction. The stacked
  legend reports each system's final proxy and gold NLL.}
  \label{fig:loop-hillclimb}
\end{figure}

A Loop Transformer~\citep{geiping_scaling_2025} repeatedly applies a shared
block, increasing effective depth without
introducing distinct parameters at every
iteration. Its central challenge is to revise the recurrent state without
erasing information needed by later applications of the block.
HillClimbBench-Loop asks the agent to improve this information flow.
We compare matched 48-hour MIRA and Codex
campaigns using GPT-6 Astra. MIRA first hypothesizes that repeated core
applications overwrite useful state. When an adaptive retention gate does not
beat constant interpolation with the previous state, it retains the simpler
leaky update. After memory and routing alternatives produce small or
inconsistent gains, MIRA hypothesizes that repeated positive attention may
reinforce information that later recurrences struggle to remove. It arrives at
what it calls \emph{signed attention}, which subtracts one attention read over
shared values from another, and tests it against a matched positive-mixture
control before scaling. This resembles a fixed-coefficient variant of
Differential Attention~\citep{ye_differential_2025} applied in a Loop
Transformer.
MIRA achieves lower NLL than Codex under both proxy and gold evaluation. Both
campaigns also outperform the task reference architecture on the proxy metric.
The selected RMT and Loop designs both regulate updates to persistent model
state: the RMT design scales how strongly attention and MLP branches write to
each matrix cell, while the Loop design allows recurrent attention to suppress
as well as reinforce information and preserves part of the previous state.
Fig.~\ref{fig:loop-hillclimb} shows the
progression of both research campaigns;
App.~\ref{app:loop-analysis} details the matched
controls, operator, initialization, and selected designs.

\begin{figure}[t]
  \centering
  \includegraphics[width=\linewidth]
    {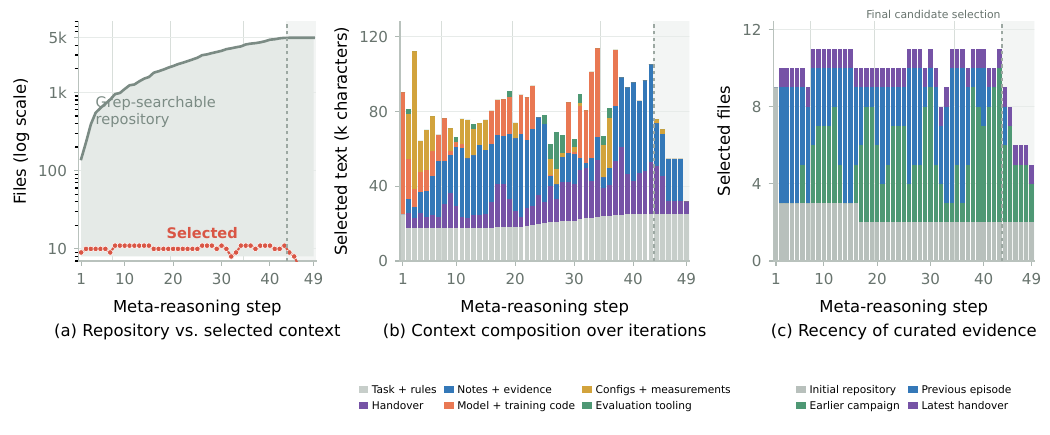}
  \caption{\textbf{MIRA keeps its working context bounded and current as the
  repository grows.}
  \textbf{(a)} Curation bounds the working state as the repository grows.
  \textbf{(b)} Its evidence mix changes with the active research direction.
  \textbf{(c)} Recent artifacts and the latest handover maintain continuity
  across context resets. The dashed line marks final candidate selection.}
  \label{fig:loop-context-curation}
\end{figure}

\paragraph{Context curation.}
Across the Loop campaign, MIRA retains a stable task scaffold while selecting
the notes, code, and measurements relevant to the active experiment. Recent
artifacts and the latest handover preserve continuity across context resets,
while the curated context contracts after final candidate selection
(Fig.~\ref{fig:loop-context-curation}).

These studies show how MIRA reallocates inference as evidence accumulates.
Making each meta-reasoning decision explicit also provides a natural boundary
for forecasting remaining progress and assigning credit from later outcomes.

\subsection{Can We Forecast Remaining Research Value?}
\label{sec:exp-value}
\begin{figure}[t]
  \centering
  \includegraphics[width=\linewidth]
    {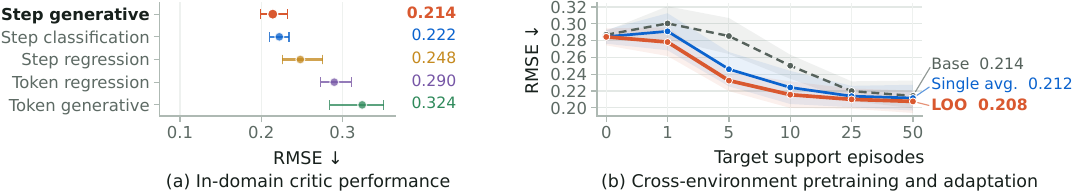}
  \caption{\textbf{Decision-level generative forecasting and cross-environment
  pretraining provide a stronger prior for partial progress.}
  \textbf{(a)} The decision-level generative critic achieves the lowest held-out
  RMSE, and \textbf{(b)} LOO pretraining achieves the lowest RMSE at every
  target-support size, improving zero-shot forecasting and adaptation to
  held-out environments. (95\% CIs from a paired bootstrap over support/query
  split seeds.)}
  \label{fig:critic-transfer-parameterization}
\end{figure}

Using proxy-marginal returns, we compare three decision-level critics---scalar,
categorical-head, and generative---with scalar and generative token-level
baselines. For each environment, we form five support/query splits, each with
task-disjoint sets of 50 support and 50 query episodes. Under matched
conditions, each critic starts from the same Qwen-27B initialization and trains
on all 50 support episodes. We
evaluate it on the query episodes using mean per-episode root mean squared
error (RMSE). For query
episodes $\mathcal Q$, this is
\begin{equation}
    \mathrm{RMSE}
    =
    \frac{1}{|\mathcal Q|}
    \sum_{\tau\in\mathcal Q}
    \sqrt{
        \frac{1}{T_\tau}
        \sum_{t=1}^{T_\tau}
        \left(
            V_\phi(o_t)-G_t
        \right)^2
    }.
    \label{eq:critic-rmse}
\end{equation}
The decision-level generative critic achieves the
lowest mean held-out RMSE among the tested designs,
reducing RMSE by an absolute 0.110
relative to its token-level generative counterpart
(Fig.~\ref{fig:critic-transfer-parameterization}(a)).
We use this design in the subsequent transfer
and policy-learning experiments.

We test whether critic pretraining improves
forecasting in an unseen research environment.
For each target environment, we compare a target-only
model with models initialized by pretraining on one source
environment or on all remaining environments.
The latter is the LOO initialization defined above. We then adapt each critic on
0, 1, 5, 10, 25, or 50
target-environment support episodes and evaluate it on the same 50 held-out
query episodes. This measures how quickly forecasting improves from each
initialization as target-environment data increases.
Across environments, LOO initialization enables better zero-shot forecasting
and faster adaptation than either single-source pretraining or no pretraining.
After adaptation
on five target-environment episodes, it reduces RMSE
by an absolute 0.053 compared with target-only training
(Fig.~\ref{fig:critic-transfer-parameterization}(b)). We use pretrained critics to assign
delayed credit and train MIRA's research-decision
policy.

\subsection{Does Decision-Level Credit Improve the Research Policy?}
\label{sec:exp-policy}

\paragraph{Experimental protocol.}
We evaluate on four environments: BNLearn causal discovery, constructed from
the Bayesian Network Repository~\citep{scutari_bnrepository}; Physics
Discovery~\citep{wiemann_discoverphysics_2026}, SRBench symbolic
regression~\citep{shojaee_llmsrbench_2025}, and HillClimbBench-CPU. The first
three each use 100 tasks.
HillClimbBench-CPU has 50 paper-derived tasks,
each with two independently initialized rollouts.
Qwen-27B is the underlying model for every MIRA agent role. In trained
configurations, the learned checkpoint is used only for action selection and value
forecasting; curation and execution remain fixed. We organize the comparison as an ablation ladder over
supervision, credit granularity, critic parameterization,
and actor--critic co-training. We begin with a
separate actor--critic that uses token-level scalar regression
and gold-binary outcomes. This \emph{Gold binary / Token regression} condition
is our analogue of single-rollout asynchronous methods such as
SAO~\citep{sao}, which typically optimize all rollout tokens. Restricting the
actor loss to the complete sampled MIRA decision response---its reasoning plus work order or
termination decision---tests the same approach for learning a meta-reasoning
policy. Holding the critic
parameterization and granularity fixed, we replace terminal
feedback with marginal improvements in the proxy metric.
Under proxy supervision, i.e., training on proxy returns rather than the
gold-binary outcome, we then move value
forecasting to MIRA's decision boundaries using decision-level scalar
regression (\emph{Step regression}) and replace scalar regression with the
decision-level generative critic (\emph{Step generative}). We evaluate the
generative critic under both supervision
regimes, first with separate policy and critic parameters and
then with MIRA-AC training.

Token regression repeats the decision return target at each
token prefix within the meta-reasoning action; the decision-level regression
and generative critics forecast once per MIRA decision.
In the separate-critic variants, the actor starts
from the base policy and the critic from its pretrained
checkpoint. MIRA-AC initializes the shared actor--critic from
the pretrained critic checkpoint. Accordingly, this comparison
evaluates the combined MIRA-AC design, not parameter sharing
alone. All comparisons hold the MIRA harness, executor,
tools, research horizon, inference budgets, and environment
feedback fixed.
Each environment-specific run makes one pass over a fixed, ordered set of 100
training episodes. The cross-environment run uses 300 episodes, comprising 100
from each of its three actor-training environments. We evaluate each run's
final policy checkpoint using gold evaluation.

\begin{table*}[t]
  \caption{\textbf{Autoresearch gains from training MIRA's meta-reasoning
  policy, measured by gold evaluation.} The primary outcomes are Solved
  (environment-specific successes) and Wins (paired HillClimbBench-CPU wins
  over untrained MIRA); adjacent columns report secondary metrics. $R^2$ is
  sealed-test coefficient of determination, Law qual. combines Physics
  trajectory fit and explanation quality, Arch. qual. is the mean task-specific
  HillClimbBench architecture score, and structural Hamming distance (SHD) gain
  is the normalized closure of the directed-SHD gap; higher is better and all
  are defined in App.~\ref{sec:environments}. Gold binary and Proxy marginal
  denote training feedback, not evaluation. Italics mark Gold-binary group
  bests, bold marks Proxy-marginal group bests, and boxes mark the best overall
  counts. The final row trains one MIRA-AC policy across all
  environments except BNLearn, which is held out from actor training but
  included in critic pretraining.}
  \label{tab:mira-ac-final-summary}
  \centering
  \footnotesize
  \setlength{\tabcolsep}{3.5pt}
  \renewcommand{\arraystretch}{1.08}
  \begin{tabularx}{\textwidth}{@{}cl*{8}{>{\centering\arraybackslash}X}@{}}
    \toprule
    &
    & \multicolumn{2}{c}{\textbf{Symbolic Regression}}
    & \multicolumn{2}{c}{\textbf{Physics Discovery}}
    & \multicolumn{2}{c}{\textbf{HillClimbBench-CPU}}
    & \multicolumn{2}{c}{\textbf{BNLearn}} \\
    \cmidrule(lr){3-4}
    \cmidrule(lr){5-6}
    \cmidrule(lr){7-8}
    \cmidrule(lr){9-10}
    & \textbf{Training variant}
    & Solved & $R^2$
    & Solved & Law qual.
    & Wins & Arch. qual.
    & Solved & SHD gain \\
    \midrule
    & MIRA (untrained)
    & 50 & 0.9864
    & 8 & 0.2118
    & --- & 0.7624
    & 76 & 0.9672 \\
    \midrule
    \multirow{3}{*}{\rotatebox[origin=c]{90}{\shortstack{Gold\\binary}}}
    & Token regression
    & 55 & 0.9957
    & \emph{9} & 0.2327
    & \emph{49} & \emph{0.7594}
    & 79 & 0.9567 \\
    & Step generative
    & 58 & \emph{0.9993}
    & \emph{9} & \emph{0.2733}
    & 43 & 0.7518
    & \emph{82} & \emph{0.9804} \\
    & MIRA-AC
    & \bestoverall{\emph{65}} & 0.9879
    & 6 & 0.1846
    & \emph{49} & 0.7481
    & 56 & 0.9261 \\
    \midrule
    \multirow{6}{*}{\rotatebox[origin=c]{90}{\shortstack{Proxy\\marginal}}}
    & Token regression
    & 55 & 0.9860
    & 6 & 0.2463
    & 49 & 0.7618
    & 63 & 0.9002 \\
    & Step regression
    & 50 & 0.9980
    & 6 & \textbf{0.3043}
    & 46 & 0.7578
    & \bestoverall{\textbf{84}} & 0.9733 \\
    & Step generative
    & 59 & 0.9990
    & 8 & 0.2463
    & 53 & 0.7693
    & 76 & 0.9106 \\
    & MIRA-AC
    & 60 & \textbf{0.9994}
    & \bestoverall{\textbf{10}} & 0.2955
    & \bestoverall{\textbf{58}} & \textbf{0.8054}
    & 48 & 0.9059 \\
    \cmidrule(l){2-10}
    & \multicolumn{9}{c}{\textit{Online actor training across all environments except BNLearn}} \\
    & MIRA-AC
    & \textbf{63} & 0.9869
    & \bestoverall{\textbf{10}} & 0.3028
    & 55 & 0.7853
    & \bestoverall{\textbf{84}} & \textbf{0.9869} \\
    \bottomrule
  \end{tabularx}
\end{table*}

\paragraph{Training with decision-level credit
assignment improves long-horizon research.}
MIRA-AC shows that long-horizon research can be improved by assigning credit
directly to the meta-reasoning decisions that shape a trajectory. Trained from
proxy hill-climbing feedback across multiple autoresearch environments, the
policy achieves higher gold-evaluation point estimates than untrained MIRA in
all four environments. Among the separate actor--critic ablations under
proxy-marginal supervision, the strongest decision-level variant improves the
primary outcome over \emph{Token regression} in every environment. \emph{Step
generative} performs best on Symbolic Regression, Physics Discovery, and
HillClimbBench-CPU, while \emph{Step regression} performs best on BNLearn. MIRA
exposes each meta-reasoning decision, and MIRA-AC applies policy loss to its
complete sampled response while masking the much longer executor trace.
MIRA-AC therefore improves research outcomes without directly optimizing the
much longer executor traces, demonstrating the leverage of targeted credit
assignment at meta-reasoning boundaries.

\paragraph{Jointly learning to forecast partial progress and act from the
same curated research observations produces a stronger policy.}
Ordinarily, the critic affects the policy only through the advantages used in
the actor loss. Unlike \emph{Step generative}, which uses separate actor and
critic adapters, MIRA-AC applies both objectives to the same model with shared
parameters. The critic loss therefore directly trains the policy's
representation to recognize evidence predictive of future progress. Under
proxy-marginal supervision, the environment-specific MIRA-AC policy further
improves the primary outcome on the three environments where
\emph{Step generative} is
strongest. BNLearn instead favors \emph{Step regression}. The critic loss
provides an outcome-supervised auxiliary objective for learning what to
investigate next.

\paragraph{MIRA-AC demonstrates the viability of training a general
meta-reasoning policy.}
Beyond the environment-specific ablations, we train a single MIRA-AC policy from
proxy hill-climbing feedback across Symbolic Regression, Physics Discovery, and
HillClimbBench-CPU. The cross-environment policy matches or exceeds the strongest
separate-adapter primary outcome in every environment. On BNLearn, it matches
\emph{Step regression} without receiving BNLearn actor updates. This experiment
builds on the transferable critic prior established in
Sec.~\ref{sec:exp-value}: MIRA-AC starts from a critic pretrained on all four
environments, including BNLearn, to forecast progress from curated research
observations. Its BNLearn result therefore provides evidence that the meta-reasoning
policy learned from proxy feedback in the other environments transfers to
BNLearn. Together, these results show that MIRA-AC can be scaled to a general
meta-reasoning policy that learns reusable research strategies for deciding
what to investigate across a variety of research domains.

\subsection{What Changes After Meta-Reasoning Policy Training?}
\label{sec:behavior-systems}

Beyond final task outcomes, we analyze how policy training changes research
behavior, which generated tokens receive actor loss, and how context curation
affects rollout input costs.

\paragraph{Analysis of trained meta-reasoning trajectories reveals notable
shifts in research behavior, particularly a greater tendency to target
unresolved uncertainty.}
We use an LLM judge to compare 400 matched trajectory pairs while hiding policy
identities, using a rubric covering hypothesis revision, experiment choice,
evidence use, robustness testing, and stopping decisions.
The judge first evaluates research behavior without seeing final outcomes.
After outcomes are revealed, it assesses which policy's outer-loop behavior
contributed more strongly to the corresponding trajectory's gold outcome.
It favors MIRA-AC in 203 pairs and untrained MIRA in 131, with 66 tied or
unclear; among clear attributions, 60.8\% favor MIRA-AC.

\begin{figure*}[t]
  \centering
  \includegraphics[width=\textwidth]
    {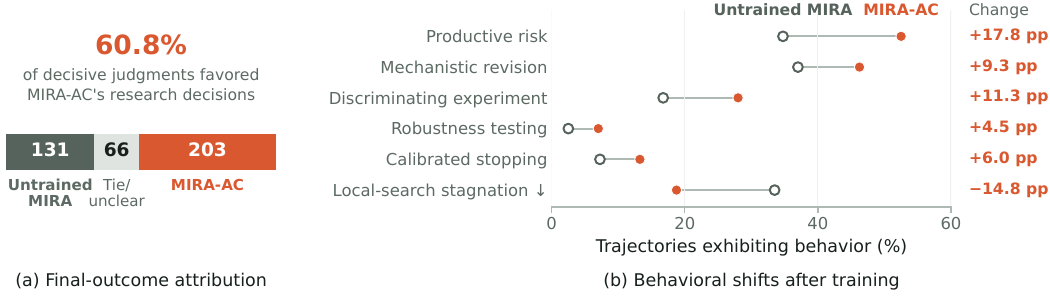}
  \caption{\textbf{MIRA-AC's meta-reasoning is more often judged to make the
  stronger contribution to gold outcomes and to target unresolved uncertainty.}
  Results cover 400 matched pairs across four environments.
  \textbf{(a)} Which outer policy was judged to have contributed more to its
  own gold outcome; 66 comparisons were tied or unclear.
  \textbf{(b)} Share of trajectories assigned each behavior label.
  The first row reports reasonable exploratory bets that improved the gold
  result; the remaining rows report behaviors attributed at least partly to
  the outer policy.
  Each trajectory contributes at most once per row.
  Lower is preferable for local-search stagnation, defined as repeatedly trying
  minor variants of an unproductive idea.}
  \label{fig:mira-ac-behavior}
\end{figure*}

Across environments, the judge more often identifies MIRA-AC trajectories that
revise explanations, test competing ideas, check whether gains persist, and
abandon stalled lines of inquiry.
The largest shifts are an increase in trajectories containing a test that
distinguishes competing ideas, from 16.8\% to 28.0\%, and a decrease in
trajectories repeatedly trying minor variants of an unproductive idea, from
33.5\% to 18.8\% (Fig.~\ref{fig:mira-ac-behavior}).
In Symbolic Regression, this appears as changing the functional form in
response to structured errors rather than continuing to tune coefficients.
In Physics Discovery, it appears as choosing experiments that vary one factor
to distinguish competing physical laws.
In HillClimbBench-CPU, it appears as leaving a plateaued sequence of local
modifications for an architectural change tied to the diagnosed bottleneck.
In BNLearn, which receives no actor updates, the transferred policy targets
specific unresolved edges before expanding to broader graph search.
App.~\ref{app:mira-ac-behavior} provides matched case analyses and
trajectory excerpts, including cases where these changes help or hurt final
performance.

\begin{figure}[t]
  \centering
  \includegraphics[width=\linewidth]
    {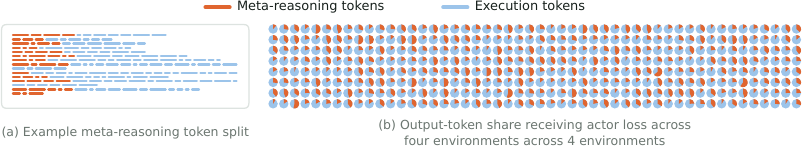}
  \caption{\textbf{MIRA-AC concentrates actor learning on research decisions.}
  Across the four environment-specific training runs (400 episodes), complete
  meta-reasoning decision responses comprise 25.8\% of accounted decision and
  execution output tokens and are the only generated tokens receiving
  actor loss.
  Left: one complete SRBench PO30 episode.
  Right: per-episode token shares across the full training set.}
  \label{fig:mira-ac-token-landscape}
\end{figure}

\paragraph{Actor updates target only the research-decision responses.}
Across the four environment-specific training runs, comprising 400 episodes,
complete sampled meta-reasoning decision responses account for 25.8\% of
accounted decision and execution output tokens and are the only generated tokens that
receive actor loss. Context-curation output is excluded because it was not
fully logged.
Fig.~\ref{fig:mira-ac-token-landscape} shows this separation within one
complete episode and across the full training set.

\begin{figure*}[t]
  \centering
  \includegraphics[width=\textwidth]
    {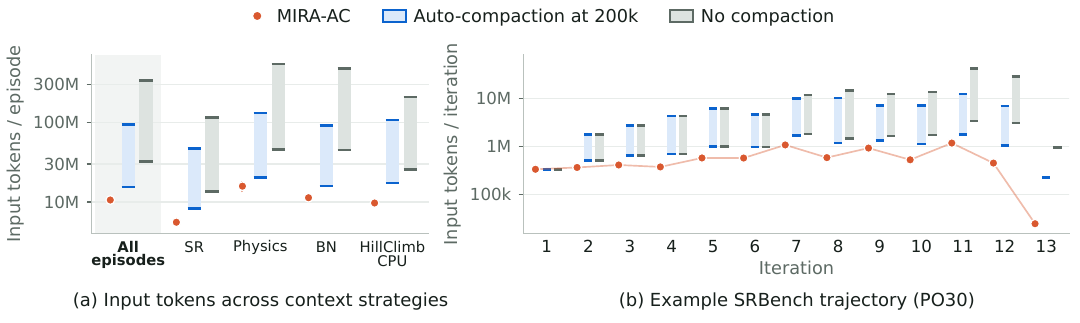}
  \caption{\textbf{Curation keeps active contexts compact and deliberately
  focused as research histories grow.}
  Across 400 recorded MIRA-AC trajectories, we compare accounted input-token
  use with projected replay of prior context either in full or with
  auto-compaction at 200k tokens.
  \textbf{(a)} Mean input tokens per episode.
  \textbf{(b)} Per-iteration input tokens for an example SRBench trajectory.
  Points show recorded MIRA-AC usage; bars span projected replay costs under
  different assumptions about how often retained history is reread. SR denotes
  SRBench, and BN denotes BNLearn.}
  \label{fig:mira-context-reset}
\end{figure*}

\paragraph{Curated context limits repeated-history input.}
At each research decision, MIRA reads selected evidence from the persistent
record instead of replaying the entire preceding trajectory.
Under our accounting, MIRA-AC processes an average of 10.6 million input tokens
per episode across the 400 training episodes.
We compare this recorded cost with fixed-trajectory projections that replay the
same trajectories while carrying prior context either in full or with
auto-compaction at 200k tokens.
Both replay alternatives require more input under this projection as research
history accumulates (Fig.~\ref{fig:mira-context-reset}).
App.~\ref{app:implementation-details} details the accounting and projection
assumptions.

\ifmetapaper
\section{Limitations and Future Work}
\label{sec:limitations}
\label{app:limitations}

MIRA-AC learns only part of the meta-reasoning process.
The policy that generates research decisions is trained, while evidence
curation and execution remain fixed.
This makes learning tractable and provides a clean credit-assignment boundary,
but it also means that the policy can act only on evidence surfaced by the
curator and can improve outcomes only through investigations the executor is
capable of carrying out.
Relevant evidence omitted during curation cannot be recovered by the actor or
critic, and improvements in the executor may change which research decisions
are valuable.
An important next step is therefore to learn curation, allocation, and
possibly execution jointly while preserving decision boundaries
for credit assignment.

The behavioral results raise a broader question: what should a general
meta-reasoner learn about how to conduct research? Policy training shapes not
only task performance, but the model's research temperament---how readily it
abandons an incumbent, gathers more evidence, or commits to a larger change.
In HillClimbBench-CPU, MIRA-AC's willingness to pursue substantial
architectural changes can escape local optima that trap untrained MIRA on some
tasks, yet the same willingness can backfire when preserving and patiently
refining the incumbent is better. The calibration problem recurs elsewhere: a
new experiment is useful while physical laws remain ambiguous but wasteful once
they are resolved; a richer symbolic expression can improve proxy fit while
harming extrapolation; and broad graph search can be inferior to a one-edge
repair. The objective is therefore not maximal exploration or aggressiveness,
but context-sensitive judgment about when to persist, refine, test, pivot, or
stop.

The environments used for policy training cover only a small fraction of the
diversity required to learn this judgment. A general policy will require a
deliberately constructed training distribution spanning distinct research
regimes---exploration and exploitation, bold pivots and incremental
refinement, continued investigation and stopping---so that the model learns
not one research style, but which style and granularity of investigation a
situation warrants. In this work, we demonstrate that decision-level training
improves outcomes, changes research behavior, and transfers to an environment
held out from actor training, providing a roadmap for scaling toward that goal.

We demonstrate that meta-reasoning updates transfer to an environment held out
from actor training: MIRA-AC improves on BNLearn despite receiving no BNLearn
actor updates. The model is, however, exposed to BNLearn during value
pretraining. Future work should examine broader transfer to environments
excluded from both value pretraining and actor training.

Our experiments use a single MIRA configuration with the same Codex-based
inner-loop harness across research environments. This holds the research
interface fixed while varying the task. Training across heterogeneous and
evolving inner-loop scaffolds is a natural next step for testing whether
learned meta-reasoning remains useful as execution capabilities change.

Our learning experiments also depend on proxy feedback.
Dense proxy signals make repeated autoresearch training feasible, but a proxy
need not rank research directions in the same way as the final gold
evaluation.
Our results show that proxy-trained policies can improve gold-evaluated
outcomes in the environments studied here, but they do not establish when such
transfer should be expected or how robust learning is to systematically
misaligned proxies.
Future work should characterize the degree of proxy--gold alignment required
for stable policy improvement and develop methods that can learn from sparse
gold feedback without giving up the sample efficiency of intermediate signals.

The behavioral analysis should likewise be interpreted as descriptive.
Its conclusions rely on an LLM judge assigning process labels and comparing
matched trajectories.
These annotations provide evidence that training changes observable research
behavior, but they do not establish that any individual behavior---such as
testing competing hypotheses, abandoning stalled directions, or making larger
pivots---causes the associated performance difference.
More direct interventions on research strategy, together with human or
task-specific behavioral evaluation, would be needed to identify which learned
behaviors are causally useful and under which research regimes.

Finally, the current optimization results are limited by scale.
We train LoRA adapters rather than all model parameters, and our asynchronous
procedure is evaluated only at investigation durations supported by the present
infrastructure.
We have not tested episodes in which individual investigations last hours or
days and a single trajectory spans many more actor and critic versions.
At those timescales, delayed and heterogeneous feedback, policy staleness, and
interference between actor and critic updates to shared parameters may become
substantially more important.
Scaling to full-parameter training and longer-lived research episodes will
therefore require studying update scheduling, value calibration under changing
policies, and stable joint optimization over much larger temporal gaps.

More broadly, MIRA-AC assumes that research can be decomposed into meaningful
decision boundaries at which accumulated evidence is summarized, future value
can be estimated, and credit can be assigned.
The environments studied here support that decomposition.
Whether the same abstraction remains effective for less structured research,
where useful investigations overlap, branch, or evolve continuously, is an
open question.
Testing that boundary is important for determining how far decision-level
meta-reasoning can scale toward more general autonomous research.

\section{Conclusion}

This work asks whether a research agent can learn what to investigate next
from the downstream consequences of its previous research decisions.
Across the settings studied here, our results support a positive answer.
MIRA represents long-horizon research as a sequence of explicit
meta-reasoning decisions over a persistent research record, treating the
potentially much longer execution of each investigation as part of an
environment transition.
This decomposition improves long-horizon inference even before policy
training and provides a natural boundary at which remaining research value
can be forecast and delayed outcomes assigned to earlier decisions.

Building on this representation, MIRA-AC jointly trains one model to forecast
remaining return from a curated research observation and choose how to allocate
research effort next.
Decision-level value forecasting outperforms token-level alternatives, and
decision-level policy learning improves gold-evaluated research outcomes,
including when trained from proxy feedback, without directly optimizing the
longer executor trajectories.
We also find initial evidence that actor updates learned across research
environments can transfer to a held-out actor-training environment when
initialized from a cross-environment value model.

The broader implication is that research strategy need not remain an
implicit byproduct of long-horizon generation.
Once decisions about what to investigate are represented explicitly and
linked to their downstream consequences, they become objects that can be
valued, credited, and improved through experience.
This suggests a path toward research agents that learn not only to execute
individual investigations more effectively, but to improve the strategy by
which they decide what research is worth doing next.

\else
\section{Conclusion and Limitations}

Long-horizon autoresearch requires allocating inference as evidence
accumulates, not merely carrying out investigations.
MIRA exposes these choices as meta-reasoning decisions over curated partial
states, separating what to investigate from execution. Through this boundary,
MIRA-AC jointly trains one model to choose what to investigate and forecast
progress, enabling decision-level credit assignment. Targeting
policy updates to these decisions can improve long-horizon research even from
imperfect proxy feedback without
directly optimizing the much longer execution trajectories. Our method assumes
the model can curate a sufficiently representative view of accumulated
research; MIRA-AC trains the core decision policy, not curation, which remains an
important direction for future work. We also hold the
MIRA configuration and Codex-based inner-loop harness fixed, leaving
cross-harness adaptation untested. Proxy hill-climbing signals may diverge
from gold evaluation. Finally, we train only LoRA adapters and have not
tested asynchronous optimization on investigations lasting hours or days.
Further limitations are in App.~\ref{app:limitations}, with related work in
App.~\ref{sec:related}. More broadly, MIRA-AC makes the strategy of conducting
research itself an object of learning.
\fi

\ifmetapaper
\else
\section*{AI Usage Statement}

This work studies how to train language-model agents to conduct research and,
in some settings, improve other language models. The task-specific hypotheses,
experiments, and results are therefore generated entirely by the agents being
trained and evaluated. This does not extend to the paper's broader research, methods, claims, or
conclusions, for which the authors take responsibility. Language models are
also used as judges for qualitative analysis, including Physics Discovery
explanation scoring, semantic exploration-graph annotation, and behavioral
trajectory comparisons.
\fi

\bibliography{paper}

@article{funsearch,
  title   = {Mathematical discoveries from program search with large language models},
  author  = {Romera-Paredes, Bernardino and Barekatain, Mohammadamin and Novikov, Alexander and Balog, Matej and Kumar, M. Pawan and Dupont, Emilien and Ruiz, Francisco J. R. and Ellenberg, Jordan S. and Wang, Pengming and Fawzi, Omar and Kohli, Pushmeet and Fawzi, Alhussein},
  year    = 2024,
  journal = {Nature},
  volume  = 625,
  number  = 7995,
  pages   = {468--475},
  doi     = {10.1038/s41586-023-06924-6},
  url     = {https://doi.org/10.1038/s41586-023-06924-6}
}

@misc{scutari_bnrepository,
  title  = {Bayesian Network Repository},
  author = {Scutari, Marco},
  year   = {n.d.},
  url    = {https://www.bnlearn.com/bnrepository/}
}

@inproceedings{shojaee_llmsrbench_2025,
  title   = {{LLM-SRBench}: A New Benchmark for Scientific Equation Discovery with Large Language Models},
  author  = {Shojaee, Parshin and Nguyen, Ngoc-Hieu and Meidani, Kazem and Farimani, Amir Barati and Doan, Khoa D. and Reddy, Chandan K.},
  year    = 2025,
  booktitle = {Proceedings of the 42nd International Conference on Machine Learning},
  note    = {Oral presentation; arXiv:2504.10415},
  url     = {https://arxiv.org/abs/2504.10415}
}

@misc{ippo,
  title   = {Iterative {GRPO}: Batch-Online Multi-Turn {RL} via Single-Turn {RLHF}},
  author  = {Jiang, Daniel R. and Samanta, Ankur and Yang, Yukai and Bhandari, Jalaj and Munos, R{\'e}mi and Lu, Tyler},
  year    = 2025,
  note    = {arXiv:2511.21638},
  url     = {https://arxiv.org/abs/2511.21638}
}

@misc{sao,
  title  = {Single-Rollout Asynchronous Optimization for Agentic Reinforcement Learning},
  author = {Hou, Zhenyu and Li, Yujiang and Tang, Jie and Dong, Yuxiao},
  year   = {2026},
  note   = {arXiv:2607.07508},
  url    = {https://arxiv.org/abs/2607.07508}
}

@inproceedings{chan_mle-bench_2024,
	title = {{MLE}-bench: {Evaluating} {Machine} {Learning} {Agents} on {Machine} {Learning} {Engineering}},
	shorttitle = {{MLE}-bench},
	url = {https://openreview.net/forum?id=6s5uXNWGIh},
	language = {en},
	urldate = {2026-06-22},
	author = {Chan, Jun Shern and Chowdhury, Neil and Jaffe, Oliver and Aung, James and Sherburn, Dane and Mays, Evan and Starace, Giulio and Liu, Kevin and Maksin, Leon and Patwardhan, Tejal and Weng, Lilian and M{\k{a}}dry, Aleksander},
	booktitle = {International Conference on Learning Representations},
	year = {2025},
	note = {arXiv:2410.07095},
}

@inproceedings{huang_mlagentbench_2024,
	title = {{MLAgentBench}: {Evaluating} {Language} {Agents} on {Machine} {Learning} {Experimentation}},
	shorttitle = {{MLAgentBench}},
	url = {https://proceedings.mlr.press/v235/huang24y.html},
	doi = {10.48550/arXiv.2310.03302},
	urldate = {2026-06-22},
	author = {Huang, Qian and Vora, Jian and Liang, Percy and Leskovec, Jure},
	booktitle = {Proceedings of the 41st International Conference on Machine Learning},
	series = {Proceedings of Machine Learning Research},
	volume = {235},
	publisher = {PMLR},
	year = {2024},
	note = {arXiv:2310.03302},
}

@misc{qiang_mle-dojo_2025,
	title = {{MLE}-{Dojo}: {Interactive} {Environments} for {Empowering} {LLM} {Agents} in {Machine} {Learning} {Engineering}},
	shorttitle = {{MLE}-{Dojo}},
	url = {https://arxiv.org/abs/2505.07782},
	doi = {10.48550/arXiv.2505.07782},
	urldate = {2026-06-22},
	publisher = {arXiv},
	author = {Qiang, Rushi and Zhuang, Yuchen and Li, Yinghao and K, Dingu Sagar V. and Zhang, Rongzhi and Li, Changhao and Wong, Ian Shu-Hei and Yang, Sherry and Liang, Percy and Zhang, Chao and Dai, Bo},
	month = may,
	year = {2025},
	note = {arXiv:2505.07782},
}

@inproceedings{chen_scienceagentbench_2025,
	title = {{ScienceAgentBench}: {Toward} {Rigorous} {Assessment} of {Language} {Agents} for {Data}-{Driven} {Scientific} {Discovery}},
	shorttitle = {{ScienceAgentBench}},
	url = {https://arxiv.org/abs/2410.05080},
	doi = {10.48550/arXiv.2410.05080},
	urldate = {2026-06-22},
	author = {Chen, Ziru and Chen, Shijie and Ning, Yuting and Zhang, Qianheng and Wang, Boshi and Yu, Botao and Li, Yifei and Liao, Zeyi and Wei, Chen and Lu, Zitong and Dey, Vishal and Xue, Mingyi and Baker, Frazier N. and Burns, Benjamin and Adu-Ampratwum, Daniel and Huang, Xuhui and Ning, Xia and Gao, Song and Su, Yu and Sun, Huan},
	booktitle = {International Conference on Learning Representations},
	year = {2025},
	note = {arXiv:2410.05080},
}

@misc{starace_paperbench_2025,
	title = {{PaperBench}: {Evaluating} {AI}'s {Ability} to {Replicate} {AI} {Research}},
	shorttitle = {{PaperBench}},
	url = {https://arxiv.org/abs/2504.01848},
	doi = {10.48550/arXiv.2504.01848},
	urldate = {2026-06-22},
	publisher = {arXiv},
	author = {Starace, Giulio and Jaffe, Oliver and Sherburn, Dane and Aung, James and Chan, Jun Shern and Maksin, Leon and Dias, Rachel and Mays, Evan and Kinsella, Benjamin and Thompson, Wyatt and Heidecke, Johannes and Glaese, Amelia and Patwardhan, Tejal},
	month = apr,
	year = {2025},
	note = {arXiv:2504.01848},
}

@misc{wang_frontierscience_2026,
	title = {{FrontierScience}: {Evaluating} {AI}'s {Ability} to {Perform} {Expert}-{Level} {Scientific} {Tasks}},
	shorttitle = {{FrontierScience}},
	url = {https://arxiv.org/abs/2601.21165},
	doi = {10.48550/arXiv.2601.21165},
	urldate = {2026-06-22},
	publisher = {arXiv},
	author = {Wang, Miles and Lin, Robi and Hu, Kat and Jiao, Joy and Chowdhury, Neil and Chang, Ethan and Patwardhan, Tejal},
	month = jan,
	year = {2026},
	note = {arXiv:2601.21165},
}

@inproceedings{wijk_re-bench_2025,
	title = {{RE}-{Bench}: {Evaluating} frontier {AI} {R}\&{D} capabilities of language model agents against human experts},
	shorttitle = {{RE}-{Bench}},
	url = {https://proceedings.mlr.press/v267/wijk25a.html},
	doi = {10.48550/arXiv.2411.15114},
	urldate = {2026-06-22},
	author = {Wijk, Hjalmar and Lin, Tao and Becker, Joel and Jawhar, Sami and Parikh, Neev and Broadley, Thomas and Chan, Lawrence and Chen, Michael and Clymer, Josh and Dhyani, Jai and Ericheva, Elena and Garcia, Katharyn and Goodrich, Brian and Jurkovic, Nikola and Karnofsky, Holden and Kinniment, Megan and Lajko, Aron and Nix, Seraphina and Sato, Lucas and Saunders, William and Taran, Maksym and West, Ben and Barnes, Elizabeth},
	booktitle = {Proceedings of the 42nd International Conference on Machine Learning},
	series = {Proceedings of Machine Learning Research},
	volume = {267},
	publisher = {PMLR},
	year = {2025},
	note = {arXiv:2411.15114},
}

@misc{rein_hcast_2025,
	title = {{HCAST}: {Human}-{Calibrated} {Autonomy} {Software} {Tasks}},
	shorttitle = {{HCAST}},
	url = {https://arxiv.org/abs/2503.17354},
	doi = {10.48550/arXiv.2503.17354},
	urldate = {2026-06-22},
	publisher = {arXiv},
	author = {Rein, David and Becker, Joel and Deng, Amy and Nix, Seraphina and Canal, Chris and O'Connel, Daniel and Arnott, Pip and Bloom, Ryan and Broadley, Thomas and Garcia, Katharyn and Goodrich, Brian and Hasin, Max and Jawhar, Sami and Kinniment, Megan and Kwa, Thomas and Lajko, Aron and Rush, Nate and Sato, Lucas Jun Koba and Arx, Sydney Von and West, Ben and Chan, Lawrence and Barnes, Elizabeth},
	month = mar,
	year = {2025},
	note = {arXiv:2503.17354},
}

@misc{press_algotune_2025,
	title = {{AlgoTune}: {Can} {Language} {Models} {Speed} {Up} {General}-{Purpose} {Numerical} {Programs}?},
	shorttitle = {{AlgoTune}},
	url = {https://arxiv.org/abs/2507.15887},
	doi = {10.48550/arXiv.2507.15887},
	urldate = {2026-06-22},
	publisher = {arXiv},
	author = {Press, Ori and Amos, Brandon and Zhao, Haoyu and Wu, Yikai and Ainsworth, Samuel K. and Krupke, Dominik and Kidger, Patrick and Sajed, Touqir and Stellato, Bartolomeo and Park, Jisun and Bosch, Nathanael and Meril, Eli and Steppi, Albert and Zharmagambetov, Arman and Zhang, Fangzhao and Perez-Pineiro, David and Mercurio, Alberto and Zhan, Ni and Abramovich, Talor and Lieret, Kilian and Zhang, Hanlin and Huang, Shirley and Bethge, Matthias and Press, Ofir},
	month = oct,
	year = {2025},
	note = {arXiv:2507.15887},
}

@misc{zhao_automated_2025,
	title = {The {Automated} {LLM} {Speedrunning} {Benchmark}: {Reproducing} {NanoGPT} {Improvements}},
	shorttitle = {The {Automated} {LLM} {Speedrunning} {Benchmark}},
	url = {https://arxiv.org/abs/2506.22419},
	doi = {10.48550/arXiv.2506.22419},
	urldate = {2026-06-22},
	publisher = {arXiv},
	author = {Zhao, Bingchen and Magka, Despoina and Jiang, Minqi and Li, Xian and Raileanu, Roberta and Shavrina, Tatiana and Gagnon-Audet, Jean-Christophe and Niu, Kelvin and Sodhani, Shagun and Shvartsman, Michael and Lupu, Andrei and Lupidi, Alisia and Toledo, Edan and Hambardzumyan, Karen and Josifoski, Martin and Foster, Thomas and Cipolina-Kun, Lucia and Charnalia, Abhishek and Dunfield, Derek and Miller, Alexander H. and Aodha, Oisin Mac and Foerster, Jakob and Bachrach, Yoram},
	month = jun,
	year = {2025},
	note = {arXiv:2506.22419},
}

@misc{rank_posttrainbench_2026,
	title = {{PostTrainBench}: {Can} {LLM} {Agents} {Automate} {LLM} {Post}-{Training}?},
	shorttitle = {{PostTrainBench}},
	url = {https://arxiv.org/abs/2603.08640},
	doi = {10.48550/arXiv.2603.08640},
	urldate = {2026-06-22},
	publisher = {arXiv},
	author = {Rank, Ben and Bhatnagar, Hardik and Prabhu, Ameya and Eisenberg, Shira and Nguyen, Karina and Bethge, Matthias and Andriushchenko, Maksym},
	month = mar,
	year = {2026},
	note = {arXiv:2603.08640},
}

@misc{chi_frontier-eng_2026,
	title = {Frontier-{Eng}: {Benchmarking} {Self}-{Evolving} {Agents} on {Real}-{World} {Engineering} {Tasks} with {Generative} {Optimization}},
	shorttitle = {Frontier-{Eng}},
	url = {https://arxiv.org/abs/2604.12290},
	doi = {10.48550/arXiv.2604.12290},
	urldate = {2026-06-22},
	publisher = {arXiv},
	author = {Chi, Yizhe and Hong, Deyao and Jiang, Dapeng and Luo, Tianwei and Yang, Kaisen and Zhang, Boshi and Cao, Zhe and Fan, Xiaoyan and He, Bingxiang and Hao, Han and Jin, Weiyang and Lei, Dianqiao and Liu, Qingle and Qian, Houde and Wang, Bowen and Wang, Situ and Zheng, Youjie and Zhou, Yifan and Xiao, Calvin and Cai, Eren and Na, Qinhuai},
	month = apr,
	year = {2026},
	note = {arXiv:2604.12290},
}

@misc{dennis_compiling_2026,
	title = {Compiling {Agentic} {Workflows} into {LLM} {Weights}: {Near}-{Frontier} {Quality} at {Two} {Orders} of {Magnitude} {Less} {Cost}},
	shorttitle = {Compiling {Agentic} {Workflows} into {LLM} {Weights}},
	url = {https://arxiv.org/abs/2605.22502},
	doi = {10.48550/arXiv.2605.22502},
	urldate = {2026-06-22},
	publisher = {arXiv},
	author = {Dennis, Simon and Patil, Rivaan and Shabahang, Kevin and Guo, Hao},
	month = may,
	year = {2026},
	note = {arXiv:2605.22502},
}

@misc{lin_harness_2026,
	title = {Harness {Updating} {Is} {Not} {Harness} {Benefit}: {Disentangling} {Evolution} {Capabilities} in {Self}-{Evolving} {LLM} {Agents}},
	shorttitle = {Harness {Updating} {Is} {Not} {Harness} {Benefit}},
	url = {https://arxiv.org/abs/2605.30621},
	doi = {10.48550/arXiv.2605.30621},
	urldate = {2026-06-22},
	publisher = {arXiv},
	author = {Lin, Minhua and Wu, Juncheng and Wang, Zijun and Shi, Zhan and Sang, Yisi and He, Bing and Liu, Zewen and Wei, Tianxin and Wu, Zongyu and Zhang, Zhiwei and Wang, Dakuo and Zhang, Xiang and Dumoulin, Benoit and Xie, Cihang and Zhou, Yuyin and Wang, Suhang and Lu, Hanqing},
	month = may,
	year = {2026},
	note = {arXiv:2605.30621},
}

@misc{wang_skillgrad_2026,
	title = {{SkillGrad}: {Optimizing} {Agent} {Skills} {Like} {Gradient} {Descent}},
	shorttitle = {{SkillGrad}},
	url = {https://arxiv.org/abs/2605.27760},
	doi = {10.48550/arXiv.2605.27760},
	urldate = {2026-06-22},
	publisher = {arXiv},
	author = {Wang, Hanyu and Lan, Yifan and Cao, Bochuan and Lin, Lu and Chen, Jinghui},
	month = may,
	year = {2026},
	note = {arXiv:2605.27760},
}

@misc{yang_skillopt_2026,
	title = {{SkillOpt}: {Executive} {Strategy} for {Self}-{Evolving} {Agent} {Skills}},
	shorttitle = {{SkillOpt}},
	url = {https://arxiv.org/abs/2605.23904},
	doi = {10.48550/arXiv.2605.23904},
	urldate = {2026-06-22},
	publisher = {arXiv},
	author = {Yang, Yifan and Gong, Ziyang and Huang, Weiquan and Yang, Qihao and Zhou, Ziwei and Huang, Zisu and Li, Yan and Gao, Xuemei and Dai, Qi and Liu, Bei and Qiu, Kai and Yang, Yuqing and Chen, Dongdong and Yang, Xue and Luo, Chong},
	month = may,
	year = {2026},
	note = {arXiv:2605.23904},
}

@misc{cheng_llf-bench_2023,
	title = {{LLF}-{Bench}: {Benchmark} for {Interactive} {Learning} from {Language} {Feedback}},
	shorttitle = {{LLF}-{Bench}},
	url = {https://arxiv.org/abs/2312.06853},
	doi = {10.48550/arXiv.2312.06853},
	urldate = {2026-06-22},
	publisher = {arXiv},
	author = {Cheng, Ching-An and Kolobov, Andrey and Misra, Dipendra and Nie, Allen and Swaminathan, Adith},
	month = dec,
	year = {2023},
	note = {arXiv:2312.06853},
}

@inproceedings{xu_formalizing_2026,
	title = {Formalizing {Learning} from {Language} {Feedback} with {Provable} {Guarantees}},
	url = {https://arxiv.org/abs/2506.10341},
	doi = {10.48550/arXiv.2506.10341},
	urldate = {2026-06-22},
	author = {Xu, Wanqiao and Nie, Allen and Zheng, Ruijie and Modi, Aditya and Swaminathan, Adith and Cheng, Ching-An},
	booktitle = {Proceedings of the 43rd International Conference on Machine Learning},
	year = {2026},
	note = {arXiv:2506.10341},
}

@inproceedings{ji_tree_2026,
	title = {Tree {Search} for {LLM} {Agent} {Reinforcement} {Learning}},
	url = {https://arxiv.org/abs/2509.21240},
	doi = {10.48550/arXiv.2509.21240},
	urldate = {2026-06-22},
	author = {Ji, Yuxiang and Ma, Ziyu and Wang, Yong and Chen, Guanhua and Chu, Xiangxiang and Wu, Liaoni},
	booktitle = {International Conference on Learning Representations},
	year = {2026},
	note = {arXiv:2509.21240},
}

@inproceedings{zhou_language_2024,
	title = {Language {Agent} {Tree} {Search} {Unifies} {Reasoning} {Acting} and {Planning} in {Language} {Models}},
	url = {https://arxiv.org/abs/2310.04406},
	doi = {10.48550/arXiv.2310.04406},
	urldate = {2026-06-22},
	author = {Zhou, Andy and Yan, Kai and Shlapentokh-Rothman, Michal and Wang, Haohan and Wang, Yu-Xiong},
	booktitle = {Proceedings of the 41st International Conference on Machine Learning},
	year = {2024},
	note = {arXiv:2310.04406},
}

@inproceedings{goldie_procedural_2026,
	title = {{DiscoGen}: {Procedural} {Generation} of {Algorithm} {Discovery} {Tasks} in {Machine} {Learning}},
	url = {https://arxiv.org/abs/2603.17863},
	doi = {10.48550/arXiv.2603.17863},
	urldate = {2026-06-22},
	author = {Goldie, Alexander D. and Wang, Zilin and Hayler, Adrian and Nathani, Deepak and Toledo, Edan and Thampiratwong, Ken and Kalisz, Aleksandra and Beukman, Michael and Erlebach, Hannah and Letcher, Alistair and Reddy, Shashank and Wibault, Clarisse and Wolf, Theo and O'Neill, Charles and Berdica, Uljad and Roberts, Nicholas and Rahmani, Saeed and Raileanu, Roberta and Whiteson, Shimon and Foerster, Jakob N.},
	booktitle = {Proceedings of the 43rd International Conference on Machine Learning},
	year = {2026},
	note = {arXiv:2603.17863},
}

@misc{zou_fml-bench_2026,
	title = {{FML}-bench: {A} {Controlled} {Study} of {AI} {Research} {Agent} {Strategies} from the {Perspective} of {Search} {Dynamics}},
	shorttitle = {{FML}-bench},
	url = {https://arxiv.org/abs/2605.17373},
	doi = {10.48550/arXiv.2605.17373},
	urldate = {2026-06-22},
	publisher = {arXiv},
	author = {Zou, Qiran and Lam, Hou Hei and Zhao, Wenhao and Chen, Tingting and Tang, Yiming and Yu, Samson and Zhu, Yingtao and Anumasa, Srinivas and Zhang, Zufeng and Zhang, Tianyi and Liu, Chang and Jiang, Zhengyao and Goyal, Anirudh and Liu, Dianbo},
	month = may,
	year = {2026},
	note = {arXiv:2605.17373},
}

@misc{song_r1-searcher_2025,
	title = {R1-{Searcher}: {Incentivizing} the {Search} {Capability} in {LLMs} via {Reinforcement} {Learning}},
	shorttitle = {R1-{Searcher}},
	url = {https://arxiv.org/abs/2503.05592},
	doi = {10.48550/arXiv.2503.05592},
	urldate = {2026-06-22},
	publisher = {arXiv},
	author = {Song, Huatong and Jiang, Jinhao and Min, Yingqian and Chen, Jie and Chen, Zhipeng and Zhao, Wayne Xin and Fang, Lei and Wen, Ji-Rong},
	month = mar,
	year = {2025},
	note = {arXiv:2503.05592},
}

@misc{wang_stepsearch_2025,
	title = {{StepSearch}: {Igniting} {LLMs} {Search} {Ability} via {Step}-{Wise} {Proximal} {Policy} {Optimization}},
	shorttitle = {{StepSearch}},
	url = {https://arxiv.org/abs/2505.15107},
	doi = {10.48550/arXiv.2505.15107},
	urldate = {2026-06-22},
	publisher = {arXiv},
	author = {Wang, Ziliang and Zheng, Xuhui and An, Kang and Ouyang, Cijun and Cai, Jialu and Wang, Yuhang and Wu, Yichao},
	month = may,
	year = {2025},
	note = {arXiv:2505.15107},
}

@misc{szot_expanding_2026,
	title = {Expanding {LLM} {Agent} {Boundaries} with {Strategy}-{Guided} {Exploration}},
	url = {https://arxiv.org/abs/2603.02045},
	doi = {10.48550/arXiv.2603.02045},
	urldate = {2026-06-22},
	publisher = {arXiv},
	author = {Szot, Andrew and Kirchhof, Michael and Attia, Omar and Toshev, Alexander},
	month = mar,
	year = {2026},
	note = {arXiv:2603.02045},
}

@misc{du_datamaster_2026,
	title = {{DataMaster}: {Data}-{Centric} {Autonomous} {AI} {Research}},
	shorttitle = {{DataMaster}},
	url = {https://arxiv.org/abs/2605.10906},
	doi = {10.48550/arXiv.2605.10906},
	urldate = {2026-06-22},
	publisher = {arXiv},
	author = {Du, Yaxin and Yang, Xiyuan and Zhou, Zhifan and Liu, Wanxu and Lei, Zixing and Chen, Zimeng and Liu, Fenyi and Wu, Haotian and Cai, Yuzhu and Liu, Zexi and Zhu, Xinyu and Wang, WenHao and Zhang, Linfeng and Qian, Chen and Chen, Siheng},
	month = may,
	year = {2026},
	note = {arXiv:2605.10906},
}

@inproceedings{chen_mars_2026,
	title = {{MARS}: {Modular} {Agent} with {Reflective} {Search} for {Automated} {AI} {Research}},
	shorttitle = {{MARS}},
	url = {https://arxiv.org/abs/2602.02660},
	doi = {10.48550/arXiv.2602.02660},
	urldate = {2026-06-22},
	author = {Chen, Jiefeng and Mishra, Bhavana Dalvi and Nam, Jaehyun and Meng, Rui and Pfister, Tomas and Yoon, Jinsung},
	booktitle = {Proceedings of the 43rd International Conference on Machine Learning},
	year = {2026},
	note = {arXiv:2602.02660},
}

@misc{wan_loongflow_2025,
	title = {{LoongFlow}: {Directed} {Evolutionary} {Search} via a {Cognitive} {Plan}-{Execute}-{Summarize} {Paradigm}},
	shorttitle = {{LoongFlow}},
	url = {https://arxiv.org/abs/2512.24077},
	doi = {10.48550/arXiv.2512.24077},
	urldate = {2026-06-22},
	publisher = {arXiv},
	author = {Wan, Chunhui and Dai, Xunan and Wang, Zhuo and Li, Minglei and Wang, Yanpeng and Mao, Yinan and Lan, Yu and Xiao, Zhiwen},
	month = dec,
	year = {2025},
	note = {arXiv:2512.24077},
}

@misc{chai_scimaster_2025,
	title = {{SciMaster}: {Towards} {General}-{Purpose} {Scientific} {AI} {Agents}, {Part} {I}. {X}-{Master} as {Foundation}: {Can} {We} {Lead} on {Humanity}'s {Last} {Exam}?},
	shorttitle = {{SciMaster}},
	url = {https://arxiv.org/abs/2507.05241},
	doi = {10.48550/arXiv.2507.05241},
	urldate = {2026-06-22},
	publisher = {arXiv},
	author = {Chai, Jingyi and Tang, Shuo and Ye, Rui and Du, Yuwen and Zhu, Xinyu and Zhou, Mengcheng and Wang, Yanfeng and E, Weinan and Zhang, Yuzhi and Zhang, Linfeng and Chen, Siheng},
	month = jul,
	year = {2025},
	note = {arXiv:2507.05241},
}

@misc{novikov_alphaevolve_2025,
	title = {{AlphaEvolve}: {A} coding agent for scientific and algorithmic discovery},
	shorttitle = {{AlphaEvolve}},
	url = {https://arxiv.org/abs/2506.13131},
	doi = {10.48550/arXiv.2506.13131},
	urldate = {2026-06-22},
	publisher = {arXiv},
	author = {Novikov, Alexander and V{\~u}, Ng{\^a}n and Eisenberger, Marvin and Dupont, Emilien and Huang, Po-Sen and Wagner, Adam Zsolt and Shirobokov, Sergey and Kozlovskii, Borislav and Ruiz, Francisco J. R. and Mehrabian, Abbas and Kumar, M. Pawan and See, Abigail and Chaudhuri, Swarat and Holland, George and Davies, Alex and Nowozin, Sebastian and Kohli, Pushmeet and Balog, Matej},
	month = jun,
	year = {2025},
	note = {arXiv:2506.13131},
}

@article{stokes_deep_2020,
	title = {A Deep Learning Approach to Antibiotic Discovery},
	author = {Stokes, Jonathan M. and Yang, Kevin and Swanson, Kyle and Jin, Wengong and Cubillos-Ruiz, Andres and Donghia, Nina M. and MacNair, Craig R. and French, Shawn and Carfrae, Lindsey A. and Bloom-Ackermann, Zohar and Tran, Victoria M. and Chiappino-Pepe, Anush and Badran, Ahmed H. and Andrews, Ian W. and Chory, Emma J. and Church, George M. and Brown, Eric D. and Jaakkola, Tommi S. and Barzilay, Regina and Collins, James J.},
	journal = {Cell},
	volume = {180},
	number = {4},
	pages = {688--702.e13},
	year = {2020},
	doi = {10.1016/j.cell.2020.01.021},
	url = {https://doi.org/10.1016/j.cell.2020.01.021},
}

@misc{yang_rd-agent_2025,
	title = {R\&{D}-{Agent}: {An} {LLM}-{Agent} {Framework} {Towards} {Autonomous} {Data} {Science}},
	shorttitle = {R\&{D}-{Agent}},
	url = {https://arxiv.org/abs/2505.14738},
	doi = {10.48550/arXiv.2505.14738},
	urldate = {2026-06-22},
	publisher = {arXiv},
	author = {Yang, Xu and Yang, Xiao and Fang, Shikai and Zhang, Yifei and Wang, Jian and Xian, Bowen and Li, Qizheng and Li, Jingyuan and Xu, Minrui and Li, Yuante and Pan, Haoran and Zhang, Yuge and Liu, Weiqing and Shen, Yelong and Chen, Weizhu and Bian, Jiang},
	month = oct,
	year = {2025},
	note = {arXiv:2505.14738},
}

@misc{jiang_aide_2025,
	title = {{AIDE}: {AI}-{Driven} {Exploration} in the {Space} of {Code}},
	shorttitle = {{AIDE}},
	url = {https://arxiv.org/abs/2502.13138},
	doi = {10.48550/arXiv.2502.13138},
	urldate = {2026-06-22},
	publisher = {arXiv},
	author = {Jiang, Zhengyao and Schmidt, Dominik and Srikanth, Dhruv and Xu, Dixing and Kaplan, Ian and Jacenko, Deniss and Wu, Yuxiang},
	month = feb,
	year = {2025},
	note = {arXiv:2502.13138},
}

@misc{yamada_ai_2025,
	title = {The {AI} {Scientist}-v2: {Workshop}-{Level} {Automated} {Scientific} {Discovery} via {Agentic} {Tree} {Search}},
	shorttitle = {The {AI} {Scientist}-v2},
	url = {https://arxiv.org/abs/2504.08066},
	doi = {10.48550/arXiv.2504.08066},
	urldate = {2026-06-22},
	publisher = {arXiv},
	author = {Yamada, Yutaro and Lange, Robert Tjarko and Lu, Cong and Hu, Shengran and Lu, Chris and Foerster, Jakob and Clune, Jeff and Ha, David},
	month = apr,
	year = {2025},
	note = {arXiv:2504.08066},
}

@misc{schmidgall_agent_2025,
	title = {Agent {Laboratory}: {Using} {LLM} {Agents} as {Research} {Assistants}},
	shorttitle = {Agent {Laboratory}},
	url = {https://arxiv.org/abs/2501.04227},
	doi = {10.48550/arXiv.2501.04227},
	urldate = {2026-06-22},
	publisher = {arXiv},
	author = {Schmidgall, Samuel and Su, Yusheng and Wang, Ze and Sun, Ximeng and Wu, Jialian and Yu, Xiaodong and Liu, Jiang and Moor, Michael and Liu, Zicheng and Barsoum, Emad},
	month = jun,
	year = {2025},
	note = {arXiv:2501.04227},
}

@misc{lu_ai_2024,
	title = {The {AI} {Scientist}: {Towards} {Fully} {Automated} {Open}-{Ended} {Scientific} {Discovery}},
	shorttitle = {The {AI} {Scientist}},
	url = {https://arxiv.org/abs/2408.06292},
	doi = {10.48550/arXiv.2408.06292},
	urldate = {2026-06-22},
	publisher = {arXiv},
	author = {Lu, Chris and Lu, Cong and Lange, Robert Tjarko and Foerster, Jakob and Clune, Jeff and Ha, David},
	month = sep,
	year = {2024},
	note = {arXiv:2408.06292},
}

@misc{pepe_agentic_2026,
	title = {Agentic {Discovery} of {Neural} {Architectures}: {AIRA}-{Compose} and {AIRA}-{Design}},
	shorttitle = {Agentic {Discovery} of {Neural} {Architectures}},
	url = {https://arxiv.org/abs/2605.15871},
	doi = {10.48550/arXiv.2605.15871},
	urldate = {2026-06-22},
	publisher = {arXiv},
	author = {Pepe, Alberto and Lin, Chien-Yu and Magka, Despoina and Acun, Bilge and Wu, Yannan Nellie and Protopopov, Anton and Wu, Carole-Jean and Bachrach, Yoram},
	month = may,
	year = {2026},
	note = {arXiv:2605.15871},
}

@misc{wiemann_discoverphysics_2026,
	title = {{DiscoverPhysics}: {Benchmarking} {LLMs} for {Out}-of-the-{Box} {Scientific} {Thinking}},
	shorttitle = {{DiscoverPhysics}},
	url = {https://arxiv.org/abs/2605.26087},
	doi = {10.48550/arXiv.2605.26087},
	urldate = {2026-06-22},
	publisher = {arXiv},
	author = {Wiemann, Matt L. and Smith, Lindsay M. and Melchior, Peter and Mishra-Sharma, Siddharth and Wilson, Andrew Gordon and Izmailov, Pavel and Cuesta-L{\'a}zaro, Carolina},
	month = may,
	year = {2026},
	note = {arXiv:2605.26087},
}

@misc{du_accelerating_2026,
	title = {Accelerating {Scientific} {Discovery} with {Autonomous} {Goal}-evolving {Agents}},
	url = {https://arxiv.org/abs/2512.21782},
	doi = {10.48550/arXiv.2512.21782},
	urldate = {2026-06-22},
	publisher = {arXiv},
	author = {Du, Yuanqi and Yu, Botao and Liu, Tianyu and Shen, Tony and Chen, Junwu and Rittig, Jan G. and Sun, Kunyang and Zhang, Yikun and Krishnan, Aarti and Zhang, Yu and Rosen, Daniel and Pirone, Rosali and Song, Zhangde and Zhou, Bo and Masschelein, Cassandra and Wang, Yingze and Wang, Haorui and Jia, Haojun and Zhang, Chao and Zhao, Hongyu and Ester, Martin and Hacohen, Nir and Head-Gordon, Teresa and Gomes, Carla P. and Sun, Huan and Duan, Chenru and Schwaller, Philippe and Jin, Wengong},
	month = dec,
	year = {2025},
	note = {arXiv:2512.21782},
}

@misc{kabra_llm-autoscilab_2026,
	title = {{LLM}-{AutoSciLab}: {Closed}-{Loop} {Scientific} {Discovery} via {Active} {Experimentation} with {LLMs}},
	shorttitle = {{LLM}-{AutoSciLab}},
	url = {https://arxiv.org/abs/2605.24043},
	doi = {10.48550/arXiv.2605.24043},
	urldate = {2026-06-22},
	publisher = {arXiv},
	author = {Kabra, Sanchit and Abhyankar, Nikhil and Desai, Saaketh and Iyer, Prasad and Reddy, Chandan K.},
	month = may,
	year = {2026},
	note = {arXiv:2605.24043 
version: 1},
}

@misc{wan_pokeeresearch_2025,
	title = {Rethinking the {Design} of {Reinforcement} {Learning}-{Based} {Deep} {Research} {Agents}},
	shorttitle = {{PokeeResearch}},
	url = {https://arxiv.org/abs/2510.15862},
	doi = {10.48550/arXiv.2510.15862},
	urldate = {2026-06-22},
	publisher = {arXiv},
	author = {Wan, Yi and Wang, Jiuqi and Li, Liam and Liu, Jinsong and Zhu, Ruihao and Zhu, Zheqing},
	month = oct,
	year = {2025},
	note = {arXiv:2510.15862 
version: 3},
}

@misc{li_autosota_2026,
	title = {{AutoSOTA}: {An} {End}-to-{End} {Automated} {Research} {System} for {State}-of-the-{Art} {AI} {Model} {Discovery}},
	shorttitle = {{AutoSOTA}},
	url = {https://arxiv.org/abs/2604.05550},
	doi = {10.48550/arXiv.2604.05550},
	urldate = {2026-06-22},
	publisher = {arXiv},
	author = {Li, Yu and Shao, Chenyang and Liu, Xinyang and Zhao, Ruotong and Liu, Peijie and Su, Hongyuan and Chen, Zhibin and Yang, Qinglong and Xu, Anjie and Fang, Yi and Zeng, Qingbin and Li, Tianxing and Xu, Jingbo and Xu, Fengli and Li, Yong and Liu, Tie-Yan},
	month = may,
	year = {2026},
	note = {arXiv:2604.05550},
}

@misc{gao_autoscientists_2026,
	title = {{AutoScientists}: {Self}-{Organizing} {Agent} {Teams} for {Long}-{Running} {Scientific} {Experimentation}},
	shorttitle = {{AutoScientists}},
	url = {https://arxiv.org/abs/2605.28655},
	doi = {10.48550/arXiv.2605.28655},
	urldate = {2026-06-22},
	publisher = {arXiv},
	author = {Gao, Shanghua and Fang, Ada and Zitnik, Marinka},
	month = may,
	year = {2026},
	note = {arXiv:2605.28655},
}

@article{gottweis_towards_2025,
	title = {Accelerating scientific discovery with {Co-Scientist}},
	url = {https://doi.org/10.1038/s41586-026-10644-y},
	doi = {10.1038/s41586-026-10644-y},
	urldate = {2026-06-22},
	journal = {Nature},
	author = {Gottweis, Juraj and Weng, Wei-Hung and Daryin, Alexander and Tu, Tao and Sirkovic, Petar and Myaskovsky, Artiom and Glowaty, Grzegorz and Weissenberger, Felix and Orlandi, Alessio and Popovici, Dan and Palepu, Anil and Rong, Keran and Tanno, Ryutaro and Saab, Khaled and Zhang, Fan and Blum, Jacob and Carroll, Andrew and Kulkarni, Kavita and Tomasev, Nenad and Zverinski, Dina and Rendulic, Ivor and Vedadi, Elahe and Hasler, Florian and Rimanic, Luka and Boia, Marina and Budiselic, Ivan and Feinstein, Ben and Bellaiche, Mathias and Sheffer, Tom and Freyberg, Jan and Ratcliff, Jeremy and Bertolli, Ottavia and Chou, Katherine and Hassidim, Avinatan and Gokturk, Burak and Vahdat, Amin and Guan, Yuan and Dhillon, Vikram and Vaishnav, Eeshit Dhaval and Lee, Byron and Costa, Tiago R. D. and Penad{\'e}s, Jos{\'e} R. and Peltz, Gary and Matias, Yossi and Manyika, James and Hassabis, Demis and Xu, Yunhan and Kohli, Pushmeet and Pawlosky, Annalisa and Karthikesalingam, Alan and Natarajan, Vivek},
	year = {2026},
	note = {arXiv:2502.18864},
}

@misc{mishra_ainstein_2026,
	title = {{AInstein}: {Can} {LLMs} {Solve} {Research} {Problems} {From} {Parametric} {Memory} {Alone}?},
	shorttitle = {{AInstein}},
	url = {https://arxiv.org/abs/2510.05432},
	doi = {10.48550/arXiv.2510.05432},
	urldate = {2026-06-22},
	publisher = {arXiv},
	author = {Mishra, Shambhavi and Sahu, Gaurav and Pedersoli, Marco and Charlin, Laurent and Dolz, Jose and Pal, Christopher},
	month = oct,
	year = {2025},
	note = {arXiv:2510.05432},
}

@misc{goel_training_2025,
	title = {Training {AI} {Co}-{Scientists} {Using} {Rubric} {Rewards}},
	url = {https://arxiv.org/abs/2512.23707},
	doi = {10.48550/arXiv.2512.23707},
	urldate = {2026-06-22},
	publisher = {arXiv},
	author = {Goel, Shashwat and Hazra, Rishi and Jayalath, Dulhan and Willi, Timon and Jain, Parag and Shen, William F. and Leontiadis, Ilias and Barbieri, Francesco and Bachrach, Yoram and Geiping, Jonas and Whitehouse, Chenxi},
	month = dec,
	year = {2025},
	note = {arXiv:2512.23707},
}

@misc{jin_toward_2026,
	title = {Toward {Generalist} {Autonomous} {Research} via {Hypothesis}-{Tree} {Refinement}},
	url = {https://arxiv.org/abs/2606.11926},
	doi = {10.48550/arXiv.2606.11926},
	urldate = {2026-06-22},
	publisher = {arXiv},
	author = {Jin, Jiajie and Hu, Yuyang and Qiu, Kai and Dai, Qi and Luo, Chong and Dong, Guanting and Li, Xiaoxi and Zhao, Tong and Ma, Xiaolong and Zhang, Gongrui and Wu, Zhirong and Liu, Bei and Yang, Zhengyuan and Li, Linjie and Wang, Lijuan and Qian, Hongjin and Zhu, Yutao and Dou, Zhicheng},
	month = jun,
	year = {2026},
	note = {arXiv:2606.11926},
}

@inproceedings{agarwal_autodiscovery_2026,
	title = {{AutoDiscovery}: {Open}-ended {Scientific} {Discovery} via {Bayesian} {Surprise}},
	shorttitle = {{AutoDiscovery}},
	url = {https://arxiv.org/abs/2507.00310},
	doi = {10.48550/arXiv.2507.00310},
	urldate = {2026-06-22},
	author = {Agarwal, Dhruv and Majumder, Bodhisattwa Prasad and Adamson, Reece and Chakravorty, Megha and Gavireddy, Satvika Reddy and Parashar, Aditya and Surana, Harshit and Mishra, Bhavana Dalvi and McCallum, Andrew and Sabharwal, Ashish and Clark, Peter},
	booktitle = {Advances in Neural Information Processing Systems},
	year = {2025},
	note = {arXiv:2507.00310},
}

@misc{duan_bayes-entropy_2025,
	title = {Bayes-{Entropy} {Collaborative} {Driven} {Agents} for {Research} {Hypotheses} {Generation} and {Optimization}},
	url = {https://arxiv.org/abs/2508.01746},
	doi = {10.48550/arXiv.2508.01746},
	urldate = {2026-06-22},
	publisher = {arXiv},
	author = {Duan, Shiyang and Tian, Yuan and Bing, Qi and Shao, Xiaowei},
	month = aug,
	year = {2025},
	note = {arXiv:2508.01746},
}

@misc{kausik2026cgdp,
  title  = {The Context Gathering Decision Process: A {POMDP} Framework for Agentic Search},
  author = {Kausik, Chinmaya and Swaminathan, Adith and Kallus, Nathan},
  year   = {2026},
  note   = {arXiv:2605.07042},
  url    = {https://arxiv.org/abs/2605.07042},
}

@article{samanta_credit_2026,
  title   = {Credit Assignment with Resets in Language Model Reasoning},
  author  = {Samanta, Ankur and Magesh, Akshayaa and Jain, Ayush and Yu, Youliang and Jiang, Daniel and Asadi, Kavosh and Hassani, Kaveh and Sajda, Paul and Bhandari, Jalaj and Efroni, Yonathan},
  journal = {arXiv preprint arXiv:2605.25507},
  year    = {2026},
  note    = {arXiv:2605.25507v2},
  url     = {https://arxiv.org/abs/2605.25507},
}

@article{russell_principles_1991,
  title   = {Principles of Metareasoning},
  author  = {Russell, Stuart J. and Wefald, Eric H.},
  journal = {Artificial Intelligence},
  volume  = {49},
  number  = {1--3},
  pages   = {361--395},
  year    = {1991},
  doi     = {10.1016/0004-3702(91)90015-C},
}

@inproceedings{mnih_async_2016,
  title     = {Asynchronous Methods for Deep Reinforcement Learning},
  author    = {Mnih, Volodymyr and Badia, Adri\`a Puigdom\`enech and Mirza,
               Mehdi and Graves, Alex and Lillicrap, Timothy P. and Harley,
               Tim and Silver, David and Kavukcuoglu, Koray},
  booktitle = {Proceedings of the 33rd International Conference on Machine Learning},
  volume    = {48},
  pages     = {1928--1937},
  year      = {2016},
  publisher = {PMLR},
  url       = {https://arxiv.org/abs/1602.01783},
}

@inproceedings{cobbe_phasic_2021,
  title     = {Phasic Policy Gradient},
  author    = {Cobbe, Karl and Hilton, Jacob and Klimov, Oleg and Schulman, John},
  booktitle = {Proceedings of the 38th International Conference on Machine Learning},
  volume    = {139},
  pages     = {2020--2027},
  year      = {2021},
  publisher = {PMLR},
  url       = {https://arxiv.org/abs/2009.04416},
}

@inproceedings{jaderberg_unreal_2017,
  title     = {Reinforcement Learning with Unsupervised Auxiliary Tasks},
  author    = {Jaderberg, Max and Mnih, Volodymyr and Czarnecki, Wojciech
               Marian and Schaul, Tom and Leibo, Joel Z. and Silver, David and
               Kavukcuoglu, Koray},
  booktitle = {International Conference on Learning Representations},
  year      = {2017},
  url       = {https://arxiv.org/abs/1611.05397},
}

@inproceedings{zhang_generative_verifiers_2024,
  title         = {Generative Verifiers: Reward Modeling as Next-Token Prediction},
  author        = {Zhang, Lunjun and Hosseini, Arian and Bansal, Hritik and
                   Kazemi, Mehran and Kumar, Aviral and Agarwal, Rishabh},
  booktitle     = {International Conference on Learning Representations},
  year          = {2025},
  eprint        = {2408.15240},
  archivePrefix = {arXiv},
  primaryClass  = {cs.LG},
  url           = {https://arxiv.org/abs/2408.15240},
}

@misc{shan_genac_2026,
  title         = {Bringing Value Models Back: Generative Critics for Value Modeling in {LLM} Reinforcement Learning},
  author        = {Shan, Zikang and Zhong, Han and Wang, Liwei and Zhao, Li},
  year          = {2026},
  eprint        = {2604.10701},
  archivePrefix = {arXiv},
  primaryClass  = {cs.LG},
  url           = {https://arxiv.org/abs/2604.10701},
}

@misc{samanta_bayesbench_2026,
  title         = {{BayesBench}: Evaluating {LLM} Belief Trajectories Under Multi-Turn Evidence Accumulation},
  author        = {Samanta, Ankur and Magesh, Akshayaa and Lancewicki, Tal and Jain, Ayush and Yu, Youliang and Sajda, Paul and Hassani, Kaveh and Modi, Aditya and Jiang, Daniel R. and Efroni, Yonathan},
  year          = {2026},
  eprint        = {2606.30850},
  archivePrefix = {arXiv},
  primaryClass  = {cs.CL},
  url           = {https://arxiv.org/abs/2606.30850},
}

@inproceedings{bellemare_distributional_2017,
  title     = {A Distributional Perspective on Reinforcement Learning},
  author    = {Bellemare, Marc G. and Dabney, Will and Munos, R{\'e}mi},
  booktitle = {Proceedings of the 34th International Conference on Machine Learning},
  volume    = {70},
  pages     = {449--458},
  year      = {2017},
  publisher = {PMLR},
  url       = {https://arxiv.org/abs/1707.06887},
}

@inproceedings{mak_residual_2025,
  title     = {Residual Matrix Transformers: Scaling the Size of the Residual Stream},
  author    = {Mak, Brian and Flanigan, Jeffrey},
  booktitle = {Proceedings of the 42nd International Conference on Machine Learning},
  year      = {2025},
  note      = {arXiv:2506.22696},
  url       = {https://arxiv.org/abs/2506.22696},
}

@inproceedings{luong_towards_2025,
  title     = {Towards Robust Mathematical Reasoning},
  author    = {Luong, Thang and Hwang, Dawsen and Nguyen, Hoang H. and Ghiasi, Golnaz and Chervonyi, Yuri and Seo, Insuk and Kim, Junsu and Bingham, Garrett and Lee, Jonathan and Mishra, Swaroop and Zhai, Alex and Hu, Clara Huiyi and Michalewski, Henryk and Kim, Jimin and Ahn, Jeonghyun and Bae, Junhwi and Song, Xingyou and Trinh, Trieu H. and Le, Quoc V. and Jung, Junehyuk},
  booktitle = {Proceedings of the 2025 Conference on Empirical Methods in Natural Language Processing},
  year      = {2025},
  doi       = {10.18653/v1/2025.emnlp-main.1794},
  url       = {https://aclanthology.org/2025.emnlp-main.1794/},
}

@article{geiping_scaling_2025,
  title = {Scaling up Test-Time Compute with Latent Reasoning: A Recurrent Depth Approach},
  author = {Geiping, Jonas and McLeish, Sean and Jain, Neel and Kirchenbauer, John and Singh, Siddharth and Bartoldson, Brian R. and Kailkhura, Bhavya and Bhatele, Abhinav and Goldstein, Tom},
  journal = {arXiv preprint arXiv:2502.05171},
  year = {2025},
  url = {https://arxiv.org/abs/2502.05171}
}

@inproceedings{ye_differential_2025,
  title     = {Differential Transformer},
  author    = {Ye, Tianzhu and Dong, Li and Xia, Yuqing and Sun, Yutao and Zhu, Yi and Huang, Gao and Wei, Furu},
  booktitle = {International Conference on Learning Representations},
  year      = {2025},
  note      = {Oral presentation},
  url       = {https://arxiv.org/abs/2410.05258},
}

@inproceedings{vaswani_attention_2017,
  title     = {Attention Is All You Need},
  author    = {Vaswani, Ashish and Shazeer, Noam and Parmar, Niki and Uszkoreit, Jakob and Jones, Llion and Gomez, Aidan N. and Kaiser, Lukasz and Polosukhin, Illia},
  booktitle = {Advances in Neural Information Processing Systems},
  year      = {2017},
  url       = {https://arxiv.org/abs/1706.03762},
}

@inproceedings{schulman_gae_2016,
  title     = {High-Dimensional Continuous Control Using Generalized Advantage Estimation},
  author    = {Schulman, John and Moritz, Philipp and Levine, Sergey and Jordan, Michael and Abbeel, Pieter},
  booktitle = {International Conference on Learning Representations},
  year      = {2016},
  url       = {https://arxiv.org/abs/1506.02438},
}

@inproceedings{hu_lora_2022,
  title     = {{LoRA}: Low-Rank Adaptation of Large Language Models},
  author    = {Hu, Edward J. and Shen, Yelong and Wallis, Phillip and Allen-Zhu, Zeyuan and Li, Yuanzhi and Wang, Shean and Wang, Lu and Chen, Weizhu},
  booktitle = {International Conference on Learning Representations},
  year      = {2022},
  url       = {https://arxiv.org/abs/2106.09685},
}

@book{sutton_barto_2018,
  title     = {Reinforcement Learning: An Introduction},
  author    = {Sutton, Richard S. and Barto, Andrew G.},
  edition   = {2},
  publisher = {MIT Press},
  year      = {2018},
  url       = {https://mitpress.mit.edu/9780262039246/reinforcement-learning/},
}

@misc{qwen_qwen36_27b_2026,
  author       = {{Qwen Team}},
  title        = {{Qwen3.6-27B}},
  year         = {2026},
  howpublished = {Hugging Face model repository},
  url          = {https://huggingface.co/Qwen/Qwen3.6-27B/tree/6a9e13bd6fc8f0983b9b99948120bc37f49c13e9},
  note         = {Revision 6a9e13bd6fc8f0983b9b99948120bc37f49c13e9},
}

@misc{openai2025codex,
  title        = {Introducing {Codex}},
  author       = {{OpenAI}},
  year         = {2025},
  howpublished = {\url{https://openai.com/index/introducing-codex/}},
}

@misc{tallec_warp_time_2018,
  title         = {Can Recurrent Neural Networks Warp Time?},
  author        = {Tallec, Corentin and Ollivier, Yann},
  year          = {2018},
  eprint        = {1804.11188},
  archiveprefix = {arXiv},
  url           = {https://arxiv.org/abs/1804.11188},
}

@misc{liu_coordconv_2018,
  title         = {An Intriguing Failing of Convolutional Neural Networks and the {CoordConv} Solution},
  author        = {Liu, Rosanne and Lehman, Joel and Molino, Piero and Such, Felipe Petroski and Frank, Eric and Sergeev, Alex and Yosinski, Jason},
  year          = {2018},
  eprint        = {1807.03247},
  archiveprefix = {arXiv},
  url           = {https://arxiv.org/abs/1807.03247},
}

@misc{you_deep_lattice_2017,
  title         = {Deep Lattice Networks and Partial Monotonic Functions},
  author        = {You, Seungil and Ding, David and Canini, Kevin and Pfeifer, Jan and Gupta, Maya},
  year          = {2017},
  eprint        = {1709.06680},
  archiveprefix = {arXiv},
  url           = {https://arxiv.org/abs/1709.06680},
}

@misc{zaheer_deep_sets_2017,
  title         = {Deep Sets},
  author        = {Zaheer, Manzil and Kottur, Satwik and Ravanbakhsh, Siamak and Poczos, Barnabas and Salakhutdinov, Ruslan and Smola, Alexander},
  year          = {2017},
  eprint        = {1703.06114},
  archiveprefix = {arXiv},
  url           = {https://arxiv.org/abs/1703.06114},
}

@article{lu_deeponet_2021,
  title   = {Learning Nonlinear Operators via {DeepONet} Based on the Universal Approximation Theorem of Operators},
  author  = {Lu, Lu and Jin, Pengzhan and Pang, Guofei and Zhang, Zhongqiang and Karniadakis, George Em},
  journal = {Nature Machine Intelligence},
  volume  = {3},
  pages   = {218--229},
  year    = {2021},
  doi     = {10.1038/s42256-021-00302-5},
  url     = {https://doi.org/10.1038/s42256-021-00302-5},
}

@misc{chen_dynamical_isometry_2018,
  title         = {Dynamical Isometry and a Mean Field Theory of {RNNs}: Gating Enables Signal Propagation in Recurrent Neural Networks},
  author        = {Chen, Minmin and Pennington, Jeffrey and Schoenholz, Samuel S.},
  year          = {2018},
  eprint        = {1806.05394},
  archiveprefix = {arXiv},
  url           = {https://arxiv.org/abs/1806.05394},
}

@misc{satorras_egnn_2021,
  title         = {{$E(n)$} Equivariant Graph Neural Networks},
  author        = {Satorras, Victor Garcia and Hoogeboom, Emiel and Welling, Max},
  year          = {2021},
  eprint        = {2102.09844},
  archiveprefix = {arXiv},
  url           = {https://arxiv.org/abs/2102.09844},
}

@misc{tancik_fourier_features_2020,
  title         = {Fourier Features Let Networks Learn High Frequency Functions in Low Dimensional Domains},
  author        = {Tancik, Matthew and Srinivasan, Pratul P. and Mildenhall, Ben and Fridovich-Keil, Sara and Raghavan, Nithin and Singhal, Utkarsh and Ramamoorthi, Ravi and Barron, Jonathan T. and Ng, Ren},
  year          = {2020},
  eprint        = {2006.10739},
  archiveprefix = {arXiv},
  url           = {https://arxiv.org/abs/2006.10739},
}

@misc{wisdom_full_capacity_2016,
  title         = {Full-Capacity Unitary Recurrent Neural Networks},
  author        = {Wisdom, Scott and Powers, Thomas and Hershey, John R. and Le Roux, Jonathan and Atlas, Les},
  year          = {2016},
  eprint        = {1611.00035},
  archiveprefix = {arXiv},
  url           = {https://arxiv.org/abs/1611.00035},
}

@misc{xu_powerful_gnn_2018,
  title         = {How Powerful are Graph Neural Networks?},
  author        = {Xu, Keyulu and Hu, Weihua and Leskovec, Jure and Jegelka, Stefanie},
  year          = {2018},
  eprint        = {1810.00826},
  archiveprefix = {arXiv},
  url           = {https://arxiv.org/abs/1810.00826},
}

@misc{velickovic_graph_attention_2017,
  title         = {Graph Attention Networks},
  author        = {Veli{\v{c}}kovi{\'c}, Petar and Cucurull, Guillem and Casanova, Arantxa and Romero, Adriana and Li{\`o}, Pietro and Bengio, Yoshua},
  year          = {2017},
  eprint        = {1710.10903},
  archiveprefix = {arXiv},
  url           = {https://arxiv.org/abs/1710.10903},
}

@misc{kalchbrenner_grid_lstm_2015,
  title         = {Grid Long Short-Term Memory},
  author        = {Kalchbrenner, Nal and Danihelka, Ivo and Graves, Alex},
  year          = {2015},
  eprint        = {1507.01526},
  archiveprefix = {arXiv},
  url           = {https://arxiv.org/abs/1507.01526},
}

@misc{cohen_group_equivariant_2016,
  title         = {Group Equivariant Convolutional Networks},
  author        = {Cohen, Taco S. and Welling, Max},
  year          = {2016},
  eprint        = {1602.07576},
  archiveprefix = {arXiv},
  url           = {https://arxiv.org/abs/1602.07576},
}

@misc{mena_gumbel_sinkhorn_2018,
  title         = {Learning Latent Permutations with {Gumbel-Sinkhorn} Networks},
  author        = {Mena, Gonzalo and Belanger, David and Linderman, Scott and Snoek, Jasper},
  year          = {2018},
  eprint        = {1802.08665},
  archiveprefix = {arXiv},
  url           = {https://arxiv.org/abs/1802.08665},
}

@misc{gu_hippo_2020,
  title         = {{HiPPO}: Recurrent Memory with Optimal Polynomial Projections},
  author        = {Gu, Albert and Dao, Tri and Ermon, Stefano and Rudra, Atri and R{\'e}, Christopher},
  year          = {2020},
  eprint        = {2008.07669},
  archiveprefix = {arXiv},
  url           = {https://arxiv.org/abs/2008.07669},
}

@misc{poli_hyena_2023,
  title         = {Hyena Hierarchy: Towards Larger Convolutional Language Models},
  author        = {Poli, Michael and Massaroli, Stefano and Nguyen, Eric and Fu, Daniel Y. and Dao, Tri and Baccus, Stephen and Bengio, Yoshua and Ermon, Stefano and R{\'e}, Christopher},
  year          = {2023},
  eprint        = {2302.10866},
  archiveprefix = {arXiv},
  url           = {https://arxiv.org/abs/2302.10866},
}

@misc{ha_hypernetworks_2016,
  title         = {{HyperNetworks}},
  author        = {Ha, David and Dai, Andrew and Le, Quoc V.},
  year          = {2016},
  eprint        = {1609.09106},
  archiveprefix = {arXiv},
  url           = {https://arxiv.org/abs/1609.09106},
}

@misc{li_indrnn_2018,
  title         = {Independently Recurrent Neural Network ({IndRNN}): Building a Longer and Deeper {RNN}},
  author        = {Li, Shuai and Li, Wanqing and Cook, Chris and Zhu, Ce and Gao, Yanbo},
  year          = {2018},
  eprint        = {1803.04831},
  archiveprefix = {arXiv},
  url           = {https://arxiv.org/abs/1803.04831},
}

@misc{battaglia_interaction_networks_2016,
  title         = {Interaction Networks for Learning about Objects, Relations and Physics},
  author        = {Battaglia, Peter W. and Pascanu, Razvan and Lai, Matthew and Rezende, Danilo and Kavukcuoglu, Koray},
  year          = {2016},
  eprint        = {1612.00222},
  archiveprefix = {arXiv},
  url           = {https://arxiv.org/abs/1612.00222},
}

@misc{westhuizen_forget_gate_2018,
  title         = {The Unreasonable Effectiveness of the Forget Gate},
  author        = {{van der Westhuizen}, Jos and Lasenby, Joan},
  year          = {2018},
  eprint        = {1804.04849},
  archiveprefix = {arXiv},
  url           = {https://arxiv.org/abs/1804.04849},
}

@misc{murphy_janossy_pooling_2018,
  title         = {Janossy Pooling: Learning Deep Permutation-Invariant Functions for Variable-Size Inputs},
  author        = {Murphy, Ryan L. and Srinivasan, Balasubramaniam and Rao, Vinayak and Ribeiro, Bruno},
  year          = {2018},
  eprint        = {1811.01900},
  archiveprefix = {arXiv},
  url           = {https://arxiv.org/abs/1811.01900},
}

@misc{liu_kan_2024,
  title         = {{KAN}: Kolmogorov-Arnold Networks},
  author        = {Liu, Ziming and Wang, Yixuan and Vaidya, Sachin and Ruehle, Fabian and Halverson, James and Solja{\v{c}}i{\'c}, Marin and Hou, Thomas Y. and Tegmark, Max},
  year          = {2024},
  eprint        = {2404.19756},
  archiveprefix = {arXiv},
  url           = {https://arxiv.org/abs/2404.19756},
}

@inproceedings{voelker_lmu_2019,
  title     = {Legendre Memory Units: Continuous-Time Representation in Recurrent Neural Networks},
  author    = {Voelker, Aaron and Kaji{\'c}, Ivana and Eliasmith, Chris},
  booktitle = {Advances in Neural Information Processing Systems},
  volume    = {32},
  pages     = {15544--15553},
  year      = {2019},
  url       = {https://papers.nips.cc/paper/9689-legendre-memory-units-continuous-time-representation-in-recurrent-neural-networks},
}

@misc{orvieto_lru_2023,
  title         = {Resurrecting Recurrent Neural Networks for Long Sequences},
  author        = {Orvieto, Antonio and Smith, Samuel L. and Gu, Albert and Fernando, Anushan and Gulcehre, Caglar and Pascanu, Razvan and De, Soham},
  year          = {2023},
  eprint        = {2303.06349},
  archiveprefix = {arXiv},
  url           = {https://arxiv.org/abs/2303.06349},
}

@article{hochreiter_lstm_1997,
  title   = {Long Short-Term Memory},
  author  = {Hochreiter, Sepp and Schmidhuber, J{\"u}rgen},
  journal = {Neural Computation},
  volume  = {9},
  number  = {8},
  pages   = {1735--1780},
  year    = {1997},
  doi     = {10.1162/neco.1997.9.8.1735},
  url     = {https://doi.org/10.1162/neco.1997.9.8.1735},
}

@misc{gu_mamba_2023,
  title         = {Mamba: Linear-Time Sequence Modeling with Selective State Spaces},
  author        = {Gu, Albert and Dao, Tri},
  year          = {2023},
  eprint        = {2312.00752},
  archiveprefix = {arXiv},
  url           = {https://arxiv.org/abs/2312.00752},
}

@misc{weston_memory_networks_2014,
  title         = {Memory Networks},
  author        = {Weston, Jason and Chopra, Sumit and Bordes, Antoine},
  year          = {2014},
  eprint        = {1410.3916},
  archiveprefix = {arXiv},
  url           = {https://arxiv.org/abs/1410.3916},
}

@misc{chen_minimalrnn_2017,
  title         = {{MinimalRNN}: Toward More Interpretable and Trainable Recurrent Neural Networks},
  author        = {Chen, Minmin},
  year          = {2017},
  eprint        = {1711.06788},
  archiveprefix = {arXiv},
  url           = {https://arxiv.org/abs/1711.06788},
}

@misc{ramsauer_hopfield_2020,
  title         = {Hopfield Networks is All You Need},
  author        = {Ramsauer, Hubert and Sch{\"a}fl, Bernhard and Lehner, Johannes and Seidl, Philipp and Widrich, Michael and Adler, Thomas and Gruber, Lukas and Holzleitner, Markus and Pavlovi{\'c}, Milena and Sandve, Geir Kjetil and Greiff, Victor and Kreil, David and Kopp, Michael and Klambauer, G{\"u}nter and Brandstetter, Johannes and Hochreiter, Sepp},
  year          = {2020},
  eprint        = {2008.02217},
  archiveprefix = {arXiv},
  url           = {https://arxiv.org/abs/2008.02217},
}

@misc{madsen_neural_arithmetic_units_2020,
  title         = {Neural Arithmetic Units},
  author        = {Madsen, Andreas and Johansen, Alexander Rosenberg},
  year          = {2020},
  eprint        = {2001.05016},
  archiveprefix = {arXiv},
  url           = {https://arxiv.org/abs/2001.05016},
}

@misc{chen_neural_ode_2018,
  title         = {Neural Ordinary Differential Equations},
  author        = {Chen, Ricky T. Q. and Rubanova, Yulia and Bettencourt, Jesse and Duvenaud, David},
  year          = {2018},
  eprint        = {1806.07366},
  archiveprefix = {arXiv},
  url           = {https://arxiv.org/abs/1806.07366},
}

@misc{garnelo_neural_processes_2018,
  title         = {Neural Processes},
  author        = {Garnelo, Marta and Schwarz, Jonathan and Rosenbaum, Dan and Viola, Fabio and Rezende, Danilo J. and Eslami, S. M. Ali and Teh, Yee Whye},
  year          = {2018},
  eprint        = {1807.01622},
  archiveprefix = {arXiv},
  url           = {https://arxiv.org/abs/1807.01622},
}

@misc{graves_neural_turing_2014,
  title         = {Neural Turing Machines},
  author        = {Graves, Alex and Wayne, Greg and Danihelka, Ivo},
  year          = {2014},
  eprint        = {1410.5401},
  archiveprefix = {arXiv},
  url           = {https://arxiv.org/abs/1410.5401},
}

@misc{amos_optnet_2017,
  title         = {{OptNet}: Differentiable Optimization as a Layer in Neural Networks},
  author        = {Amos, Brandon and Kolter, J. Zico},
  year          = {2017},
  eprint        = {1703.00443},
  archiveprefix = {arXiv},
  url           = {https://arxiv.org/abs/1703.00443},
}

@misc{chilkuri_parallel_lmu_2021,
  title         = {Parallelizing Legendre Memory Unit Training},
  author        = {Chilkuri, Narsimha and Eliasmith, Chris},
  year          = {2021},
  eprint        = {2102.11417},
  archiveprefix = {arXiv},
  url           = {https://arxiv.org/abs/2102.11417},
}

@misc{long_pde_net_2017,
  title         = {{PDE-Net}: Learning {PDEs} from Data},
  author        = {Long, Zichao and Lu, Yiping and Ma, Xianzhong and Dong, Bin},
  year          = {2017},
  eprint        = {1710.09668},
  archiveprefix = {arXiv},
  url           = {https://arxiv.org/abs/1710.09668},
}

@misc{jaegle_perceiver_io_2021,
  title         = {Perceiver {IO}: A General Architecture for Structured Inputs \& Outputs},
  author        = {Jaegle, Andrew and Borgeaud, Sebastian and Alayrac, Jean-Baptiste and Doersch, Carl and Ionescu, Catalin and Ding, David and Koppula, Skanda and Zoran, Daniel and Brock, Andrew and Shelhamer, Evan and H{\'e}naff, Olivier and Botvinick, Matthew M. and Zisserman, Andrew and Vinyals, Oriol and Carreira, Jo{\~a}o},
  year          = {2021},
  eprint        = {2107.14795},
  archiveprefix = {arXiv},
  url           = {https://arxiv.org/abs/2107.14795},
}

@misc{choromanski_performer_2020,
  title         = {Rethinking Attention with Performers},
  author        = {Choromanski, Krzysztof and Likhosherstov, Valerii and Dohan, David and Song, Xingyou and Gane, Andreea and Sarlos, Tamas and Hawkins, Peter and Davis, Jared and Mohiuddin, Afroz and Kaiser, Lukasz and Belanger, David and Colwell, Lucy and Weller, Adrian},
  year          = {2020},
  eprint        = {2009.14794},
  archiveprefix = {arXiv},
  url           = {https://arxiv.org/abs/2009.14794},
}

@misc{neil_phased_lstm_2016,
  title         = {Phased {LSTM}: Accelerating Recurrent Network Training for Long or Event-based Sequences},
  author        = {Neil, Daniel and Pfeiffer, Michael and Liu, Shih-Chii},
  year          = {2016},
  eprint        = {1610.09513},
  archiveprefix = {arXiv},
  url           = {https://arxiv.org/abs/1610.09513},
}

@misc{vinyals_pointer_networks_2015,
  title         = {Pointer Networks},
  author        = {Vinyals, Oriol and Fortunato, Meire and Jaitly, Navdeep},
  year          = {2015},
  eprint        = {1506.03134},
  archiveprefix = {arXiv},
  url           = {https://arxiv.org/abs/1506.03134},
}

@misc{zilly_recurrent_highway_2016,
  title         = {Recurrent Highway Networks},
  author        = {Zilly, Julian Georg and Srivastava, Rupesh Kumar and Koutn{\'i}k, Jan and Schmidhuber, J{\"u}rgen},
  year          = {2016},
  eprint        = {1607.03474},
  archiveprefix = {arXiv},
  url           = {https://arxiv.org/abs/1607.03474},
}

@misc{gu_s4_2021,
  title         = {Efficiently Modeling Long Sequences with Structured State Spaces},
  author        = {Gu, Albert and Goel, Karan and R{\'e}, Christopher},
  year          = {2021},
  eprint        = {2111.00396},
  archiveprefix = {arXiv},
  url           = {https://arxiv.org/abs/2111.00396},
}

@misc{helfrich_scaled_cayley_2017,
  title         = {Orthogonal Recurrent Neural Networks with Scaled {Cayley} Transform},
  author        = {Helfrich, Kyle and Willmott, Devin and Ye, Qiang},
  year          = {2017},
  eprint        = {1707.09520},
  archiveprefix = {arXiv},
  url           = {https://arxiv.org/abs/1707.09520},
}

@misc{sitzmann_siren_2020,
  title         = {Implicit Neural Representations with Periodic Activation Functions},
  author        = {Sitzmann, Vincent and Martel, Julien N. P. and Bergman, Alexander W. and Lindell, David B. and Wetzstein, Gordon},
  year          = {2020},
  eprint        = {2006.09661},
  archiveprefix = {arXiv},
  url           = {https://arxiv.org/abs/2006.09661},
}

@misc{locatello_slot_attention_2020,
  title         = {Object-Centric Learning with Slot Attention},
  author        = {Locatello, Francesco and Weissenborn, Dirk and Unterthiner, Thomas and Mahendran, Aravindh and Heigold, Georg and Uszkoreit, Jakob and Dosovitskiy, Alexey and Kipf, Thomas},
  year          = {2020},
  eprint        = {2006.15055},
  archiveprefix = {arXiv},
  url           = {https://arxiv.org/abs/2006.15055},
}

@misc{jaderberg_spatial_transformer_2015,
  title         = {Spatial Transformer Networks},
  author        = {Jaderberg, Max and Simonyan, Karen and Zisserman, Andrew and Kavukcuoglu, Koray},
  year          = {2015},
  eprint        = {1506.02025},
  archiveprefix = {arXiv},
  url           = {https://arxiv.org/abs/1506.02025},
}

@misc{arik_tabnet_2019,
  title         = {{TabNet}: Attentive Interpretable Tabular Learning},
  author        = {Arik, Sercan O. and Pfister, Tomas},
  year          = {2019},
  eprint        = {1908.07442},
  archiveprefix = {arXiv},
  url           = {https://arxiv.org/abs/1908.07442},
}

@misc{jolicoeur_martineau_tiny_recursive_2025,
  title         = {Less is More: Recursive Reasoning with Tiny Networks},
  author        = {Jolicoeur-Martineau, Alexia},
  year          = {2025},
  eprint        = {2510.04871},
  archiveprefix = {arXiv},
  url           = {https://arxiv.org/abs/2510.04871},
}

@misc{dehghani_universal_transformer_2018,
  title         = {Universal Transformers},
  author        = {Dehghani, Mostafa and Gouws, Stephan and Vinyals, Oriol and Uszkoreit, Jakob and Kaiser, {\L}ukasz},
  year          = {2018},
  eprint        = {1807.03819},
  archiveprefix = {arXiv},
  url           = {https://arxiv.org/abs/1807.03819},
}

@misc{arora_zoology_2023,
  title         = {Zoology: Measuring and Improving Recall in Efficient Language Models},
  author        = {Arora, Simran and Eyuboglu, Sabri and Timalsina, Aman and Johnson, Isys and Poli, Michael and Zou, James and Rudra, Atri and R{\'e}, Christopher},
  year          = {2023},
  eprint        = {2312.04927},
  archiveprefix = {arXiv},
  url           = {https://arxiv.org/abs/2312.04927},
}
\ifmetapaper
  \bibliographystyle{assets/plainnat}
\else
  \bibliographystyle{iclr2027_conference}
\fi

\clearpage
\newpage
\ifmetapaper
  \beginappendix
\else
  \appendix
\fi

\section*{Appendix Contents}

\begingroup
\setlength{\fboxsep}{12pt}
\noindent\colorbox{miraorangebg}{%
  \parbox{\dimexpr\linewidth-2\fboxsep\relax}{%
    \small
    \ifmetapaper
      \textcolor{miraorange}{\textbf{Background}}\par
    \else
      \textcolor{miraorange}{\textbf{Context and limitations}}\par
    \fi
    \vspace{0.35em}
    \ifmetapaper
    \else
      \appendixcontentsline{app:limitations}
    \fi
    \appendixcontentsline{sec:related}

    \vspace{0.8em}
    \textcolor{miraorange}{\textbf{Experimental setup}}\par
    \vspace{0.35em}
    \appendixcontentsline{sec:environments}
    \appendixcontentsline{app:mira-implementation}

    \vspace{0.8em}
    \textcolor{miraorange}{\textbf{Additional MIRA analyses}}\par
    \vspace{0.35em}
    \appendixcontentsline{app:semantic-graphs}
    \appendixcontentsline{app:rmt-analysis}
    \appendixcontentsline{app:loop-analysis}

    \vspace{0.8em}
    \textcolor{miraorange}{\textbf{Critic and policy-learning details}}\par
    \vspace{0.35em}
    \appendixcontentsline{app:critic-details}
    \appendixcontentsline{app:implementation-details}
    \appendixcontentsline{app:trajectory_stitching}
    \appendixcontentsline{app:mira-ac-results}
    \appendixcontentsline{app:mira-ac-behavior}
  }%
}
\endgroup

\ifmetapaper
\else

\fi

\section{Related Work}
\label{sec:related}

\paragraph{Rational meta-reasoning.}
MIRA builds on the classical account in Principles of
Metareasoning~\citep{russell_principles_1991}, which asks which computation a
resource-bounded agent should perform next and when it should stop. In long-horizon research, the
relevant computations are investigations: testing a hypothesis, revising an
implementation, gathering evidence, or terminating the search. MIRA makes
these allocation decisions explicit as work orders, thereby turning a
classical meta-reasoning problem into a trainable sequential decision process.

\paragraph{Benchmarks for autonomous research.}
Recent benchmarks expose complementary aspects of long-horizon research.
MLAgentBench~\citep{huang_mlagentbench_2024} and
MLE-bench~\citep{chan_mle-bench_2024} evaluate iterative machine-learning
engineering over executable tasks, while
RE-Bench~\citep{wijk_re-bench_2025} compares agents with human experts over
increasing time budgets. PaperBench~\citep{starace_paperbench_2025} evaluates
end-to-end replication using author-developed hierarchical rubrics, and
HCAST~\citep{rein_hcast_2025} grounds task difficulty in the time required by
skilled humans. ScienceAgentBench~\citep{chen_scienceagentbench_2025} similarly
derives data-driven discovery tasks from published scientific workflows.

Other environments broaden either the training interface or the space of
research problems. MLE-Dojo~\citep{qiang_mle-dojo_2025} provides interactive
Kaggle environments with normalized rewards suitable for collecting learning
trajectories; PostTrainBench~\citep{rank_posttrainbench_2026} extends
autoresearch to coupled post-training decisions; and
FrontierScience~\citep{wang_frontierscience_2026} targets expert-level
scientific tasks. DiscoGen~\citep{goldie_procedural_2026} supplies procedurally
generated algorithm-discovery tasks with meta-train/meta-test splits, whereas
AlgoTune~\citep{press_algotune_2025} evaluates measurable speedups to established
numerical programs. Finally, Frontier-Eng~\citep{chi_frontier-eng_2026}
characterizes how improvements diminish over extended engineering runs, and
FML-bench~\citep{zou_fml-bench_2026} compares search strategies across
controlled landscapes. In addition to introducing HillClimbBench, we cast
several existing environments as instances of a common research decision
process and ask whether research strategy can be learned across trajectories.

\paragraph{Scientific and algorithmic discovery agents.}
Several systems automate substantial portions of the research lifecycle. The
AI Scientist~\citep{lu_ai_2024} and
AI Scientist-v2~\citep{yamada_ai_2025} combine experiment execution,
evaluation, and paper generation, with the latter adding tree search and
structured memory. FunSearch~\citep{funsearch} established
language-model-guided evolution as a mechanism for discovering programs, and
AlphaEvolve~\citep{novikov_alphaevolve_2025} scales this approach through model
ensembles, evaluation cascades, and quality-diversity search.
SAGA~\citep{du_accelerating_2026} extends automation to the evolution of
scientific objectives, while the AI co-scientist~\citep{gottweis_towards_2025}
and AutoDiscovery~\citep{agarwal_autodiscovery_2026} organize hypothesis
development through evolutionary selection and Bayesian surprise,
respectively.

Research agents also differ in how they divide and retain work.
R\&D-Agent~\citep{yang_rd-agent_2025} separates researcher and developer roles,
AutoScientists~\citep{gao_autoscientists_2026} coordinates decentralized teams
through shared experimental memory, and MARS~\citep{chen_mars_2026} performs
budget-aware tree search over repository states.
AutoSOTA~\citep{li_autosota_2026} targets reproduction and improvement of
existing models, LLM-AutoSciLab~\citep{kabra_llm-autoscilab_2026} casts
discovery as active experimental design, and Arbor~\citep{jin_toward_2026}
refines explicit hypothesis trees.
DiscoverPhysics~\citep{wiemann_discoverphysics_2026} evaluates open-ended
scientific reasoning in physics, whereas AInstein~\citep{mishra_ainstein_2026}
studies iterative critique using parametric knowledge alone. These systems
primarily improve a fixed model through inference-time organization. MIRA is
compatible with such execution mechanisms, but our central concern is learning
the policy that chooses the next research objective from prior experience.

\paragraph{Search, memory, and evolving agent structure.}
Existing agents encode a wide range of search biases.
AIDE~\citep{jiang_aide_2025} uses explicit drafting, improvement, and debugging
operators; LATS~\citep{zhou_language_2024} combines tree search with acting,
planning, reflection, and model-generated values; and
LoongFlow~\citep{wan_loongflow_2025} couples plan--execute--summarize cycles
with evolutionary archives. DataMaster~\citep{du_datamaster_2026} schedules
exploration over the available budget, AIRA~\citep{pepe_agentic_2026} applies
UCT-style search to neural-architecture discovery, and
SciMaster~\citep{chai_scimaster_2025} distributes exploration across multiple
agents. Automated AI-research Speedrunning~\citep{zhao_automated_2025} provides
a controlled autoresearch setting, while Agent
Laboratory~\citep{schmidgall_agent_2025} organizes research into fixed phases. The
diversity of successful scaffolds, together with the strategy-dependent results
of Frontier-Eng~\citep{chi_frontier-eng_2026} and
FML-bench~\citep{zou_fml-bench_2026}, motivates selecting research actions
conditionally rather than committing to one search heuristic for an entire
investigation.

A complementary line of work changes the agent's persistent artifacts.
SkillGrad~\citep{wang_skillgrad_2026} treats trajectory diagnoses as textual
gradients, and SkillOpt~\citep{yang_skillopt_2026} uses bounded edits and
validation to improve reusable skill documents. Harness self-evolution
analysis~\citep{lin_harness_2026} distinguishes the ability to modify a harness
from the ability to benefit from those modifications, while Compiling Agentic
Workflows into LLM Weights~\citep{dennis_compiling_2026} shows that procedural
workflows can be distilled into model weights. MIRA keeps the execution harness fixed in our
learning experiments and instead updates the model that allocates research
effort; evolving the harness and learning the meta-reasoning policy are
therefore complementary directions.

\paragraph{Feedback and value learning.}
Language feedback can convey substantially more structure than a scalar
reward, as illustrated empirically by LLF-Bench~\citep{cheng_llf-bench_2023}
and formalized by Formalizing Learning from Language Feedback with Provable
Guarantees~\citep{xu_formalizing_2026}. Training AI Co-Scientists Using Rubric
Rewards~\citep{goel_training_2025} uses rubric rewards to train research-plan
generation, while HypoAgents~\citep{duan_bayes-entropy_2025} uses Bayesian
belief and entropy updates to guide which hypotheses to refine. Our experiments instead use
scalar, outcome-grounded targets: either a gold-binary outcome or changes in
a proxy experimental metric. This choice isolates whether delayed research
outcomes alone can train value estimates at MIRA's decision boundaries;
versioned artifacts could support richer feedback in future work.

MIRA-AC also relates to shared actor--critic representation learning.
A3C~\citep{mnih_async_2016} uses a shared backbone with separate policy and
scalar-value outputs, while Phasic Policy Gradient~\citep{cobbe_phasic_2021}
studies the balance between useful representation transfer and interference
between policy and value objectives. UNREAL~\citep{jaderberg_unreal_2017}
demonstrates how auxiliary prediction objectives can shape a shared policy
representation. More recent generative approaches express verification or
value estimation through language modeling itself: Generative
Verifiers~\citep{zhang_generative_verifiers_2024} formulates verification as
next-token prediction, and GenAC~\citep{shan_genac_2026} uses a generative
critic for value modeling. MIRA-AC similarly uses
one language model and one next-token interface for both research actions and
return prediction, without adding a scalar head or maintaining a separately
resident critic.

\paragraph{Credit assignment in multi-turn agents.}
Several methods localize outcome supervision within long interactions. Credit
assignment can also proceed through a learned step-level critic: Iterative
GRPO~\citep{ippo} fits a turn-level $Q^\pi$ from logged Monte Carlo returns and
uses it as the reward model for single-turn policy improvement in batch-online
multi-turn settings.
Credit Assignment with Resets~\citep{samanta_credit_2026} uses counterfactual
continuations from intermediate states, and Tree-GRPO~\citep{ji_tree_2026}
obtains step-level relative signals from sibling branches that share a prefix.
R1-Searcher~\citep{song_r1-searcher_2025} learns retrieval behavior from
outcome-only reinforcement learning, StepSearch~\citep{wang_stepsearch_2025}
applies step-wise policy optimization to search-augmented reasoning, and
strategy-guided exploration~\citep{szot_expanding_2026} operates over
high-level language strategies. PokeeResearch~\citep{wan_pokeeresearch_2025}
studies reinforcement-learning design choices for deep-research agents.
MIRA-AC differs in where it places the
learning boundary: it reconstructs complete research episodes but applies
policy loss only to the complete sampled decision responses that allocate future
computation. Its decision-level generative critic assigns delayed credit at those
boundaries, and buffered asynchronous optimization accommodates investigations
with heterogeneous durations.

\section{Experimental Setup}
\label{sec:environments}

We study four autoresearch environments: BNLearn causal discovery, Physics
Discovery, SRBench symbolic regression, and HillClimbBench-CPU architecture
research. In the reported policy evaluation, BNLearn, Physics, and SRBench each
contain 100 tasks with one rollout per task. HillClimbBench-CPU contains 50
distinct paper-derived tasks with two independently initialized rollouts per
task. Each rollout permits at most 20 MIRA iterations.

\subsection{Shared Interaction Protocol}

Each task is an isolated, persistent workspace. At every iteration, a fresh
executor receives the MIRA work order, the fixed task prompt, and stable
environment-specific instructions. The prompt identifies the available data
and tools, interaction budget, editable files, submission artifact, and hidden
information. The executor may write analyses and programs, run proxy
evaluators, update its artifact, and leave evidence for later iterations. The
shared procedure governing outer-loop and executor interactions is described in
App.~\ref{app:mira-implementation}; below we specify the environment-facing
part of the prompt.

For the three scientific environments, policy training and reported evaluation
use independently initialized trajectories on the same fixed task identities
and seeds. They therefore test new research trajectories and held-out
evaluations within the same tasks, not generalization to unseen task identities.

\subsection{Defining Proxy Hill-Climbing Signals}
\label{sec:proxy-signals}

Each environment exposes a different proxy measure of progress: graph-edit
distance, trajectory error, expression error, or architecture performance. We
map that measure to a higher-is-better quality $q_t\in[0,1]$ after iteration
$t$, with $q_0$ denoting the fixed initial quality. We construct each map from
constants in the environment specification and freeze it before training.

The four maps use natural task-level anchors. BNLearn measures how much of the
empty graph's structural Hamming distance (SHD) has been removed. Physics
scales trajectory mean squared error (MSE) by position variance and applies a
fixed reciprocal map capped at $0.95$. SRBench measures log-normalized mean
squared error (log-NMSE) improvement from the initial expression toward the float64 numerical
floor. HillClimbBench-CPU measures improvement from the supplied reference
architecture toward the task's ideal score of one. The exact equations appear
below. The endpoints need not be attainable; they are placed to distinguish
plausible within-budget improvements.

These proxies provide a task-local ordering of partial progress. They turn
changes across successive research iterations into marginal rewards at a
resolution suited to the current model class and fixed research budget. The
mapping remains imperfect: it can miss real progress, reward a brittle shortcut,
or saturate too early. Gold evaluation therefore remains the final measure of
the selected artifact. It stays hidden during research, while reward depends on
increases in best-so-far proxy quality.

For a new environment, the construction begins with a repeatable proxy metric
that usually moves with the desired outcome. Natural anchors include a supplied
baseline, an empty prediction, or a task-defined target. A monotone transform
can spread raw errors that span orders of magnitude, while its scale should keep
plausible improvements away from a narrow or saturated range. Invalid artifacts
receive no improvement. The completed map is frozen before policy training.
Development tasks then establish its usable range and its association with gold
outcomes; poor alignment calls for redesign before final evaluation.

The $q_t$ values define best-so-far progress and marginal reward in
Sec.~\ref{sec:returns}; other visible diagnostics do not affect reward. After
selection, a hidden gold evaluation produces the score reported in the results
table. It uses the same hidden target as the proxy for BNLearn and separate
held-out data or conditions for the other environments. A task is solved when
this gold evaluation meets the environment-specific success criterion.

\subsection{Environment Interfaces}

\paragraph{BNLearn causal discovery.}
We construct 100 tasks from the Alarm (32 tasks), Child (17), Insurance (24),
and Water (27) networks in the Bayesian Network
Repository~\citep{scutari_bnrepository}. Each task is a randomly
permuted, anonymized 14--34-node induced subgraph. A linear-Gaussian structural
equation model provides 1,800 training and 600 visible validation observations.

The agent may analyze the visible data, query validation Bayesian information
criterion (BIC) diagnostics, and,
on an $n$-node task, request $n$ interventions that report estimated total
effects. These diagnostics do not determine reward, and the executor cannot
query graph accuracy while carrying out an investigation. The final artifact
is a directed acyclic graph (DAG).

After each completed iteration, the harness reports directed structural
Hamming distance (SHD), including missing, extra, and reversed edges. For the
empty-graph SHD $S_0$ and submitted-graph SHD $S_t$, proxy quality is
$q_t=\operatorname{clip}((S_0-S_t)/S_0,0,1)$.
The best model-authored graph is retained and restored for final scoring under
the same hidden-DAG objective. Its gold score is the normalized SHD gap
closure above, evaluated on the submitted graph; the task is solved only by
exact directed recovery.

\paragraph{Physics discovery.}
We adapt DiscoverPhysics~\citep{wiemann_discoverphysics_2026} into 100 anonymous
worlds spanning 12 mechanisms and three difficulty levels. The agent receives a
mission and topology-specific laboratory interface, but not the governing law.
Across 20 MIRA iterations, it may allocate 15 experiment rounds among the first
19; the last iteration is reserved for finalization. It may validate experiment
plans and repeatedly fit executable candidate laws to all acquired trajectories.

The agent selects an executable law and a short explanation. For
acquired-trajectory MSE $m_t$ and position variance $v_t$, proxy quality is
$q_t=0.95/(1+\sqrt{m_t/v_t})$.
After the fixed horizon, the law is evaluated once on held-out trajectories and
the explanation is graded once by a hosted GPT-5.5 judge using a hidden
world-specific reference and rubric. Let $n$ be trajectory MSE normalized by
the held-out world's ground-truth trajectory variance, and let $e$ be the
explanation score. For an unsolved task, the gold score is the product of
the threshold-scaled components $0.1/\max(n,0.1)$ and $\min(1,e/0.75)$, capped
strictly below one. A task is solved, and receives a gold score of one, when
$n<0.1$ and $e\geq0.75$.

\paragraph{SRBench symbolic regression.}
We use 100 LLM-SRBench problems~\citep{shojaee_llmsrbench_2025} from chemical
reactions (30 tasks), physical oscillators (27), materials science (23), and
population growth (20). Within each family, tasks span a frozen
reference-expression complexity ranking. The official training data are split
deterministically 80/20 into visible fit and validation sets; the
in-distribution (ID) test set, optional out-of-distribution (OOD) set, and
reference expression remain hidden.

The agent may analyze the available data freely and evaluate expressions within a
constrained SymPy-compatible grammar. It must select one final expression. The
environment reports proxy metrics but does not restore the candidate with the
highest proxy metric, so final selection remains with the agent.

For validation NMSE $N_t$, fixed initial-expression NMSE $B$, and float64 floor
$\epsilon$, proxy quality is
\begin{equation}
q_t=\operatorname{clip}\!\left(
\frac{\log(B/\max\{N_t,\epsilon\})}{\log(B/\epsilon)},0,1
\right).
\end{equation}
The final expression is evaluated once on sealed ID-test data, with OOD metrics
reported when available. Its gold score is $R^2$ on the sealed ID test. The
task is solved when pointwise relative error is at most $0.1$ everywhere (and
the prediction is exact when the target is zero).

\paragraph{HillClimbBench-CPU architecture research.}
We select 50 research papers whose central architectural contribution can be
implemented, trained, evaluated, and iteratively improved within a CPU-only
budget, and construct one task from each paper. We run two independent
trajectories per task. Each task prompt combines a technical paper summary,
research brief, description of how gold evaluation differs from proxy
evaluation, evaluation contract, and reference implementation. The agent tests
architectural hypotheses while the host fixes the data, targets, loss,
optimizer, training budget, evaluation seeds, and
resource limits. Every candidate is trained from scratch on one CPU thread.

The 50 source papers are:
\begin{enumerate}
  \item \emph{Can Recurrent Neural Networks Warp Time?} \citep{tallec_warp_time_2018};
  \item \emph{An Intriguing Failing of Convolutional Neural Networks and the CoordConv Solution} \citep{liu_coordconv_2018};
  \item \emph{Deep Lattice Networks and Partial Monotonic Functions} \citep{you_deep_lattice_2017};
  \item \emph{Deep Sets} \citep{zaheer_deep_sets_2017};
  \item \emph{Learning Nonlinear Operators via DeepONet Based on the Universal Approximation Theorem of Operators} \citep{lu_deeponet_2021};
  \item \emph{Dynamical Isometry and a Mean Field Theory of RNNs: Gating Enables Signal Propagation in Recurrent Neural Networks} \citep{chen_dynamical_isometry_2018};
  \item \emph{E(n) Equivariant Graph Neural Networks} \citep{satorras_egnn_2021};
  \item \emph{Fourier Features Let Networks Learn High Frequency Functions in Low Dimensional Domains} \citep{tancik_fourier_features_2020};
  \item \emph{Full-Capacity Unitary Recurrent Neural Networks} \citep{wisdom_full_capacity_2016};
  \item \emph{How Powerful are Graph Neural Networks?} \citep{xu_powerful_gnn_2018};
  \item \emph{Graph Attention Networks} \citep{velickovic_graph_attention_2017};
  \item \emph{Grid Long Short-Term Memory} \citep{kalchbrenner_grid_lstm_2015};
  \item \emph{Group Equivariant Convolutional Networks} \citep{cohen_group_equivariant_2016};
  \item \emph{Learning Latent Permutations with Gumbel-Sinkhorn Networks} \citep{mena_gumbel_sinkhorn_2018};
  \item \emph{HiPPO: Recurrent Memory with Optimal Polynomial Projections} \citep{gu_hippo_2020};
  \item \emph{Hyena Hierarchy: Towards Larger Convolutional Language Models} \citep{poli_hyena_2023};
  \item \emph{HyperNetworks} \citep{ha_hypernetworks_2016};
  \item \emph{Independently Recurrent Neural Network (IndRNN): Building A Longer and Deeper RNN} \citep{li_indrnn_2018};
  \item \emph{Interaction Networks for Learning about Objects, Relations and Physics} \citep{battaglia_interaction_networks_2016};
  \item \emph{The Unreasonable Effectiveness of the Forget Gate} \citep{westhuizen_forget_gate_2018};
  \item \emph{Janossy Pooling: Learning Deep Permutation-Invariant Functions for Variable-Size Inputs} \citep{murphy_janossy_pooling_2018};
  \item \emph{KAN: Kolmogorov-Arnold Networks} \citep{liu_kan_2024};
  \item \emph{Legendre Memory Units: Continuous-Time Representation in Recurrent Neural Networks} \citep{voelker_lmu_2019};
  \item \emph{Resurrecting Recurrent Neural Networks for Long Sequences} \citep{orvieto_lru_2023};
  \item \emph{Long Short-Term Memory} \citep{hochreiter_lstm_1997};
  \item \emph{Mamba: Linear-Time Sequence Modeling with Selective State Spaces} \citep{gu_mamba_2023};
  \item \emph{Memory Networks} \citep{weston_memory_networks_2014};
  \item \emph{MinimalRNN: Toward More Interpretable and Trainable Recurrent Neural Networks} \citep{chen_minimalrnn_2017};
  \item \emph{Hopfield Networks is All You Need} \citep{ramsauer_hopfield_2020};
  \item \emph{Neural Arithmetic Units} \citep{madsen_neural_arithmetic_units_2020};
  \item \emph{Neural Ordinary Differential Equations} \citep{chen_neural_ode_2018};
  \item \emph{Neural Processes} \citep{garnelo_neural_processes_2018};
  \item \emph{Neural Turing Machines} \citep{graves_neural_turing_2014};
  \item \emph{OptNet: Differentiable Optimization as a Layer in Neural Networks} \citep{amos_optnet_2017};
  \item \emph{Parallelizing Legendre Memory Unit Training} \citep{chilkuri_parallel_lmu_2021};
  \item \emph{PDE-Net: Learning PDEs from Data} \citep{long_pde_net_2017};
  \item \emph{Perceiver IO: A General Architecture for Structured Inputs \& Outputs} \citep{jaegle_perceiver_io_2021};
  \item \emph{Rethinking Attention with Performers} \citep{choromanski_performer_2020};
  \item \emph{Phased LSTM: Accelerating Recurrent Network Training for Long or Event-based Sequences} \citep{neil_phased_lstm_2016};
  \item \emph{Pointer Networks} \citep{vinyals_pointer_networks_2015};
  \item \emph{Recurrent Highway Networks} \citep{zilly_recurrent_highway_2016};
  \item \emph{Efficiently Modeling Long Sequences with Structured State Spaces} \citep{gu_s4_2021};
  \item \emph{Orthogonal Recurrent Neural Networks with Scaled Cayley Transform} \citep{helfrich_scaled_cayley_2017};
  \item \emph{Implicit Neural Representations with Periodic Activation Functions} \citep{sitzmann_siren_2020};
  \item \emph{Object-Centric Learning with Slot Attention} \citep{locatello_slot_attention_2020};
  \item \emph{Spatial Transformer Networks} \citep{jaderberg_spatial_transformer_2015};
  \item \emph{TabNet: Attentive Interpretable Tabular Learning} \citep{arik_tabnet_2019};
  \item \emph{Less is More: Recursive Reasoning with Tiny Networks} \citep{jolicoeur_martineau_tiny_recursive_2025};
  \item \emph{Universal Transformers} \citep{dehghani_universal_transformer_2018}; and
  \item \emph{Zoology: Measuring and Improving Recall in Efficient Language Models} \citep{arora_zoology_2023}.
\end{enumerate}

The proxy evaluator measures both in-distribution performance and a
task-specific structural extrapolation split. Each task fixes in advance a
scoring function that combines these measurements into a scalar proxy metric
$u_t\in[0,1]$, with one denoting the task's declared ideal. Only valid candidates
whose proxy metric exceeds that of the retained incumbent replace it. For the
frozen reference metric $u_0$, proxy quality is
$q_t=\operatorname{clip}((u_t-u_0)/(1-u_0),0,1)$.
After selection, the retained architecture and task reference are retrained on
the same five gold-evaluation seeds under harder structural conditions. The task's
gold evaluator maps each run to a score in $[0,1]$, and the submission's
gold score is the mean across its five runs. A submission meets the
benchmark's success criterion when the lower endpoint of the two-sided 95\%
Student-$t$ confidence interval over the five paired candidate-minus-reference
scores is positive. In the policy comparison, a win is a paired evaluation in
which the trained-policy submission has a higher gold score than the matched
untrained-MIRA submission.

\subsection{Evaluation}

We report the proxy best-so-far quality $\bar q_t$ and the gold score
separately. Comparisons use matched task instances and interaction budgets;
confidence intervals use a task-clustered bootstrap.

\section{MIRA Implementation}
\label{app:mira-implementation}

This appendix describes how MIRA constructs each decision context, executes the
resulting work order, and delegates evaluation and submission to the task
environment.

\subsection{Context Construction}

The outer-loop context is rebuilt at every iteration. Its fixed portion contains
the task statement, a compact workspace summary, the latest proxy evaluation,
a structured summary of the immediately preceding work episode and a pointer
to its handover when available, and the remaining budget. The full episode
record and handover text are not copied automatically. Instead, the fixed
curator has read-only access to repository files, diffs, commit history, and
prior work records, from which it selects a bounded set of decision-relevant
evidence. The
fixed portion and this selected evidence form the new decision context.

The policy is instructed to select a bounded objective or coherent experiment
set while considering its information value, cost, remaining time, and risk to
the current solution. Its work order states the objective, relevant evidence,
resource guidance, acceptance criteria, and stopping condition. The same
structure is used for theorem proving, with experiments replaced by proof
construction and verification.

\subsection{Decision and Execution}

A fresh executor in the Codex harness receives the work order, its selected repository
references, and the stable environment instructions. It may inspect and modify
the workspace, run tests and authorized tools, and evaluate candidates. It
validates substantive work, commits useful artifacts and evidence, and leaves a
concise handover. Recorded failures and contradictory results remain available
to later decisions through the repository.

The repository---including the current candidate, experiments, results,
analyses, and handovers---persists across iterations; the model contexts do not.
Any proxy evaluation produced after execution becomes evidence for the next
decision. When an environment maintains an incumbent, it protects the strongest
validated artifact against later regressions.

\subsection{Environment-Controlled Evaluation and Submission}

The environment adapter supplies the task-specific prompt, tools, proxy
feedback, reward mapping, submission artifact, and terminal evaluator. It also
determines who authorizes submission: Codex may submit under inner-loop control,
the policy chooses between further investigation and submission under outer-loop
control, and the environment evaluates only after research ends under external
control.

A run ends when a model-visible success condition is met, submission is
authorized, or the iteration or wall-clock budget is exhausted. Gold
evaluation remains outside both model contexts and is applied only to the
selected repository artifact. Environments that retain an incumbent may restore
it before final evaluation.

\section{Semantic Exploration Graphs and Strategic Inertia}
\label{app:semantic-graphs}

We construct the exploration graphs after each run has terminated. The graphs
are used only for analysis and are never shown to the evaluated system.

\paragraph{Comparable evidence projection.}
We project every run into the same chronological evidence format: the common
task statement, text authored by the evaluated models, and the submitted proof.
For MIRA, model-authored text includes the outer-loop policy's work orders and
the executors' reports; for Codex, it includes assistant messages. We exclude
commands, tool outputs, diffs, commits, repository metadata, handovers,
submission reviews, and external evaluations from every condition. The same
projection is applied at each forced-continuation checkpoint, and the same
downstream extraction procedure is applied to every inference system.

\paragraph{Semantic graph extraction.}
A GPT-5.5 pass at temperature zero converts the projected trace into a
graph. A node represents a semantically meaningful proof or
research step, such as an approach, claim, lemma, counterexample, experiment,
candidate, result, or submission. Several raw turns may therefore map to one
node, while one long response may contain several nodes. Nodes record their
status, such as supported, proved, refuted, stalled, abandoned, or submitted.
Typed edges record evidence-supported relationships including refinement,
dependence, support, refutation, pivoting, supersession, and production of a
result. Chronology alone does not create an edge.

The extractor receives the trajectory content but not the method name, model
name, grader feedback, or final grader score. Every node and edge must cite a
source segment in the projected evidence; outputs without valid provenance are
rejected. The extracted nodes are ordered by first appearance in the trace. For
visualization, the submission ancestry is placed on a central lane and other
connected components are assigned separate lanes. Shortened labels and display
coordinates affect only presentation, not the analysis.

\paragraph{Research-direction annotation.}
A separate GPT-5.5 pass at temperature zero groups the fixed semantic
nodes into research directions. A direction is a high-level hypothesis, proof plan,
construction, mechanism, or algorithmic approach. Lemmas, tests, repairs, and
refinements remain in the same direction when they serve the same high-level
plan; a materially different plan receives a new identifier, and a later return
to a previous plan reuses its identifier. This pass may correct the order of
fixed nodes when their provenance establishes a different first-introduction
order, but it may not add, remove, split, or merge nodes. It observes node
content, provenance episode identifiers, and graph relations, but not the
method or model name, final grader score, grader feedback, node status, terminal
submission, graph title, or graph summary.

\paragraph{Strategic inertia.}
Two consecutive non-submission semantic steps form a continuation when they
belong to the same annotated research direction and a pivot otherwise. For
graph $\mathcal{G}$, we define
$I(\mathcal{G})=N_{\mathrm{continue}}/
(N_{\mathrm{continue}}+N_{\mathrm{pivot}})$.
The root and terminal submission are excluded. Graphs with fewer than two
non-submission nodes have no transitions and receive no inertia value. The
statistic is descriptive: high inertia can sustain a promising direction or
prolong a local minimum, while low inertia can expose alternatives or reflect
premature switching.

For the natural GPT-5.5 comparison, the analysis contains 30 problems, three
trials, and three inference systems. We compute inertia for each run, average
trials within each problem, and obtain 95\% confidence intervals by resampling
problems with 10,000 bootstrap draws. Pairwise contrasts match runs by problem
and trial, average the paired differences within each problem, and again
bootstrap problems. Natural and forced-continuation runs are analyzed
separately.

The four visualizations below are the problems on which a gap remains after
output-token-matched continuation. They show the first natural-stopping trial
and use the same extraction and layout procedure for all three inference
systems.

\begin{figure}[H]
  \centering
  \includegraphics[width=\textwidth]{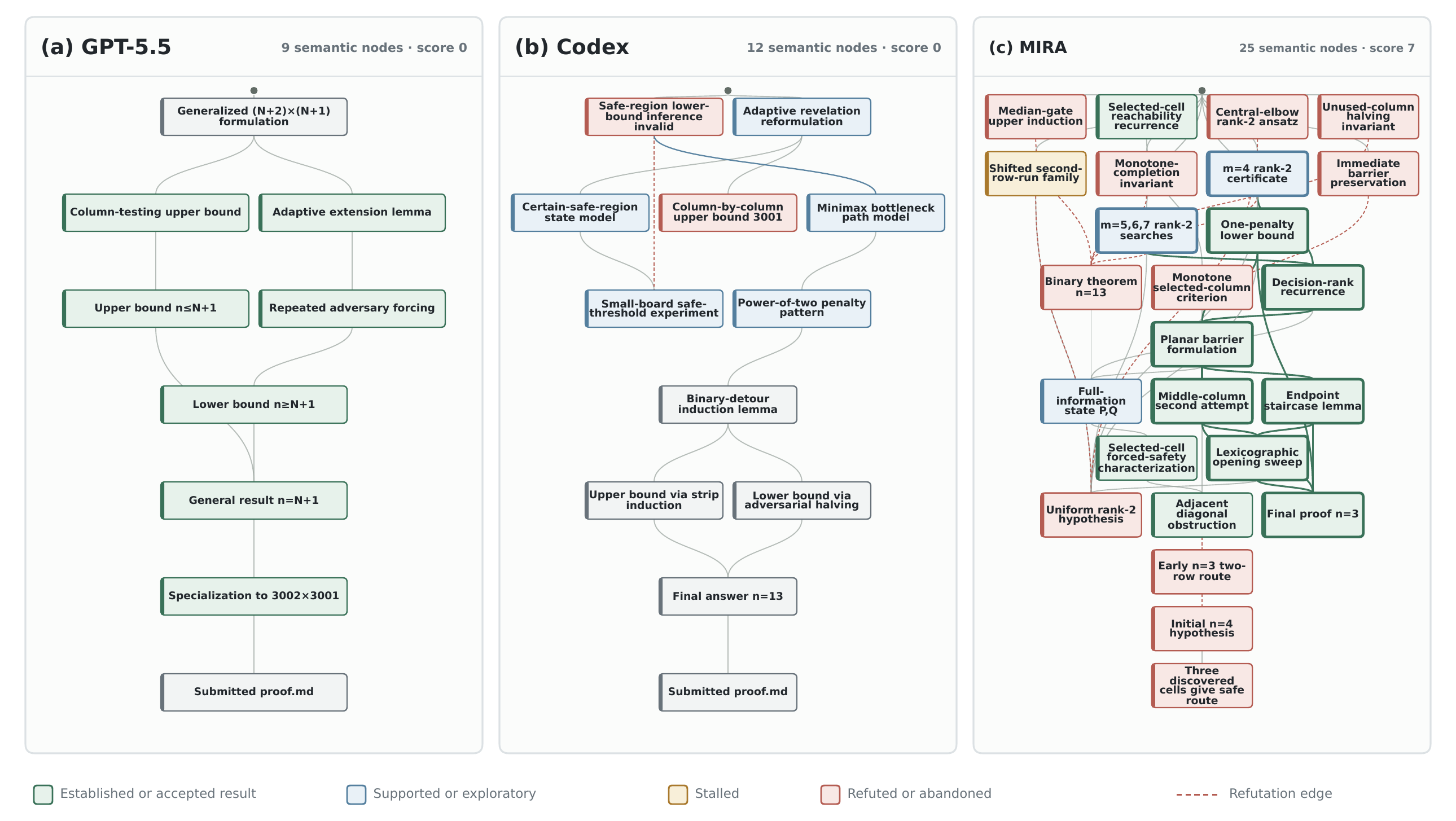}
  \caption{\textbf{MIRA explores multiple proof directions and reaches the
  full-credit proof on PB-Advanced-023.} Panels show (a) GPT-5.5, (b) Codex,
  and (c) MIRA; all conditions use GPT-5.5. Dashed red edges denote
  refutations.}
  \label{fig:imoproofbench-pb23}
\end{figure}

\begin{figure}[H]
  \centering
  \includegraphics[width=\textwidth]{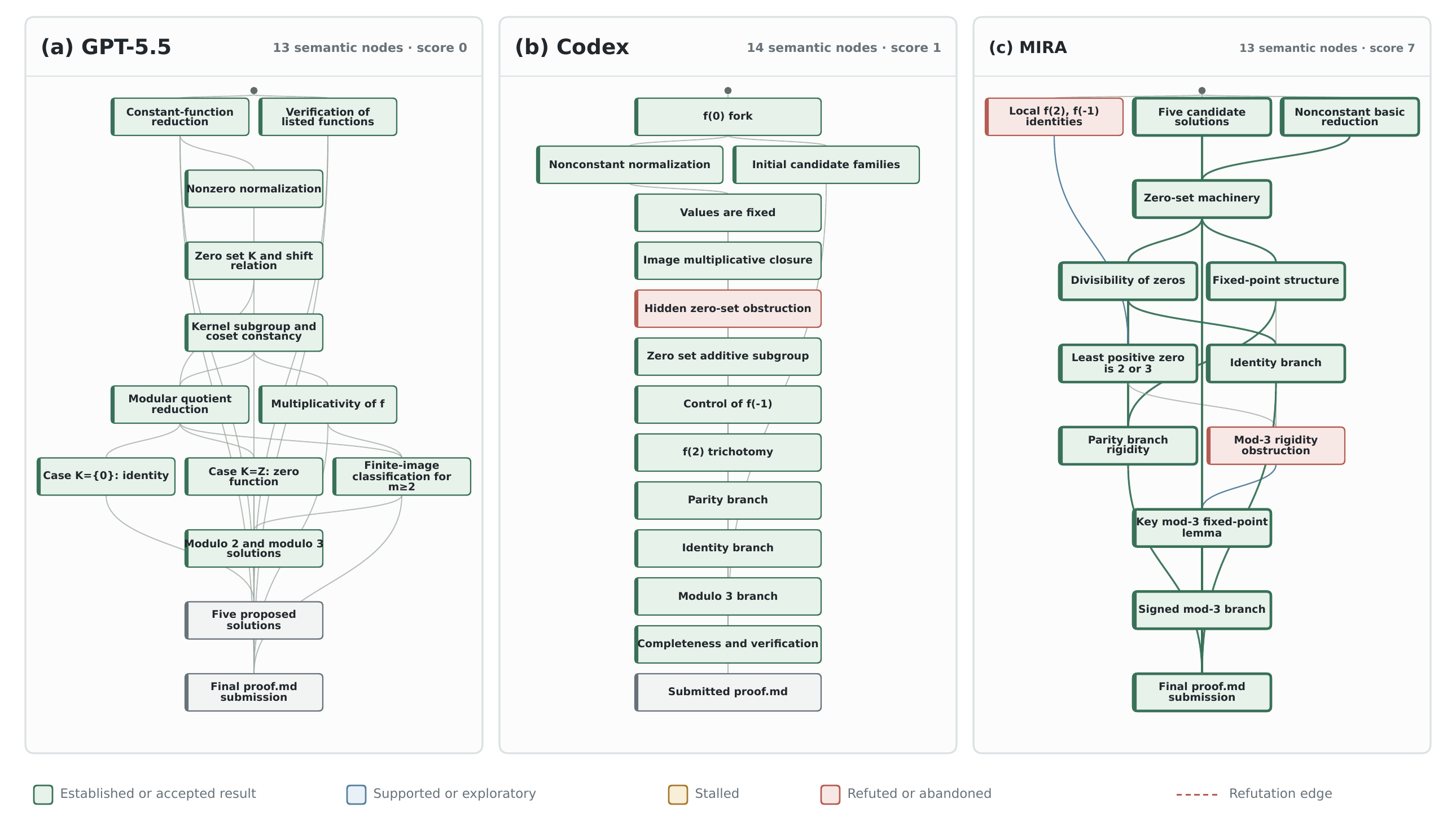}
  \caption{\textbf{MIRA develops and combines multiple proof branches into a
  complete classification on PB-Advanced-006.} GPT-5.5 and Codex
  identify candidate functions but leave central steps unsupported or
  incomplete, whereas MIRA produces a complete proof receiving 7/7. Systems and notation
  follow Fig.~\ref{fig:imoproofbench-pb23}.}
  \label{fig:imoproofbench-pb06}
\end{figure}

\begin{figure}[H]
  \centering
  \includegraphics[width=\textwidth]{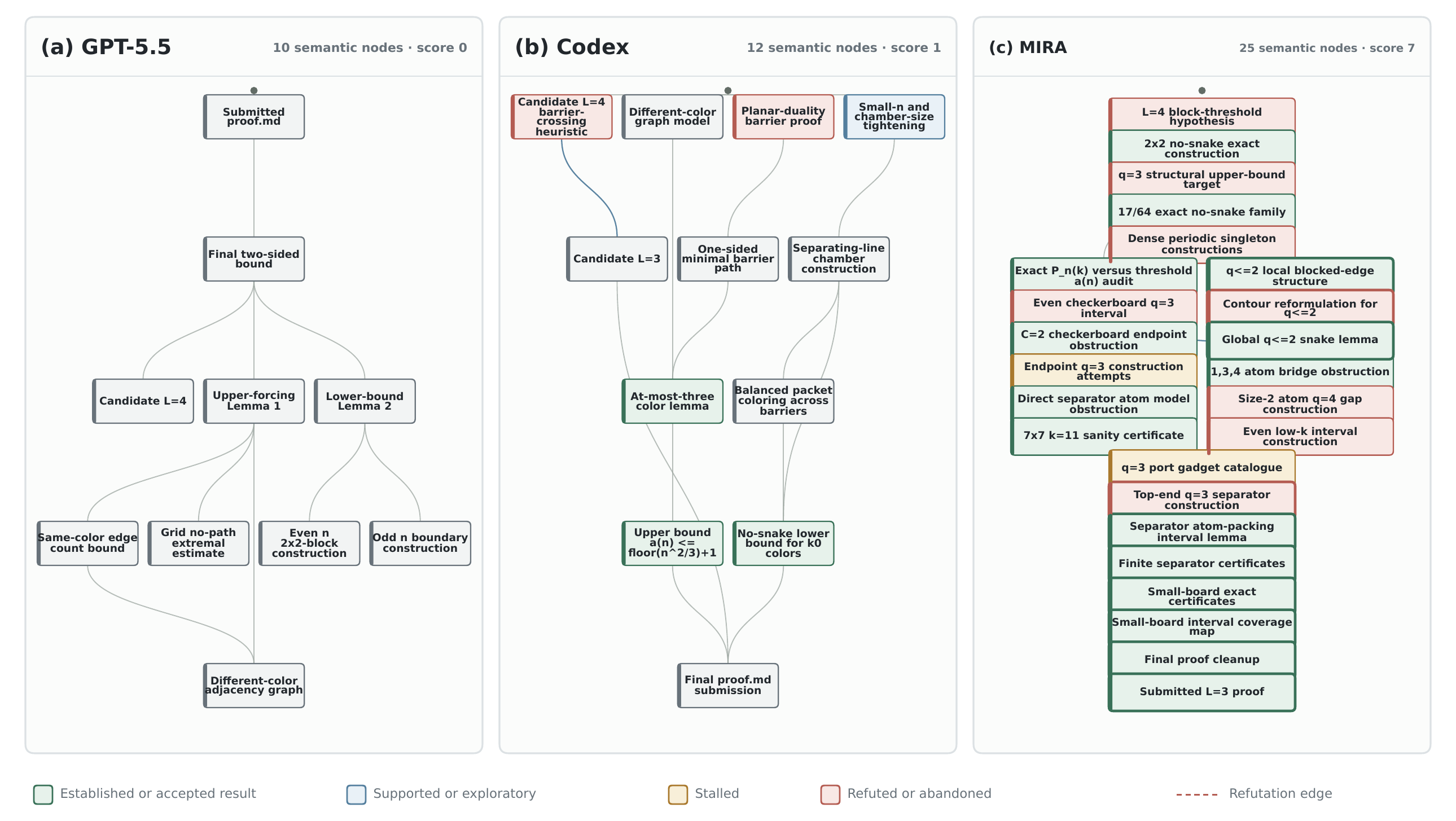}
  \caption{\textbf{MIRA redirects from a refuted $L=4$ approach and completes
  an $L=3$ proof on PB-Advanced-018.} GPT-5.5 remains within the unsuccessful
  $L=4$ direction, while Codex pivots to $L=3$ but leaves the proof
  incomplete. MIRA develops the redirected approach into a proof receiving 7/7.
  Systems and notation follow Fig.~\ref{fig:imoproofbench-pb23}.}
  \label{fig:imoproofbench-pb18}
\end{figure}

\begin{figure}[H]
  \centering
  \includegraphics[width=\textwidth]{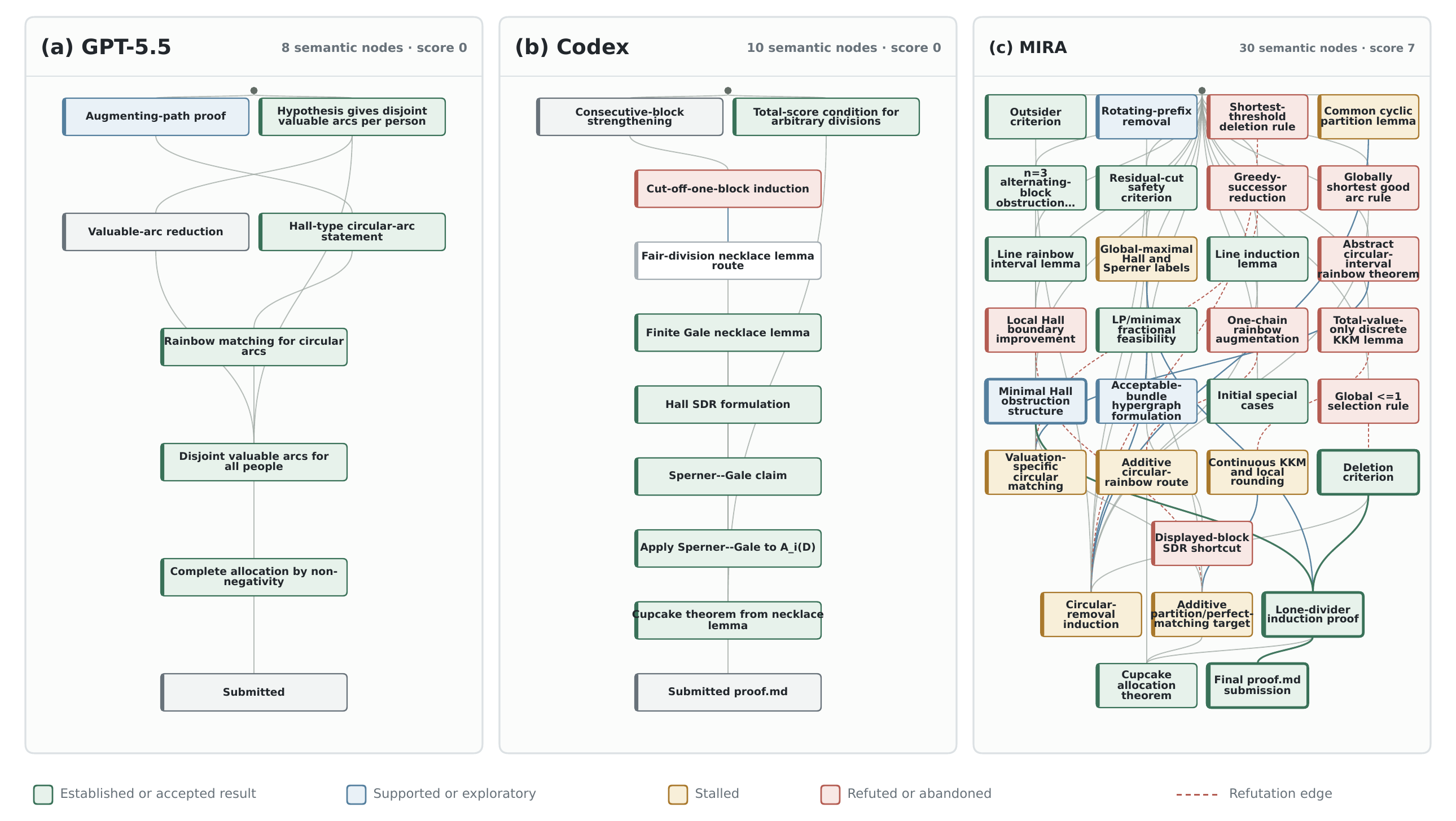}
  \caption{\textbf{MIRA explores competing combinatorial directions and
  synthesizes a correct proof on PB-Advanced-030.} GPT-5.5 and Codex
  terminate in unsuccessful proof directions, whereas MIRA explores allocation,
  deletion, and induction arguments before reaching a complete 7/7 proof.
  Systems and notation follow Fig.~\ref{fig:imoproofbench-pb23}.}
  \label{fig:imoproofbench-pb30}
\end{figure}

\section{RMT Architecture Identified by MIRA}
\label{app:rmt-analysis}

\paragraph{Matched search protocol.}
We run one 24-hour GPT-5.5 campaign for each orchestration condition. All three
campaigns use the same task reference architecture, proxy evaluator, compute
budget, and parameter limit. Proxy evaluations guide research, and the final
submitted architecture is scored separately by the gold evaluator.

\paragraph{Selected architecture.}
An ordinary RMT stores each token's residual state as a matrix
$X_\ell\in\mathbb{R}^{D_k\times D_v}$. Attention and MLP branches read this
state through a lower-rank interface, produce a matrix update
$\Delta X_\ell^{(b)}$, and add it uniformly as
$X_{\ell,\mathrm{out}}^{(b)}=X_{\ell,\mathrm{in}}^{(b)}+
\Delta X_\ell^{(b)}$ for $b\in\{\mathrm{attn},\mathrm{mlp}\}$.
MIRA's final model retains this matrix state and low-rank read/write interface,
uses a GELU-gated linear unit (GEGLU) feed-forward block, and introduces a
separate learned gate for
each branch and layer:
$X_{\ell,\mathrm{out}}^{(b)}=X_{\ell,\mathrm{in}}^{(b)}+
G_\ell^{(b)}\odot\Delta X_\ell^{(b)}$, where
$G_\ell^{(b)}\in\mathbb{R}^{D_k\times D_v}$.
Every coefficient is initialized to one, so the new update is exactly the
ordinary additive update at initialization. The two gates add only 0.056\% to
the model's parameter count. Unlike a generic GEGLU block, this mechanism
is specific to the matrix residual stream: it learns where each branch can
amplify or suppress a proposed matrix-state write.

MIRA first tested the gate at small scale, then reran a final-scale comparison
matched in geometry, seed, optimizer, and training tokens. Relative to the
ungated GEGLU control, the gated model improved last-token cross-entropy by
0.009 at small scale and by 0.021 at the final scale.
Both comparisons use one seed, and the ungated final-geometry control was not
submitted to the gold evaluator. These ablations support the gate
as a promising mechanism, but they do not isolate it as the cause of the full
gold-loss improvement in Fig.~\ref{fig:rq1-hillclimbs}(b).

The learned gates did not remain near their identity initialization. Attention gates generally
amplify writes, whereas MLP gates generally suppress them, and both vary with
depth. A two-way additive decomposition into row and column effects explains
roughly three quarters of the
variation in the depth-averaged attention map and nearly nine tenths in the MLP
map, with stronger variation along the residual-key axis.
This supports a structured, branch-, depth-, and address-dependent
\emph{write-allocation} interpretation. It does not establish dynamic routing:
the gates are fixed across tokens and examples, and their low exact-map
correlation across layers does not support a clean permanent partition into
``durable'' and ``temporary'' cells.

\section{Loop Transformer Search and Selected Designs}
\label{app:loop-analysis}

\paragraph{Matched search protocol.}
We compare matched 48-hour MIRA and Codex campaigns using GPT-6 Astra. Both
campaigns use the same model-scale and training-token rungs.
At each rung, candidates and the task reference architecture are evaluated on the
same 512-context proxy set with eight recurrent iterations. Intermediate
search decisions use these results; after submission, the final incumbent from
each campaign and the task reference are evaluated on a separate
4,873-context gold set. The proxy and gold results are reported
separately in Fig.~\ref{fig:loop-hillclimb}.

\begin{figure}[t]
  \centering
  \includegraphics[width=\textwidth]
    {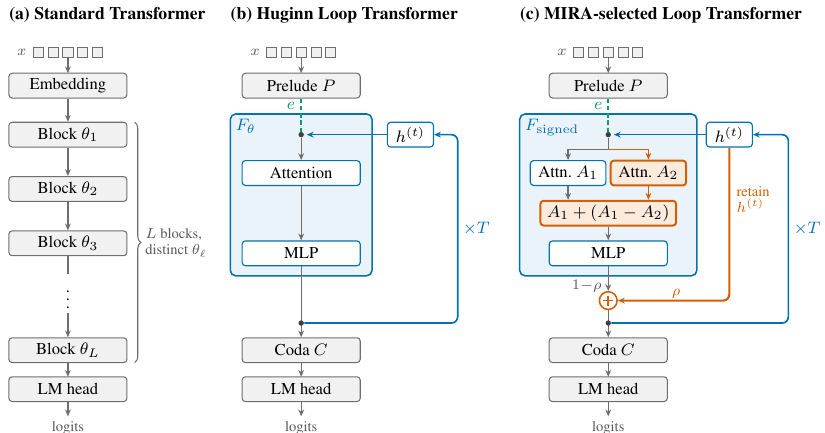}
  \caption{\textbf{Architecture progression for the Loop Transformer study.}
  A standard Transformer obtains depth from blocks with distinct parameters.
  Huginn instead repeatedly applies one weight-tied core while reinjecting the
  input representation. The MIRA-selected architecture retains this recurrent
  structure and adds what the campaign calls signed attention, which contrasts
  two attention reads over shared values. It also interpolates between the
  previous and proposed states at each recurrent iteration---a leaky recurrent
  update. Blue
  denotes weight-tied recurrence, dashed green denotes
  input reinjection, and orange denotes components introduced by MIRA.}
  \label{fig:loop-architecture}
\end{figure}

\paragraph{MIRA search trajectory.}
MIRA begins by hypothesizing that repeated applications of the shared core
overwrite useful state. It compares a learned adaptive retention gate with a
constant interpolation between the previous and proposed states. The learned
gate does not outperform this control, so subsequent candidates retain the
simpler leaky update. After explicit fast and slow states, iteration-history
mechanisms, and other routing alternatives produce small or inconsistent gains,
MIRA hypothesizes that repeated positive attention may reinforce information
that later recurrences struggle to remove. It tests the resulting subtractive
attention mechanism against a matched positive-mixture control, then promotes
it only after the advantage persists across seeds and model scales.
The selected design combines what the campaign calls ``signed attention''
with a leaky recurrent update (Fig.~\ref{fig:loop-architecture}). If the two
branches produce attention reads $A_1$
and $A_2$ over shared values, the attention module returns
$A_1+(A_1-A_2)$. Each branch separately uses ordinary softmax attention;
the subtraction creates effective positive and negative coefficients over the
shared values. The second branch copies the first branch's query
and key parameters at initialization, making the additional contrastive term
exactly zero before training. Training can then break this symmetry, allowing
the effective coefficients to suppress as well as reinforce value components.
After the block proposes $F(h^{(k)})$ from recurrent state $h^{(k)}$, the
leaky update mixes the previous and proposed states,
$h^{(k+1)}=\rho h^{(k)}+(1-\rho)F(h^{(k)})$, with fixed $\rho=0.25$. The
attention variant and leaky update therefore play complementary roles: one supports
subtractive routing, while the other limits how completely a recurrent step
can overwrite its input. Applied in a Loop Transformer, this mechanism
resembles a fixed-coefficient variant of Differential
Attention~\citep{ye_differential_2025}. It uses two full query--key branches
initialized identically and the fixed mixture $2A_1-A_2$.

\paragraph{Comparison.}
Codex selects a tangent-state update with local/global attention
routing and learned write gates on both the attention and MLP branches, a
complementary strategy that regulates how aggressively each recurrent pass
changes the state. MIRA's incumbent improves proxy loss at every
completed scale. Its final proxy NLL is 3.0331, compared with 3.2419 for
Codex and 3.2940 for the task reference. On the separate gold evaluation, its
NLL is 3.0464, compared with 3.2275 for Codex and 3.2653 for the task reference.

\section{Critic Training and Validation}
\label{app:critic-details}

\paragraph{Targets and value contracts.}
All critic variants use the stitched Monte Carlo target $G_t$. Under
proxy-marginal supervision, $G_t$ is the future increase in best-so-far proxy
quality, where $q_t$ is defined for each environment in
App.~\ref{sec:proxy-signals}. The corresponding proxy qualities are
directed-SHD gap closure for BNLearn, acquired-trajectory normalized fit for
Physics Discovery, validation log-NMSE gap closure for SRBench, and
initial-to-ideal proxy-score gap closure for HillClimbBench-CPU.

Under gold-binary supervision, the label $Y\in\{0,1\}$ records whether
the final submission meets the environment's gold success criterion. This
means exact directed recovery for BNLearn; the joint hidden-trajectory and
explanation criterion for Physics Discovery; the ID-test pointwise-accuracy
criterion for SRBench; and a positive lower endpoint of the paired 95\%
confidence interval for improvement over the reference architecture for
HillClimbBench-CPU. These are the same success criteria specified in
App.~\ref{sec:environments}; the policy-comparison win rate is not a critic
target.

\paragraph{Critic parameterizations and objectives.}
For proxy-marginal targets, a scalar critic minimizes
$\mathcal{L}_{\mathrm{scalar}}(\phi)=M^{-1}\sum_{m=1}^{M}
(V_\phi(o_m)-G_m)^2$.
For the categorical critics, let $z_1,\ldots,z_K$ be the ordered support. If
$z_j<G_t<z_{j+1}$, we set
$y_j(G_t)=(z_{j+1}-G_t)/(z_{j+1}-z_j)$ and
$y_{j+1}(G_t)=(G_t-z_j)/(z_{j+1}-z_j)$, with all other entries zero; if
$G_t=z_j$, then $y_j(G_t)=1$. A categorical
head produces $q_\phi(z_i\mid o)=\operatorname{softmax}(h_\phi(o))_i$, while
the generative critic uses the restricted next-token distribution
$p_\phi(z_i\mid o)$ from Sec.~\ref{sec:value_function}. They minimize
$\mathcal{L}_{\mathrm{cat}}(\phi)=M^{-1}\sum_{m=1}^{M}
D_{\mathrm{KL}}(y(G_m)\,\|\,p_\phi(\cdot\mid o_m))$, replacing $p_\phi$ with
$q_\phi$ for the categorical head. For the generative critic, the
implementation gathers the next-token log probabilities of the support labels,
renormalizes them across that support, and evaluates the KL objective in
Sec.~\ref{sec:critic-objective}; no probability mass outside the support enters
the critic loss.

Under gold-binary supervision, the scalar critic outputs a single
logit $\ell_\phi(o)$ and represents value as
$V_\phi(o)=\sigma(\ell_\phi(o))$. It is trained with binary cross-entropy with
logits,
\begin{equation}
  \mathcal{L}_{\mathrm{bin}}(\phi)
  =-\frac{1}{M}\sum_{m=1}^{M}
  \left[Y_m\log\sigma(\ell_\phi(o_m))
  +(1-Y_m)\log\!\left(1-\sigma(\ell_\phi(o_m))\right)\right].
\end{equation}
The gold-binary generative critic instead restricts the next-token
distribution to the support $\{0,1\}$ and uses the corresponding one-hot
target. Token-level variants repeat the decision return at each causal token
prefix within the action; decision-level variants forecast once from the curated
observation.

\paragraph{Target and forecast validation.}
Proxy-marginal returns are validated and clipped to
$G_t\in[0,1-\bar q_{t-1}]$. Before bootstrapping, temporal-difference targets,
or advantage estimation, proxy-marginal forecasts must be finite and are
clipped to the same interval. For a generative critic,
$V_\phi(o_t)=\sum_i p_\phi(z_i\mid o_t)z_i$. Gold-binary targets lie in
$\{0,1\}$, and both the sigmoid head value and the expected value of the
two-point generative distribution lie in $[0,1]$.

\paragraph{Pretraining and target-environment adaptation.}
We pretrain Qwen3.6-27B for one epoch with batch size 8, microbatch size 1,
maximum sequence length 262,144, and seed 0. The optimizer is AdamW with
learning rate $10^{-5}$, weight decay 0.01, gradient clipping at 1.0, 5\%
linear warmup, and cosine decay to 10\% of the peak learning rate. Training
uses bfloat16 and the LoRA configuration in
Table~\ref{tab:mira-ac-hparams}. The
value-forecasting experiments use four H200s; the four-environment critic used
to initialize the general MIRA-AC policy uses eight H200s.
For target-environment adaptation, we evaluate support sizes
$\{0,1,5,10,25,50\}$ over five paired support/query splits with 50 query
episodes per split. Adaptation makes one episode-ordered pass using AdamW at
learning rate $10^{-5}$ and LoRA dropout 0.05 on one H200.

For each target environment and paired support/query split, we first average
the single-source pretraining condition across source environments. We then
report averages over target environments and splits with 95\% confidence
intervals from a paired bootstrap over support/query splits.

\paragraph{Generative critic prompts.}
The generative critic receives the same curated decision-scoped conversation
as the actor, followed by the value query in Fig.~\ref{fig:critic-prompt}.
When forecasting at termination, the query marks the normal research budget as
exhausted.

\begin{figure}[H]
\centering
\begin{minipage}{0.96\linewidth}
\fontsize{5.5}{6.5}\selectfont
\begin{verbatim}
Value contract: <ENVIRONMENT-SPECIFIC VALUE-CONTRACT NAME>
<ENVIRONMENT-SPECIFIC VALUE-CONTRACT DESCRIPTION>

Observable value state:
Current best normalized quality before this decision: <CURRENT BEST NORMALIZED QUALITY>
Total decision horizon: <TOTAL DECISION HORIZON>
Decisions remaining including the current decision: <DECISIONS REMAINING>
<OPTIONAL TERMINAL-PHASE STATUS>

You are estimating the value of the research state in the conversation above.

Predict the future increase in the final best solution quality, using the normalized 0-to-1 quality scale.

0.0 means no further improvement over the best solution already found.
Higher values mean larger expected improvement in final best solution quality.

Use the raw metrics, errors, tests, benchmark scores, and experiment results as evidence, but do not answer in raw environment units.

On this normalized 0-to-1 scale, how much additional improvement over the current best solution do you expect from here?
A. 0.0
B. 0.1
C. 0.2
D. 0.3
E. 0.4
F. 0.5
G. 0.6
H. 0.7
I. 0.8
J. 0.9
K. 1.0

Answer with only the single letter for your estimate.
\end{verbatim}
\end{minipage}
\caption{\textbf{Canonical prompt for eliciting proxy-marginal generative
value forecasts.} Environment-specific fields specify the value contract,
current normalized quality, and remaining budget. Although the model emits a
single letter, its next-token probabilities over all labels define the critic's
distribution over remaining return.}
\label{fig:critic-prompt}
\end{figure}

For gold-binary supervision, the environment-specific contract states the
gold success event above, and the proxy-marginal question and eleven-point
support are replaced by:
\begin{center}
\begin{minipage}{0.90\linewidth}
\fontsize{7}{8}\selectfont
\begin{verbatim}
Predict the binary terminal outcome of the complete research episode.

0.0 means the environment's authoritative terminal success condition will not be met.
1.0 means the environment's authoritative terminal success condition will be met.

Use the raw metrics, errors, tests, benchmark scores, experiment results,
current progress, and remaining budget as evidence.

Will this research episode ultimately succeed?
A. 0.0
B. 1.0

Answer with only the single letter for your estimate.
\end{verbatim}
\end{minipage}
\end{center}

\section{MIRA-AC Training Details}
\label{app:implementation-details}

All policy-training runs use Qwen3.6-27B~\citep{qwen_qwen36_27b_2026}
(Qwen-27B) with LoRA adapters. In separate actor--critic configurations, actor
adapters target the standard attention and MLP projections,
mixture-of-experts (MoE) experts, and latent fully connected projections. MIRA-AC targets the
standard attention and MLP projections together with the model's recurrent
input and output projections. Separate-critic variants use distinct policy and
critic adapters. In MIRA-AC, one shared adapter serves both roles, with separate
optimizer states for actor and critic updates; online critic updates modify only
the shared adapter's MLP LoRA parameters. Each critic is initialized from a pretrained checkpoint
matched to its critic parameterization and target environment.
Separate actors start from the untrained MIRA policy. For each single-environment
experiment, MIRA-AC initializes its shared adapter from a generative critic
pretrained on trajectories from that environment under the matching supervision
target. The cross-environment policy instead starts from a proxy-marginal
generative critic pretrained jointly across all four environments, then receives
online actor--critic updates on SRBench, Physics Discovery, and
HillClimbBench-CPU.

Table~\ref{tab:mira-ac-hparams} reports the configuration of the general
cross-environment policy used for the main result. The reported checkpoint is
the epoch-one shared adapter at version 94, after 49 actor updates and 90 critic
updates. When both learners update in one transaction, their updates share a
single published version.

\begin{table*}[t]
\centering
\caption{\textbf{General MIRA-AC training configuration.} Settings are from
the reported epoch-one, version-94 cross-environment run.}
\label{tab:mira-ac-hparams}
\small
\setlength{\tabcolsep}{5pt}
\renewcommand{\arraystretch}{1.08}
\begin{tabularx}{\textwidth}{@{}p{0.22\textwidth}X@{}}
\toprule
Setting & Value \\
\midrule
Base model & Qwen3.6-27B, revision \texttt{6a9e13bd6fc8}; policy thinking enabled and critic thinking disabled \\
Initialization & Weights-only initialization from a proxy-marginal decision-level generative critic pretrained jointly across four environments \\
Actor-training distribution & SRBench, Physics Discovery, and HillClimbBench-CPU; 100 episodes per environment, one epoch, run seed 42 \\
LoRA & Rank 32, scale 64, dropout 0; shared actor--critic adapter; online critic updates restricted to MLP projections \\
Research and sequence limits & 20 research decisions; 262,144-token maximum sequence; 8,192-token maximum generation \\
Policy sampling & Temperature 1.0, top-$p$ 0.95, top-$k$ 20, minimum-$p$ 0, presence penalty 0, repetition penalty 1 \\
Actor optimizer & AdamW, learning rate $10^{-5}$, $\beta=(0.9,0.999)$, weight decay 0.01, gradient clipping 1.0, constant schedule \\
Online critic optimizer & AdamW, learning rate $5\times10^{-6}$, $\beta=(0.9,0.999)$, $\epsilon=10^{-8}$, weight decay 0.01, gradient clipping 1.0; one pass and optimizer step per sealed batch \\
Precision & 32-bit floating-point (FP32) actor optimization and reduction; bfloat16 critic backbone with FP32 logits and targets \\
Return and GAE & Nonnegative marginal improvement in best proxy quality; $\gamma=1$; $\lambda_T=1-1/(\alpha T)$ with $\alpha=20$ \\
Actor objective & $\epsilon_{\mathrm{low}}=0.3$, $\epsilon_{\mathrm{high}}=5.0$, $\tau_A=1$, $\tau_{\mathrm{KL}}=10^{-3}$; normalized by research decision \\
Active episodes & Up to 32 asynchronous episodes; completed slots are refilled immediately; new parameters are adopted at decision boundaries \\
Actor buffer & At least 4 episodes, 64 decisions, and 262,144 trainable tokens; 15-minute fallback after episode and decision minima \\
Critic buffer & At least 2 episodes, 32 decisions, and 131,072 prompt tokens; 7.5-minute fallback after episode and decision minima \\
Publication & When ready, critic and actor updates execute in that order within one atomic transaction, followed by one shared-version publication \\
Compute & 16 H200s for actor--critic inference and optimization, plus 16 H200s serving frozen executors \\
\bottomrule
\end{tabularx}
\end{table*}

We train one policy per configuration. Each environment-specific run makes one
pass over a fixed ordered set of 100 episodes. The cross-environment run uses
300 episodes: 100 from each of its three actor-training environments. The
scientific environments use 100 tasks with one rollout each;
HillClimbBench-CPU uses 50 tasks with two rollouts each. Within each environment,
the task order is deterministically shuffled from run seed 42; environments are
interleaved by equal-weight deficit round robin, and generation seeds are tied
to logical decision identifiers.

The actor packs each sealed learner batch dynamically into microbatches of at
most 262,144 tokens and takes one optimizer step after accumulating their
gradients. The online critic uses microbatch size 1 across four GPUs and takes
one optimizer step after a complete pass without replacement through each sealed
batch. The
proxy generative critic uses support $\{0.0,0.1,\ldots,1.0\}$; the gold-binary
critic forecasts the gold-binary outcome on support $\{0,1\}$. We use $\gamma=1$
and exact Monte Carlo critic targets.
Actor advantages use the episode-length-adaptive GAE configuration in
Table~\ref{tab:mira-ac-hparams}.

The implementation uses the token-level importance ratio and range mask from
Sec.~\ref{sec:actor-objective}, together with the decision-normalized actor
objective in Eq.~\ref{eq:actor-loss}.
Tokens outside the ratio interval
are excluded from the policy-gradient term but remain in the squared-log-ratio
penalty. The trainable span contains only the complete sampled decision
response---including its emitted reasoning and end-of-turn token---whether it
ends in a work order or a termination decision. Prompts, curated
observations, tool calls, and executor outputs are masked.

The buffered schedule maintains at most 32 active episodes and immediately
refills completed slots while budget remains. Every completed episode enters
both buffers, which seal independently according to
Table~\ref{tab:mira-ac-hparams}; remaining examples form final underfilled
batches. In MIRA-AC, all ready role-specific updates are applied within one
atomic transaction. The critic update precedes the actor update, recorded values remain
detached during actor training, and the resulting shared adapter is published
once. Live episodes adopt it only at the next meta-reasoning boundary.

Value-forecasting critic pretraining uses four H200 GPUs. The four-environment
critic used to initialize the general policy instead uses eight H200 GPUs. The
reported general-policy run uses two eight-H200 nodes for actor--critic
inference and optimization and a separate two-node, 16-H200 pool for frozen
executors.

\paragraph{Context-cost accounting.}
The context-cost analysis in Fig.~\ref{fig:mira-context-reset} compares
MIRA-AC with fixed-trajectory replays that preserve the recorded work orders,
executor actions, and tool interactions. The MIRA-AC baseline counts policy
input together with exact or imputed executor input. The no-compaction replay
carries all prior cross-iteration history into each later model call, while the auto-compaction
projection caps that history at 200k tokens. The displayed range captures how
often the carried history is reread. Its lower endpoint adds the history once
during each active stage---curation, work-order selection, and execution. Its
upper endpoint follows a sequential tool loop in which the updated context is
read after each recorded curation request and executor turn. Frozen-curator
base prompts and outputs, critic inputs, compaction-summary costs, and
behavioral changes are not modeled. The comparison therefore concerns
rollout-generation cost, not optimization cost or sample efficiency.

\section{Trajectory Stitching Across Artificial Boundaries}
\label{app:trajectory_stitching}

MIRA reconstructs a fresh curated context at every research decision. Context-window
limits, compaction, sequence packing, and learner batching may further split a
single logical episode into multiple samples. None of these boundaries is an
environment termination. Treating them as terminal either removes future TD
residuals from earlier advantages or places later rewards at the wrong temporal
distance.

Let a logical episode $\tau=(\xi_1,\ldots,\xi_T)$ be partitioned into stored
samples $\mathcal S_1,\ldots,\mathcal S_K$. The partition does not alter
the termination indicator: $d_t=1$ only when the environment truly terminates.
Each stored transition contains $o_t$, $a_t$, $r_t$, $o_{t+1}$, actor behavior
log probabilities and version $\theta^{b_t}$, and the critic version $c_t$
assigned to the transition. For that critic version, the system records both the current baseline
$v_t=V_{\theta^{c_t}}(o_t)$ and its successor bootstrap
$\widetilde v_{t+1}^{(c_t)}=V_{\theta^{c_t}}(o_{t+1})$.

An asynchronous episode may adopt a different critic before decision $t+1$.
At a nonterminal boundary where this occurs, we use the bridge $B_t$ from
Sec.~\ref{sec:batch_online}, and set $B_t=0$ otherwise. Terminal successor
values and bridges are zero. The implemented residual and backward recurrence
follow Sec.~\ref{sec:batch_online} and Eq.~\ref{eq:decision-gae}.
The bridge aligns the outgoing critic's successor baseline with the baseline
recorded by the critic version assigned to the next decision.

After this episode-level pass, $\hat A_t$ is scattered back to its original
sample and token span for actor training. The critic target is the Monte Carlo
return $G_t$, computed separately over the same reconstructed episode. The
recorded value forecasts are treated as constants during the actor update, so
the policy loss does not differentiate through the shared critic baseline.

\paragraph{Partition invariance.}
For any partition of a fixed logical episode into stored samples, stitching
produces the same actor advantages and Monte Carlo critic targets as processing
the unpartitioned episode, provided the logged transitions, critic versions, and
successor bootstraps are preserved. Stitching restores the same ordered rewards,
terminal indicators, baselines, and cross-version bridges. A backward induction
on the recurrence therefore yields the same $\hat A_t$ at every decision, while
every reward remains in the same Monte Carlo return.

In our implementation, $\gamma=1$ and the critic is trained on the
exact Monte Carlo return $G_t=\sum_{u=t}^{T}r_u$. Actor advantages use
episode-length-adaptive $\lambda_T=1-1/(\alpha T)$, giving
$\hat A_t=\delta_t+\lambda_T(1-d_t)(\hat A_{t+1}+B_t)$.
The full Monte Carlo return receives weight
$\lambda_T^{T-1}\geq1-1/\alpha$ at the earliest decision and greater weight at
later decisions. The bridges correct baseline-coordinate changes but do not
alter the Monte Carlo critic target.
Stitching remains necessary: $G_t$ must include rewards from all later MIRA
iterations, even when those iterations occupy different contexts, samples, or
learner batches.

The governing invariant is that only true environment termination may stop
value bootstrapping; infrastructure boundaries never do.

\section{Additional MIRA-AC Results}
\label{app:mira-ac-results}

\begin{figure*}[p]
  \centering
  \includegraphics[
    width=\textwidth,
    height=0.92\textheight,
    keepaspectratio
  ]{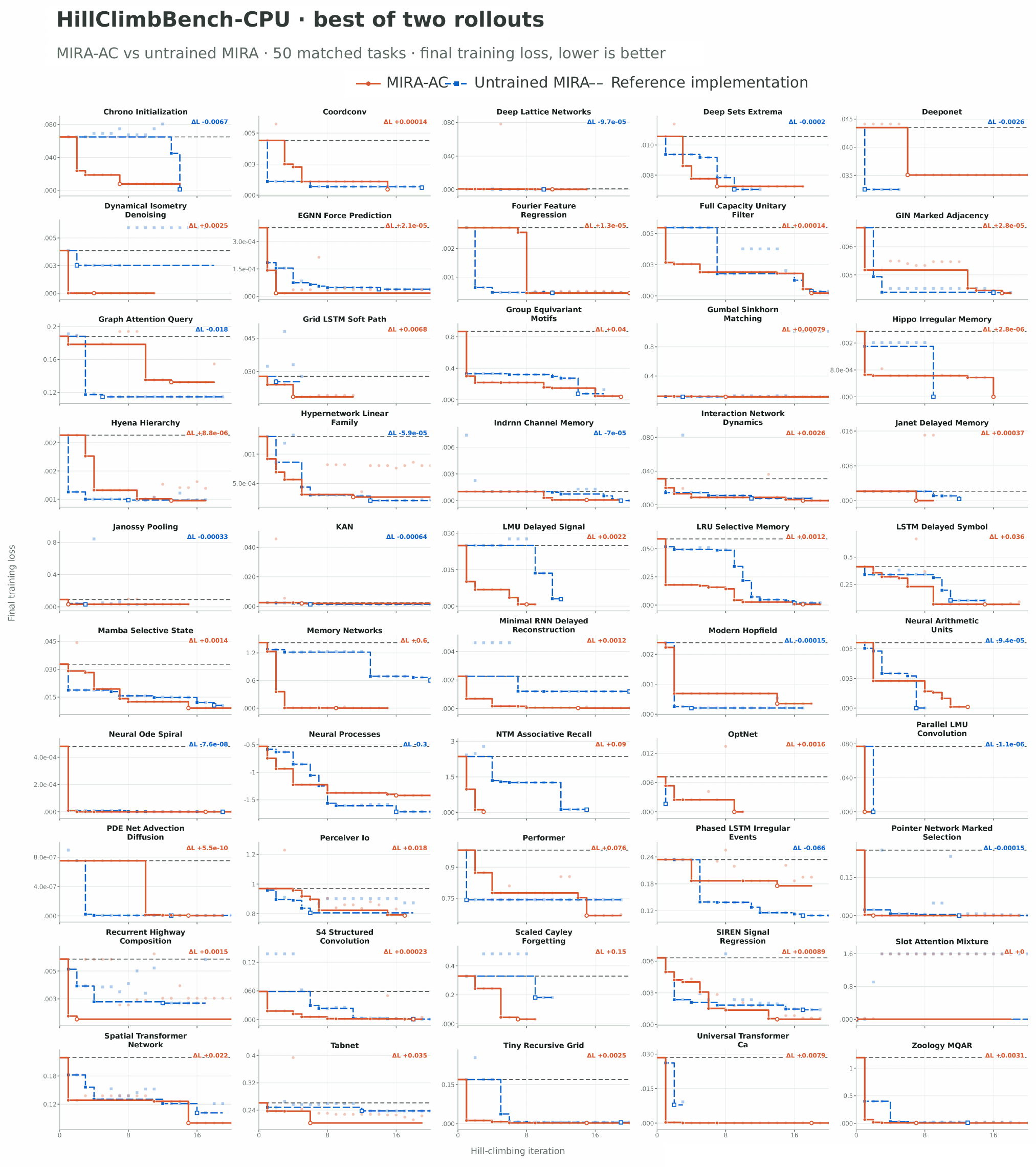}
  \caption{\textbf{MIRA-AC achieves a lower best-observed loss than untrained
  MIRA on 33 of 50 HillClimbBench-CPU tasks.} Each panel shows, for each policy,
  the better of two matched rollouts as measured by best-observed raw loss.
  Faint points show successful evaluations, staircases track the best
  loss observed so far, and the horizontal dashed line marks the reference
  implementation.}
  \label{fig:mira-ac-hillclimbbench-per-task}
\end{figure*}

\section{Behavioral Analysis of MIRA-AC}
\label{app:mira-ac-behavior}

This appendix examines the changes we observe in the model's meta-reasoning
and scientific behavior as a result of training. We compare the
cross-environment MIRA-AC policy with untrained MIRA on 400 matched evaluation
pairs (800 trajectories): 100 pairs each from symbolic regression, Physics
Discovery, HillClimbBench-CPU, and BNLearn. Each pair contains the two policies'
trajectories on the same evaluation instance. We call the held-out evaluation
result the \emph{gold outcome}. In both systems, an outer policy
selects the next research direction, an inner executor carries it out, and the
harness manages execution and can force continuation after an incomplete
iteration.

\subsection{Analysis Protocol}
\label{app:mira-ac-behavior-protocol}

We use GPT-5.5 as an LLM judge in two passes. In the first pass, the judge sees
the sequence of work orders, executor reports, and proxy feedback under
anonymous labels A and B, but not the policy identities or gold outcomes.
It records what each trajectory's policy and executor did, how they responded
to evidence, and whether the policy's
next action could resolve an uncertainty visible at the time. This assessment
is fixed before the second pass.

In the second pass, the judge receives this fixed assessment and each side's
gold outcome; policy identity remains hidden. It separately assesses
whether a decision was reasonable when made and whether it ultimately helped
or harmed the gold result. The judge also attributes each claimed
contribution among the outer policy, inner executor, harness, and unobserved
factors. The pair-level verdict used below asks which side's outer-policy
decisions contributed more to its own final outcome. The
verdict excludes executor-only and harness-only contributions. Policy identity
is revealed only after both passes are complete. All 400 pairs produced valid judgments in both
passes.

For behavior rates, the unit is a trajectory, not an action. A trajectory is
counted once for a behavior if the second-pass judge identifies at least one
instance attributable at least partly to the outer policy; repeated instances
do not increase the count. Behaviors attributed only to the executor, harness,
or unobserved factors are excluded. The exception is \emph{productive risk}, a
whole-trajectory label indicating that a reasonable exploratory bet ultimately
improved the gold result.
Every rate therefore has a denominator of 400 trajectories per policy.

The matched comparisons describe observed differences between the policies.
They do not estimate the effect of independently inducing a single behavior:
executor stochasticity, interactions among decisions, and the evaluation
environment can also affect the final result. The aggregate rates establish
how common each behavior was; the excerpts below show what those behaviors
looked like in individual trajectories.

\subsection{Pooled Behavioral Changes}
\label{app:mira-ac-behavior-pooled}

The outcome-aware judge credited MIRA-AC's outer-policy decisions with the
larger contribution in 203 pairs and untrained MIRA's in 131; 66 pairs were
tied or unclear. MIRA-AC therefore received 60.8\% of the 334 decisive
attributions. It was also more often classified as taking a defensible risk
that improved the gold result (210/400 trajectories, 52.5\%, versus
139/400, 34.8\%).

The remaining rows of Fig.~\ref{fig:mira-ac-behavior} describe what changed
in the research process. Relative to untrained MIRA, MIRA-AC more often
changed its explanation of the underlying mechanism or the form of its solution
(185/400 vs.\ 148/400),
ran a test that distinguished competing ideas (112/400 vs.\ 67/400), checked
whether an improvement held across conditions or edge cases (28/400 vs.\
10/400), and made an evidence-supported decision about whether to continue
(53/400 vs.\ 29/400). It less often repeated minor variants that yielded
little new information (75/400 vs.\ 134/400).

\subsection{Selecting and Attributing Trajectory Examples}
\label{app:mira-ac-behavior-attribution}

For each environment, the examples include at least one
case in which MIRA-AC produced the better gold result and one in which it
performed worse than untrained MIRA. We select cases only after computing the
aggregate statistics across all trajectories, favoring cases with a clear chain from the
available evidence to an outer decision, an executor result, and the gold
outcome. Additional cases are included only when they clarify a distinct
behavior or the boundary between policy and executor credit. The examples are
thus illustrative rather than estimates of how common each behavior was, and
every quotation is linked to its original
decision or result record.

A case in which training hurt performance is included only when it reveals an interpretable decision
tendency consistent with the training feedback or data distribution, not
merely because MIRA-AC lost the pair. Its discussion reconstructs why the
choice was reasonable given the information available at the time, explains
why the proxy evidence favored a decision that later performed poorly on the
gold evaluation, and states any suggested remedy
as a testable hypothesis rather than an established correction.

A useful idea reported by the executor can still have been elicited by the
outer loop. We distinguish five relationships between the outer decision and
the eventual idea:

\begin{enumerate}
  \item \textbf{Direct outer specification.} The outer work order names the
  useful hypothesis, test, or planned response before execution.

  \item \textbf{Outer-seeded executor discovery.} The outer policy isolates
  the relevant uncertainty or asks for a test likely to resolve it. The
  executor then supplies the concrete mechanism or implementation.

  \item \textbf{State scaffolding.} Earlier outer decisions create or preserve
  a useful best candidate or body of evidence, but do not point the executor
  toward the eventual idea.

  \item \textbf{Executor-independent pivot.} The executor introduces a useful
  direction outside the requested experiment and its stated contingencies.

  \item \textbf{Harness-enabled opportunity.} Forced continuation or retry
  makes later work possible but contributes no research hypothesis. Any later
  outer decision or executor discovery is credited separately.
\end{enumerate}

These categories describe an idea's provenance, not whether it was good. We
require a specific link in content between a work order and a later discovery.
An executor idea is not outer-seeded merely because the work order came first,
requested a progress update, or was followed by an iteration forced by the
harness.

\subsection{Symbolic Regression}
\label{app:behavior-symbolic}

In symbolic regression, productive research means identifying which variables
and mathematical form explain the data, testing that explanation where it is
most likely to fail, and resisting improvements that depend on a feature
specific to the proxy validation set. Across the 100 matched tasks, MIRA-AC obtained the
higher gold score in 58 cases, compared with 32 for untrained MIRA (10
ties). The LLM trajectory judge more often found that MIRA-AC ran a test
designed to distinguish two explanations (14\% vs.\ 6\%), checked robustness
or an edge case (12\% vs.\
4\%), and made a well-supported decision to stop or continue (25\% vs.\ 11\%). It
also found fewer premature stops (13\% vs.\ 29\%) and fewer stretches of
repeated minor variants that added little information (3\% vs.\ 13\%). These rates count whether a
behavior appeared at least once in a trajectory; the matched cases below
illustrate how the same tendency to act on specific evidence could help or
hurt.

\paragraph{Testing a specific error reveals a better equation.}
In MatSci25 (pair \texttt{symbolic:047}), the task asks for a compact
equation for stress as a function of
strain $\epsilon$ and temperature $T$. The proxy fit suggested that the
coefficient relating stress to strain varied with temperature, but its worst
relative errors remained too
large. MIRA-AC made that error the target of the next experiment:

\begin{quote}\small
\textbf{Outer policy (\texttt{B-D002}):} ``Scan alternative parametric forms for the
temperature-dependent modulus $E(T)$ in Stress $\approx \epsilon E(T)$ to
reduce max relative error and potentially achieve
\texttt{acc\_0\_1}.''

\textbf{Executor (\texttt{B-R002}):} ``epsilon and T are perfectly correlated ($r\approx1.0$)
in the data, so fitting $\epsilon^p E(T)$ with $p\ne1$ reduces MRE
substantially.''
\end{quote}

The resulting expression reduced the maximum relative
error from 0.140 to 0.052. MIRA-AC scored 0.9949 on the gold evaluation;
untrained MIRA scored 0. This was an outer-seeded executor discovery: the outer policy
selected the relationship to inspect and the error criterion, which exposed the
$\epsilon$--$T$ correlation, while the executor introduced the decisive
generalization from $\epsilon$ to $\epsilon^p$. The result therefore
supports credit for the policy's choice of question, not sole credit for the
formula.

\paragraph{Optimizing proxy error preserves a model that does not
extrapolate.}
In the CRK25 reaction-kinetics task (pair \texttt{symbolic:020}), the target
depends on time $t$ and
concentration $A$. Untrained MIRA tested whether time added information after
conditioning on concentration; its executor reported
``Pearson(residual,t)=0.0145, confirming t carries no independent predictive
power once A is conditioned on'' (\texttt{A-R001}), then refined equations using concentration alone
until it passed the proxy relative-error threshold. MIRA-AC instead found a
more elaborate candidate with the best proxy score, but that expression used
a boundary value estimated from the observed data. Its own work order
identified the danger:

\begin{quote}\small
\textbf{Outer policy (\texttt{B-D005}):} ``The incumbent F34 expression [\ldots] achieves
\texttt{acc\_0\_1=True} with \texttt{max\_re=0.052} and
\texttt{NMSE=0.000689}, but has 5+ free parameters plus a hardcoded
\texttt{A\_max} from data maximum --- raising OOD generalization risk.''

\textbf{Executor (\texttt{B-R005}):} ``Discovered Family 38 (F38) --- combining rational
correction with A-correction in exponent --- achieving a 1133$\times$ NMSE
improvement.''
\end{quote}

The proxy gain was substantial: validation NMSE fell from
$6.89\times10^{-4}$ to $6.08\times10^{-7}$ while satisfying the relative-error
criterion. Retaining the model was therefore reasonable given the available
metrics, despite the known extrapolation risk. The gold result
reversed the proxy ordering: untrained MIRA scored 0.9998 and MIRA-AC scored 0. The
executor fitted and compared the candidate equations, but the outer policy
directly set the criterion for choosing the final equation and selected the formula after naming
its weakness. Forced continuation preserved later opportunities to
work, but supplied neither the candidate equation nor the selection criterion.

\emph{Interpretation.} Training did not simply make the model change the
mathematical form of its answer more often: the judge attributed such a change
to the outer policy, or jointly to the outer policy and executor, in 27\%
of MIRA-AC trajectories and 36\% of untrained MIRA trajectories, but those trajectories
produced the higher gold score 66.7\% and 47.2\% of the time,
respectively. The two cases show when this tendency helps and when it hurts. A
test tied to a systematic pattern in the errors can reveal a better equation;
the same pressure to improve a proxy error can favor an expression that uses
a constant fitted at the edge of the observed data and may not extrapolate.
One testable remedy is to evaluate expressions containing such constants on
held-out examples near and beyond the range seen during fitting.

Forced continuation appeared in 57 MIRA-AC trajectories and 17 untrained MIRA
trajectories in this environment. The attribution analysis excludes
harness-only contributions, but forced continuation gave the policies unequal
opportunities to make later decisions. This weakens a causal interpretation of
the aggregate difference.

\subsection{Physics Discovery}
\label{app:behavior-physics}

Physics Discovery gives the agent a limited budget for simulator experiments and
then evaluates its submitted law on hidden trajectories. Productive research
therefore uses an experiment to distinguish physical explanations or test
conditions in
which they disagree; repeatedly fitting the same acquired trajectories need
not produce new evidence. Across 100 matched tasks, MIRA-AC obtained the
higher gold score in 65 cases and untrained MIRA in 26 (nine ties). The LLM judge more
often observed MIRA-AC revising its mechanism (51\% vs.\ 22\%), running a
discriminating experiment (28\% vs.\ 15\%), and testing an edge case (13\%
vs.\ 4\%). Repeated minor variations of an unproductive idea fell from 51\% to
23\%, and premature stopping from 27\% to 6\%. Perfect gold scores rose
only from eight to ten;
much of the outcome gain was better partial recovery of the hidden law.

\paragraph{Varying one parameter separates two candidate laws.}
In pair \texttt{physics:012}, both trajectories inferred a plausible
distance-dependent interaction, but the dependence on the second particle's
parameter remained unresolved. Untrained MIRA stopped after five research
iterations even though
its executor reported that ``14 experiment rounds and unlimited MSE
evaluations remain for further improvement'' (\texttt{A-R002}). MIRA-AC eventually isolated
the ambiguity:

\begin{quote}\small
\textbf{Outer policy (\texttt{B-D013}):} ``The cheapest discriminating action is one experiment
round varying p2 independently.''

\textbf{Executor (\texttt{B-R014}):} ``Discovered the correct charge model: $p_1/p_2$ (not
$p_1/\sqrt{p_2}$ as previously believed).''
\end{quote}

MIRA-AC then preserved its best prior candidate when later changes to the
numerical integration failed to
improve the fit. It received a gold score of 1.000, compared with 0.005
for untrained MIRA. The outer policy selected the controlled parameter
variation and, in the following work order, made the competing coupling laws
explicit. The executor ran the experiment and identified the exact relation.
This is not an independent executor pivot. A
forced continuation preserved the
opportunity to run the following iteration, but the preceding policy decision
had already selected the informative parameter variation.

\paragraph{After finding the law, further exploration chases a proxy metric.}
Pair \texttt{physics:001} shows the same willingness to act on uncertainty
after the model had largely resolved which physical law applied. MIRA-AC's
executor first reported, ``Discovered the force law: F = -p1*p2 * r\_hat /
r$^2$'' (\texttt{A-R002}). The outer policy nevertheless redirected the remaining experiment
budget:

\begin{quote}\small
\textbf{Outer policy (\texttt{A-D007}):} ``pivot directly to acquiring diverse trajectories to
improve \texttt{trajectory\_shaping\_quality}.''

\textbf{Executor (\texttt{A-R007}):} ``0.006 gain over 6 rounds.''
\end{quote}

Untrained MIRA instead instructed its executor to ``probe distance scaling, sign
reversal, and magnitude coupling'' (\texttt{B-D002}), validated the inverse-square law, and
scored 1.000 on the gold evaluation. MIRA-AC exhausted all 15 experiment
rounds and scored 0.00085. Acquiring broader trajectories could reasonably
have exposed a hidden failure and did improve the visible
\texttt{trajectory\_shaping\_quality} score, but
the new trajectories no longer distinguished between competing physical laws. The costly
pivot was directly specified by the outer policy; the executor implemented it.
Forced continuation made later work possible but supplied no scientific
direction.

\emph{Interpretation.} The principal change in Physics Discovery was not
simply that MIRA-AC did more work. It was more willing to run a new experiment
when uncertainty remained.
That tendency helped when the uncertainty distinguished physical laws and
hurt when the policy continued optimizing a proxy metric after competing
physical laws were no longer in question. A policy trained to seek incremental
improvements in proxy score could learn
this preference, although the matched trajectories alone cannot establish
that explanation. A direct follow-up is to require each costly experiment to
state which competing physical laws it is meant to distinguish, or stop
collecting data solely to improve the proxy metric after several rounds with
negligible gains.

\subsection{HillClimbBench-CPU: Larger Architectural Changes}
\label{app:behavior-hcb}

HillClimbBench-CPU asks the agent to improve a neural architecture under a
fixed CPU training budget. Productive research must therefore do more than
change hyperparameters: it must identify a computational bottleneck, propose
an architecture that addresses it, and distinguish a genuine improvement from
a proxy-evaluation shortcut. Across 100 matched pairs, MIRA-AC achieved the
higher gold score in 55, compared with 45 for untrained MIRA. Its trajectories were more often
judged to change the model's computation rather than only its settings
(73\% vs.\ 60\%), or to take a
productive risk (66\% vs.\ 55\%), and less often remained in an unproductive
sequence of minor variants after they stopped improving (23\% vs.\ 41\%). The shift was not uniformly beneficial:
continuing despite contrary evidence was also more common (26\% vs.\ 10\%).
When the judge examined the research process without seeing the final scores,
it preferred untrained MIRA in 53 pairs and MIRA-AC in 45 (two ties).
MIRA-AC's gold-score advantage therefore came from decisions with larger
effects on the final outcomes, not from a uniformly better research process.

\paragraph{An architecture change that matched the task.}
In the scaled-Cayley forgetting task, the model had to retain a Cayley-rotation
structure while extrapolating to longer horizons and larger amplitudes. After
the current architecture reached an extrapolation ceiling, MIRA-AC's outer
policy identified the restrictive component: ``Allowing
\texttt{scale > 1} enables growth modes while preserving the Cayley rotation
structure'' (\texttt{B-D006}). The executor implemented a clamped exponential scale and
reported, ``Score: 0.5710 (delta +0.4008 vs incumbent 0.1703),'' with
extrapolation rising from 0.072 to 0.615 (\texttt{B-R006}). The final gold score was 0.683,
compared with 0.195 for the matched untrained trajectory, which continued to explore
nearby direct-mapping variants.

The ability to model growth was specified directly by the outer policy. A
later executor report introduced a diagonal preconditioner, which scales
different internal modes by different amounts, but only after the outer policy
had focused the search on scaling those modes; we therefore
treat that detail as an outer-seeded executor discovery rather than an
independent pivot. The executor supplied part of the implementation, while the
outer loop had already identified the model's inability to scale different
internal modes as the bottleneck.

\paragraph{The same willingness to change the architecture can follow the
wrong signal.}
The Universal Transformer task evaluates a model that repeatedly applies the
same block and learns when to stop while solving cellular automata. MIRA-AC
instead ordered the executor to
``Switch to a direct local-neighbor transition'' (\texttt{B-D001}). This was plausible given the available evidence:
cellular-automaton updates are local, and successive refinements raised the
proxy score to 0.985. The gold evaluator nevertheless assigned the final
candidate a score of 0 after marking it invalid. Untrained MIRA preserved the benchmark's
intended design---``Single weight-tied attention+MLP block reused
across steps'' (\texttt{A-D001})---and scored 0.809 on the gold evaluation.

Both architectures were specified directly by their respective outer
policies; the contrast is not explained by an independent executor discovery
or a harness-provided hypothesis. The trained policy made a meaningful
architectural bet, but the proxy signal rewarded a solution tailored to the proxy evaluator
that did not survive the gold evaluation. The zero alone does not
show whether the candidate violated the intended architecture or merely failed
because of an implementation error.
Because the actor was trained to seek incremental improvements in proxy
score, selecting a rapidly improving proxy candidate was aligned with its
training signal even
though that candidate failed gold evaluation.
One useful safeguard would require architecture-changing work orders to
state which intended architectural constraint they preserve and to pass basic
validity tests across input shapes before being selected as the best candidate.
Whether this retains the gains from architecture
search while filtering proxy shortcuts remains to be tested.

Together, the two cases clarify the aggregate shift. MIRA-AC more often pivoted
from repeated minor variants to changes in the computation itself. That behavior
created value when the proposed architecture addressed the true bottleneck,
but produced a larger downside when the proxy evaluator rewarded an easier
shortcut.

\subsection{BNLearn: Resolving Remaining Graph Uncertainty}
\label{app:behavior-bnlearn}

BNLearn asks the agent to recover a causal directed acyclic graph from
observational data and a limited intervention budget. An intervention fixes
one variable to a chosen value and reveals how other variables respond. Useful exploration is
usually conditional: choose an intervention that separates a few remaining
graph hypotheses, or search a small, explicitly enumerated set of edge edits
after localizing the remaining errors.
We summarize those errors using structural Hamming distance (SHD), the number
of missing, extra, or reversed edges.
Because many trajectories reach the score ceiling, different research paths
often have the same final score. The outcome-aware judge credited MIRA-AC
with decisions that contributed more to its own final outcome in 25 of 100
pairs and untrained MIRA in 22, with 53 ties. MIRA-AC trajectories more often
contained a test aimed at resolving uncertainty about a specific edge (55\%
vs.\ 34\%) and a productive risk (61\%
vs.\ 40\%), but less often preserved the strongest current graph while deciding
what to try next (19\%
vs.\ 32\%). BNLearn was held out from actor training, although it was included
in critic pretraining; these differences therefore reflect behavior transferred
from other training environments.

\paragraph{Targeted interventions fix the remaining graph errors.}
On \texttt{bnlearn:004}, MIRA-AC reduced its graph to one missing, one extra,
and one reversed edge. Its outer policy then selected two interventions:
``Intervene on \textbf{X0}'' and ``Intervene on \textbf{X16}''
(\texttt{A-D003}). The executor
reported, ``Fixed all 3 SHD errors using 2 strategic interventions (X0,
X16)'' (\texttt{A-R003}), identifying the precise reversal, deletion, and addition implied by
the new evidence about which variables were affected. MIRA-AC achieved a
gold score of 1.0;
untrained MIRA repeatedly launched related combinatorial searches after
visible wall-time failures and finished at 0.833.

The outer policy receives direct credit for selecting the two informative
interventions and the narrowly defined correction problem. The executor receives credit
for identifying the exact three edge edits; because those edits followed from
the requested interventions, this is an outer-seeded executor discovery rather
than an executor-independent pivot. The useful behavior was not exploration in
the abstract, but spending scarce interventions on observations that could
change a small set of decisions.

\paragraph{Broad search loses to a one-edge repair.}
The reverse pattern appears on \texttt{bnlearn:057}. Untrained MIRA used two
interventions to raise its score for recovering which variables affect one
another from 0.935 to 0.982, then issued
the narrow work order ``close the SHD=1 gap by identifying and adding the
single missing edge'' (\texttt{B-D009}). The executor compared the effects
implied by the current graph with the variables affected by each intervention
and reported, ``Identified and added the single
missing edge X15->X14 to close the SHD=1 gap'' (\texttt{B-R009}). Untrained MIRA therefore reached a
gold score of 1.0. MIRA-AC's reconstruction based on its interventions made
substantial progress, but its early searches incurred wall-time failures and
later graph-learning methods and exhaustive edge-swap searches concluded that the current
SHD=7 graph was ``the best achievable'' (\texttt{A-R011}). It finished at 0.844.

Untrained MIRA's outer policy directly asked the executor to find the one
missing edge; the
executor supplied the exact edge, making the solution an outer-seeded executor
discovery. MIRA-AC's broader search was reasonable after several
methods agreed on the same best graph, but it failed to turn the available
evidence into a sufficiently narrow correction. This is consistent with a
preference learned in other environments for gathering new evidence, even in
an environment
where many near-ceiling solutions require only a small repair to the
remaining graph errors. A late-stage work order could instead test the smallest
set of edge edits consistent with the intervention
evidence before rebuilding the graph or expanding to a broad search.

BNLearn therefore shows both sides of the same tendency. Targeted experiments
can close the last graph errors efficiently, but exploration
that begins before the remaining errors are identified can spend the remaining budget
without improving the graph. The large number of tied outcomes also means that
many differences in behavior do not appear in final-score comparisons.

\subsection{Cross-Environment Interpretation}
\label{app:behavior-synthesis}

Across the four environments, training did not simply increase the amount of
exploration. The trained outer policy more often acted on unresolved uncertainty:
run a test that could separate competing explanations, change the form of the
solution when the current class of solutions stopped improving, or abandon a line of minor variants
that had stopped yielding new information. In Physics Discovery and
HillClimbBench-CPU this often produced
larger changes to the proposed physical law or neural architecture. In
symbolic regression and BNLearn
it more often appeared as a controlled test or a focused attempt to
resolve the remaining errors.

The same shift explains the observed degradations. Acting on uncertainty was
valuable when resolving that uncertainty improved the gold evaluation, but
costly when a proxy metric rewarded a shortcut, a constant fitted to the edge
of the observed data, or additional data
collection after the central question was already resolved. The trained policy
therefore made the model's research decisions more likely to change the final
result, for better or worse, rather than uniformly more rigorous. The paired examples suggest three concrete directions for future
evaluation: require costly experiments to name the hypotheses they separate,
test architecture and expression changes on deliberately shifted or
out-of-distribution cases before selecting them as the final answer, and prefer
the smallest set of possible edits once only a few uncertainties remain.

\end{document}